\RequirePackage{etoolbox}
\csdef{input@path}%
{%
 {sty/},
 {img/},
}
\documentclass{elsevierbook}

\usepackage{csvsimple,booktabs,longtable,siunitx}
\usepackage{tabularx}

\usepackage{comment}

\usepackage{natbib}

\usepackage{graphicx}
\usepackage{subcaption} 

\usepackage[normalem]{ulem}
\usepackage{xcolor}

\usepackage{hyperref}

\usepackage{booktabs}  

\usepackage{algorithm}
\usepackage{algorithmic}

\usepackage{amsthm}
\newtheoremstyle{remarkstyle}
  {3pt}   
  {3pt}   
  {\normalfont} 
  {}      
  {\bfseries} 
  {.}     
  {1.0em}  
  {}      

\theoremstyle{remarkstyle}
\newtheorem{remark}{Remark}
\newtheorem{example}{Example}

\newcommand{\remarkend}{\hfill$\diamond$}

\usepackage{cleveref} 
\crefformat{equation}{(#2#1#3)}
\crefrangeformat{equation}{(#3#1#4--#5#2#6)}
\crefmultiformat{equation}{(#2#1#3)}{, (#2#1#3)}{, (#2#1#3)}{, (#2#1#3)}

\newcommand{\mb}[1]{\textcolor{blue}{Martin: \ifmmode #1 \else \textnormal{#1} \fi}}
\newcommand{\am}[1]{\textcolor{magenta}{Andreas: \ifmmode #1 \else \textnormal{#1} \fi}}
\newcommand{\xw}[1]{\textcolor{purple}{Victor: \ifmmode #1 \else \textnormal{#1} \fi}}
\newcommand{\ar}[1]{\textcolor{cyan}{Audrey: \ifmmode #1 \else \textnormal{#1} \fi}}

\definecolor{alexpink}{RGB}{231,0,115}
\newcommand{\alexdel}[1]{%
  \ifmmode
    \text{\sout{\ensuremath{#1}}}
  \else
    \sout{#1}
  \fi
}
 
\newcommand{\obs}{\ensuremath{y}}
\newcommand{\Meas}{\ensuremath{H}}
\newcommand{\Lin}{\ensuremath{L}}
\newcommand{\var}{\ensuremath{x}}
\newcommand{\dual}{\ensuremath{u}}
\newcommand{\aux}{\ensuremath{u}}

\newcommand{\GLin}{\ensuremath{\boldsymbol{W}}}
\newcommand{\Glin}{\ensuremath{{W}}}
\newcommand{\GBias}{\ensuremath{\boldsymbol{b}}}
\newcommand{\GBiasd}{\ensuremath{\boldsymbol{d}}}
\newcommand{\Gbias}{\ensuremath{{b}}}
\newcommand{\Gvar}{\ensuremath{\boldsymbol{u}}}
\newcommand{\Gaux}{\ensuremath{\boldsymbol{z}}}
\newcommand{\GM}{\ensuremath{\boldsymbol{M}}}
\newcommand{\GK}{\ensuremath{\boldsymbol{K}}}
\newcommand{\GV}{\ensuremath{\boldsymbol{V}}}
\newcommand{\net}{\ensuremath{\mathcal{N}}}
\newcommand{\lr}{\ensuremath{\eta}}
\newcommand{\diff}{G}

\newcommand{\nosamples}{s} 
\newcommand{\nobatches}{{N_B}} 
\newcommand{\nolayers}{J} 
\newcommand{\codedim}{S}

\newcommand{\reghyp}{\rho} 

\newcommand{\minimize}[1]{\ensuremath{\underset{#1}{\operatorname{minimise}\;}}}

\newcommand{\argmind}[1]{\ensuremath{\underset{#1}{\operatorname{argmin}\;}}}

\newcommand{\Id}{\ensuremath{\operatorname{Id}}}
\newcommand{\soft}[1]{\ensuremath{\mathcal{T}_{#1}}}
\newcommand{\proj}[1]{\ensuremath{\operatorname{proj}_{#1}}}
\newcommand{\RR}{\ensuremath{\mathbb{R}}}
\newcommand{\NN}{\ensuremath{\mathbb{N}}}

\newcommand{\ab}{\ensuremath{\mathbf{a}}}

\newcommand{\ub}{\ensuremath{\mathbf{u}}}
\newcommand{\vb}{\ensuremath{\mathbf{v}}}

\newcommand{\Kb}{\ensuremath{\mathbf{K}}}
\newcommand{\Mb}{\ensuremath{\mathbf{M}}}

\newcommand{\Wb}{\ensuremath{\mathbf{W}}}

\newcommand{\actfct}{\ensuremath{\boldsymbol{\sigma}}}

\newcommand{\loss}{\ensuremath{\mathcal{L}_{\var_i}}}
\newcommand{\uniloss}{\ensuremath{\mathcal{E}_{\actfct}}}

\newcommand{\defeq}{\ensuremath{\mathrel{\mathop:}=}}
\newcommand{\eqdef}{\ensuremath{=\mathrel{\mathop:}}}

\makeatletter
\renewcommand{\partmark}[1]{} 

\makeatother
\begin{document}

\Frontmatter

\Mainmatter
  \begin{frontmatter}

\chapter{A Unified Framework for Lifted Training and Inversion Approaches}\label{chap1}

\begin{aug}
\author[addressrefs={ad1}]%
  {\fnm{Xiaoyu}   \snm{Wang}}%
\author[addressrefs={ad2}]%
  {\fnm{Alexandra}   \snm{Valavanis}}%
\author[addressrefs={ad3}]%
  {\fnm{Azhir}   \snm{Mahmood}}%
\author[addressrefs={ad4}]%
  {\fnm{Andreas}   \snm{Mang}}%
\author[addressrefs={ad3}]%
  {\fnm{Martin}   \snm{Benning}}%
\author[addressrefs={ad1}]%
  {\fnm{Audrey}   \snm{Repetti}}%
\address[id=ad1]%
  {Heriot-Watt University, Edinburgh, UK}%
\address[id=ad2]%
  {Queen Mary University of London, London, UK}%
\address[id=ad3]%
  {University College London, London, UK}%
\address[id=ad4]%
  {University of Houston, Houston, TX, USA}%
\end{aug}


\begin{abstract}
The training of deep neural networks predominantly relies on a combination of gradient-based optimisation and back-propagation for the computation of the gradient. While incredibly successful, this approach faces challenges such as vanishing or exploding gradients, difficulties with non-smooth activations, and an inherently sequential structure that limits parallelisation. Lifted training methods offer an alternative by reformulating the nested optimisation problem into a higher-dimensional, constrained optimisation problem where the constraints are no longer enforced directly but instead penalised with penalty terms. This chapter introduces a unified framework that encapsulates various lifted training strategies---including the \emph{Method of Auxiliary Coordinates} (\emph{MAC}), \emph{Fenchel Lifted Networks}, and \emph{Lifted Bregman Training}---and demonstrates how diverse architectures, such as \emph{Multi-Layer Perceptrons} (\emph{MLPs}), \emph{Residual Neural Networks} (\emph{ResNets}), and \emph{Proximal Neural Networks} (\emph{PNNs}), fit within this structure. By leveraging tools from convex optimisation, particularly Bregman distances, the framework facilitates distributed optimisation, accommodates non-differentiable proximal activations, and can improve the conditioning of the training landscape. We discuss the implementation of these methods using \emph{block-coordinate descent} (\emph{BCD}) strategies—including deterministic implementations enhanced by accelerated (e.g., Nesterov, Heavyball) and adaptive (e.g., Adam) optimisation techniques—as well as \emph{implicit stochastic gradient methods} (\emph{ISGM}). Furthermore, we explore the application of this framework to inverse problems, detailing methodologies for both the training of specialised networks (e.g., unrolled architectures) and the stable inversion of pre-trained networks. Numerical results on standard imaging tasks validate the effectiveness and stability of the lifted Bregman approach compared to conventional training, particularly for architectures employing proximal activations.
\end{abstract}

\begin{keywords}
\kwd{Lifted Training}
\kwd{Inverse Problems}
\kwd{Bregman Distances}
\kwd{Proximal Neural Networks}
\kwd{Distributed Optimisation}
\kwd{Algorithm Unrolling}
\kwd{Network Inversion}
\end{keywords}

\end{frontmatter}

\section{Introduction}\label{sec1}

Deep neural networks have achieved state-of-the-art performance across a wide range of machine learning tasks. The present paper proposes a unified framework for lifted training with applications to solving (\emph{linear}) inverse problems. Lifted training provides an alternative to gradient-based optimisation with back-propagation by reformulating network training as a structured optimisation problem. Its ability to accommodate non-smooth activations, improve conditioning, and support distributed computation makes it a powerful and theoretically grounded framework for modern deep learning.

The predominant training strategy for deep neural networks relies on (stochastic) first-order optimisation, where gradients are typically computed based on the back-propagation algorithm~\cite{rumelhart1986learning}. Despite its success, this approach faces significant challenges. The optimisation landscape of neural networks is highly non-convex and complex; early theoretical work highlighted this difficulty, demonstrating, for instance, the existence of exponentially many local minima even for single neurons~\cite{auer1995exponentially}. In deep architectures, back-propagation suffers from vanishing and exploding gradients~\cite{he2016deep, haber2017stable, pascanu2013difficulty, bengio1994learning}, requires subgradient methods for non-differentiable activations, and is inherently sequential, hindering efficient distributed training. Moreover, the performance of gradient-based methods can be suboptimal due to poor conditioning of the optimisation problem. While numerous stabilisation techniques have been proposed~\cite{haber2017stable, he2016deep, ioffe2015batch, pascanu2013difficulty, maas2013rectifier}, achieving stable and accurate generalisation on unseen data remains challenging and often requires extensive hyperparameter tuning.
To overcome these limitations, several alternative training strategies have emerged~\cite{zach2019contrastive, zhang2017convergence, gu2020fenchel, taylor2016training, carreira2014distributed, frecon2022bregman, combettes2023variational, wang2023liftedtraining}.

A promising line of research is \emph{lifted training}, which reformulates the training problem by augmenting the variable space with auxiliary variables. Early approaches such as the \emph{Method of Auxiliary Coordinates with Quadratic Penalty} (\emph{MAC-QP})~\cite{carreira2014distributed} replaced hard constraints with quadratic penalties. While conceptually appealing, these methods often still required differentiating non-smooth activations and were largely restricted to affine-linear networks. Instead, recent works build on this idea using \emph{lifted Bregman training}~\cite{wang2023liftedtraining, wang2023liftedinversion, wang2024lifted, wang_2023}, which leverages Bregman distances to couple auxiliary variables with network activations. This reformulation offers several advantages. First, it enables distributed optimisation and circumvents the sequential bottleneck of back-propagation. Second, it grounds the architecture in established principles from convex optimisation, leading to more interpretable and potentially more robust models~\cite{hoier2020lifted, monga2021algorithm, chouzenoux2024stability}. By lifting the problem into a higher-dimensional space and employing Bregman distances, the method improves conditioning, supports flexible regularisation, and provides mathematical guarantees for convergence. Notably, gradients with respect to network parameters no longer require differentiating activation functions, allowing the use of non-smooth proximal maps as activations.
This structure naturally decomposes the training problem across layers, resulting in a collection of bi-convex subproblems.
Such formulations are well-suited for incorporating regularisation on the network's hidden variables, e.g., to enforce sparsity in latent representations—an increasingly important property in applications such as sparse autoencoders~\cite{bungert2022bregman, bungert2021neural, scardapane2017group, hoefler2021sparsity}. Compared to subgradient-based methods, lifted Bregman training leverages the problem structure and enables a wider range of algorithmic approaches to potentially achieve faster convergence and more stable training dynamics.

To place the proposed unified framework into context, we briefly introduce the main conceptual ideas underpinning inverse problems in the following section.

\subsection{Inverse and Ill-posed Problems}

In many scientific and engineering disciplines, we are faced with the challenge of determining the internal properties or causes of a system based on indirect and often incomplete observations~\cite{engl1996regularization, scherzer2009variational,benning2018modern, kaipio2006statistical, hansen2010discrete}. This fundamental challenge is the essence of an inverse problem, which can be abstractly formulated as solving the equation
\begin{equation}\label{eq:inverse_problem}
\Meas(\var) = \obs
\end{equation}
for an unknown quantity of interest, $\var$, given a set of measurements, $\obs$. The forward operator $\Meas$ models the physical process that maps the unknown $\var$ to the observable data $\obs$. In practice, the available measurements $\obs^\delta$ are corrupted by some form of noise, i.e., $\obs^\delta = \text{noise}^\delta(\obs)$. A typical example is additive noise in which case $\text{noise}^\delta(\obs) = \obs + \eta^\delta$ for some additive perturbation $\eta$ such that $\obs^\delta$ satisfies $\| \obs - \obs^\delta \| \leq \delta$.

Framework \eqref{eq:inverse_problem} encompasses a vast range of important applications~\cite{benning2018modern}. Typical examples in signal processing include denoising where the goal is to restore a clear image from a noisy one~\cite{tian2020deep,buades2005review}, inpainting where missing parts of an image must be filled~\cite{schonlieb2015partial,elharrouss2020image, bertalmio2000image}, or compressed sensing where the aim is to reconstruct a full signal from a limited number of samples~\cite{donoho2006compressed,candes2006robust,zhang2018ista}. Other important examples include deblurring (e.g., for a shaky photograph) or recovering phase information from intensity-only measurements (known as phase retrieval~\cite{shechtman2015phase, klibanov1995phase}).
In medical imaging, one might perform image reconstruction from computed tomography~\cite{natterer_mathematics_1986, natterer2001mathematical, adler2018learned, baguer2020computed} or magnetic resonance~\cite{alma9924436385902466, ansorge2016physics, Suetens_2017} to see inside a patient's body. In all these cases, a direct or naive attempt to solve for $\var$ is often doomed to fail due to a fundamental mathematical challenge: inverse problems are typically ill-posed~\cite{engl1996regularization, benning2018modern}.

A problem is considered well-posed in the sense of Hadamard and John if a solution exists, is unique, and depends continuously on the data~\cite{engl1996regularization, benning2018modern}. While non-existence or non-uniqueness can be significant issues, the most pervasive and critical challenge in practice is the violation of the third condition: \textit{stability}. For most inverse problems of interest, the inverse mapping $\Meas^{-1}$ is discontinuous or may not even exist. This implies that even small perturbations in the data $\obs$ can lead to large and physically meaningless errors in the reconstructed solution $\var$~\cite{scherzer2009variational}. This inherent instability makes any naive inversion approach fundamentally unworkable.

This abstract concept of a discontinuous inverse is typically analysed in infinite-dimensional function spaces like $L^2(\RR^2)$, where operators such as blurring are naturally defined. In this setting, the operator $\Meas$ is often compact, meaning its range is not closed and its Moore-Penrose inverse $\Meas^\dagger$ is unbounded~\cite{engl1996regularization}. In the finite-dimensional setting required for numerical computation, this instability manifests itself as ill-conditioning ~\cite{benning2018modern}.

When the problem is discretised, the image $\var$ becomes a vector and the operator $\Meas$ becomes a matrix. While the linear inverse problem is no longer formally discontinuous, the ill-posedness translates into ill-conditioning that lets the singular values of the matrix $\Meas$ decay rapidly towards zero~\cite{hansen2010discrete}. A naive (generalised) inversion, such as computing $\var = \Meas^{\dagger} \obs^\delta$ (where $\obs^\delta$ denotes noisy data), involves division by these small singular values. Since measurement noise typically has components across all frequencies, this division massively amplifies the noise associated with small singular values, resulting in a reconstructed solution that is completely dominated by noise and artefacts.

To overcome this fundamental instability, one has to abandon the idea of finding an exact solution to $\Meas(\var) = \obs^\delta$ and instead seek a stable, meaningful approximation. The mathematical framework for achieving this is known as \emph{regularisation} (cf.~\cite{engl1996regularization}). Regularisation methods replace the original ill-posed problem with a family of related, well-posed problems, controlled by a so-called regularisation parameter. The goal is to select a solution that is both stable under data perturbations and a good approximation of the true, underlying solution.

For decades, variational regularisation methods have been state-of-the-art to solve the inverse problem~\eqref{eq:inverse_problem}. This powerful and flexible paradigm recasts the inverse problem as an optimisation task (cf.~\cite{scherzer2009variational, chambolle2016introduction, benning2018modern, burger2021variational}). Instead of solving the operator equation directly, one seeks to find a solution $\var_\reghyp^\delta$ that minimises a functional of the general form
\begin{equation}
\var_\reghyp^\delta \in
\argmind{\var} \left\{\mathcal{D}(\Meas(\var), \obs^\delta)
+ \reghyp \mathcal{R}(\var) \right\} \,. \label{eq:variational_reg}
\end{equation}
This functional consists of three key components. The first term, $\mathcal{D}(\cdot, \cdot)$, is the \emph{data-fidelity term}, which enforces consistency between the reconstruction $\var$ and the measured data $\obs^\delta$. A common choice is the squared $\ell_2$-norm, $\mathcal{D}(\Meas(\var), \obs^\delta) = \frac{1}{2}\|\Meas(\var) - \obs^\delta\|_2^2$, which means that \cref{eq:variational_reg} corresponds to a  maximum a-posteriori estimate under the assumption of i.i.d. additive Gaussian noise.\footnote{We note that there is a direct relation between the statistical and deterministic interpretation of the inverse problem. In Bayesian terms we can view \cref{eq:variational_reg} as the negative log posterior, where the data fidelity term corresponds to the negative log likelihood and the regularisation term corresponds to the negative log prior. We refer to~\cite{calvetti2018inverse, stuart2010inverse, bui2013computational, martin2012stochastic} for more details about the connection between the Bayesian and the deterministic variational formulation.} The second term, $\mathcal{R}(\,\cdot\,)$, is the \emph{regularisation functional}, often simply called the \emph{prior} (in a statistical setting). This is the heart of the variational regularisation strategy. It encodes \emph{a priori} knowledge about the expected properties of the solution $\var$. By penalising undesirable features (such as excessive oscillation or noise), it effectively restricts the solution space to a set of ``plausible'' candidates, thereby ensuring stability. Standard choices include Tikhonov regularisation \cite{tikhonov1943stability,tikhonov1963solution,tikhonov1966stability} to enforce smoothness, and sparsity-based priors such as the Total Variation (TV) regularisation~\cite{rudin1992nonlinear} or $\ell_1$-based priors within wavelet domains~\cite{mallat1999wavelet}. Finally, the third component, $\reghyp > 0$, is the \emph{regularisation parameter} which is a crucial hyperparameter that controls the trade-off between fidelity to the data (a small value of $\mathcal{D}$) and adherence to the prior knowledge (a small value of $\mathcal{R}$). The selection of an appropriate $\reghyp$ is a classic problem in the field, as it balances the bias introduced by the regularisation functional against the variance resulting from noise amplification. Classical parameter-choice strategies include Morozov’s discrepancy principle criterion~\cite{tikhonov1977solution}, the L-curve~\cite{morozov1966regularization} or Bayesian hyperparameter selection~\cite{snoek2012practical}.

\subsection{The Rise of Deep Learning Strategies for Inverse Imaging Problems}

During the last decade, deep learning methods have been developed to solve~\eqref{eq:inverse_problem}.
Two main strategies can be distinguished: building neural networks which approximate the inverse of $\Meas$ for directly solving~\eqref{eq:inverse_problem}, and building regularising neural networks that can be incorporated in traditional optimisation approaches, i.e., replacing $\mathcal{R}$ by a learned prior (cf.~\cite{arridge2019solving}).
In both cases, the neural networks must be trained to perform the task of interest (e.g., deblurring, denoising, etc.). While such networks are typically trained using standard gradient-based methods, the nested nature of these architectures often leads to difficult non-convex optimisation landscapes. The lifted training strategies at the core of this chapter provide a powerful alternative framework for training neural networks in this and other contexts. Lifting is a general optimisation scheme designed to handle highly structured problems by decoupling the layers of the network. This transforms a deeply nested non-convex problem into a more manageable (often block-convex) problem; a problem that is amenable to parallel computation and particularly well-suited for the large-scale networks used in imaging.

In the context of this work, our goal is to solve inverse problems of the general form \cref{eq:inverse_problem} by training neural networks (or decoders) $\mathcal{N}$ that represent approximate surrogates of $\Meas^{-1}$, i.e., composing $\net$ with $\Meas$ exposes $x$: $\net(\Meas(x)) \approx x$.

\subsection{Contributions}\label{subsec:contributions}

The main contributions of this chapter are summarised as follows:
\begin{itemize}
    \item \textbf{A Unified Framework for Lifted Approaches:} We introduce a comprehensive mathematical framework (\Cref{sec:unified}) that encapsulates diverse neural network architectures (Perceptrons, MLPs, ResNets, and Proximal Neural Networks) and generalises various lifted training strategies (MAC-QP, Fenchel Lifted, Lifted Contrastive, and Lifted Bregman) under a single optimisation paradigm.
    \item \textbf{Synthesis of Lifted Methodologies:} We provide a detailed review of the evolution of lifted training (\Cref{sec:lifted_training}), clarifying the relationships between different approaches, such as the connection between Fenchel duality and Bregman distances in the context of proximal activations.
    \item \textbf{Implementation Strategies:} We detail practical implementation strategies, covering both deterministic block-coordinate descent (BCD) and the Implicit Stochastic Gradient Method (ISGM). We further highlight the integration of accelerated (Nesterov, Heavyball) and adaptive (Adam) optimisation methods within the linearised BCD variants for efficiently solving the lifted optimisation problem (\Cref{sec:implementation}).
    \item \textbf{Applications to Inverse Problems:} We bridge the gap between lifted optimisation and inverse problems by explicitly considering network inversion as the target inverse problem, summarising recent work on stable lifted network inversion~\cite{wang2023liftedinversion}, which focuses on input recovery from outputs and establishes convergence results for single-layer networks in the small-noise regime.
    \item \textbf{Numerical Evaluation:} We provide a numerical evaluation of the lifted Bregman training framework for canonical (imaging) inverse problems, demonstrating its effectiveness and stability compared to conventional training for architectures with proximal activations (\Cref{sec:numerical_results}).
    \item \textbf{Software Release:} The release (upon acceptance of this manuscript) of a modular software package implementing the unified framework for lifted neural network training.
\end{itemize}

\subsection{Outline and Notation}

In what follows, we first present how a variety of neural network architectures can be unified within a compact framework in \Cref{sec:unified}. \Cref{sec:mac} and \Cref{sec:lifted_training} then trace the evolution of distributed optimisation methods and lifted training strategies. Practical aspects of the lifted formulation, including implementation details and computational strategies, are discussed in \Cref{sec:implementation}. Applications to inverse problems are examined in \Cref{sec:applications_inverse_problems}, and numerical results on prototypical linear inverse problems and network inversion problems are presented in \Cref{sec:numerical_results}. Finally, the notation used throughout the paper is summarised in \Cref{tab:notation}. Please note that we use boldface notation for variables that contain all auxiliary variables for each layer, and normal font for individual, layer-independent variables. The same holds true for operators that act on variables that contain all auxiliary variables, for which we use boldface notation, while we use normal font for operators that act on individual variables. We denote the end of a remark by $\diamond$.

\begin{table}
\centering
\caption{Mathematical notation and symbols used throughout this chapter.}
\label{tab:notation}
\begin{tabular}{ll}
\toprule
\textbf{Symbol} & \textbf{Description} \\
\midrule
\multicolumn{2}{l}{\textit{Sets and Spaces}} \\
$\RR$ & Set of real numbers \\
$\RR^n$ & $n$-dimensional real vector space \\
$\NN$ & Set of natural numbers \\
$\mathbb{E}_i[\cdot]$ & Expectation over training samples \\
\\
\multicolumn{2}{l}{\textit{Network Architecture}} \\
$\net(\obs)$ & Neural network mapping input $\obs$ to output \\
$\actfct(\cdot)$ & Nonlinear activation function \\
$\GK, \GM, \GV, \GLin$ & Network operators (matrices) \\
$\GBias, \GBiasd$ & Bias vectors \\
$\Gvar, \Gaux$ & Hidden/auxiliary variables \\
$\theta$ & Collection of all learnable parameters \\

\\
\multicolumn{2}{l}{\textit{Training and Loss Functions}} \\
$\{(\obs_i, \var_i)\}_{i\in \mathbb{I}}$ & Training dataset with inputs $\obs_i$ and labels $\var_i$ \\
$\loss(\cdot)$ & Loss function for sample $i$ \\
$\uniloss$ & Total empirical risk functional \\
$\mathcal{C}(\cdot, \cdot)$ & Architectural constraint penalty function \\
$\mathcal{D}_{\actfct}(\cdot, \cdot)$ & Activation constraint penalty function \\
$\lambda, \mu$ & Penalty parameters (hyperparameters) \\
\\
\multicolumn{2}{l}{\textit{Inverse Problems}} \\
$\obs$ & Measurements/observations \\
$\obs^\delta$ & Noisy measurements/observations with noise level $\delta>0$ \\
$\Meas$ & Forward operator\\
$\mathcal{R}$ & regularisation functional\\
$\var$ & Signal/image to be reconstructed \\
$\dual$ & Dual/auxiliary variables \\
$\reghyp$ & regularisation strength hyperparameter \\
\\
\multicolumn{2}{l}{\textit{Lifted Training Methods}} \\
$\mathcal{B}_{\Psi}(\cdot, \cdot)$ & Bregman penalty function \\
$B_{\actfct}(\cdot, \cdot)$ & Biconvex function (Fenchel duality) \\
$\Psi(\cdot)$ & Convex, proper, lower-semicontinous potential function for proximity operators \\
$\Phi(\cdot)$ & Potential function for Bregman distances \\
$D_{\Phi}^{\xi}(\cdot, \cdot)$ & Bregman distance with respect to $\Phi$ and subgradient $\xi$. \\
$\text{prox}_{\Psi}(\cdot)$ & Proximity operator of function $\Psi$, defined as $\text{prox}_{\Psi}(\vb) = \argmind{\ub}  \frac{1}{2} \|\ub - \vb\|^2_2 + \Psi(\ub)$ \\
\\
\multicolumn{2}{l}{\textit{Proximal Networks and Operators}} \\
$\soft{\gamma\lambda}(\cdot)$ & Soft-thresholding operator \\
$\proj{C}(\cdot)$ & Projection operator onto set $C$ \\
$\Lin$ & Linear regularisation operator \\
$\gamma, \tau$ & Step-size parameters \\
$\chi_S(\cdot)$ & Characteristic function of set $S$ \\
$\|\cdot\|_1, \|\cdot\|_2$ & $\ell_1$ and $\ell_2$ norms, respectively \\
\\
\multicolumn{2}{l}{\textit{Optimisation}} \\
$\minimize{x}$ & Minimisation with respect to variable $x$ \\
$\argmind{x}$ & argmin with respect to variable $x$ \\
$\nabla_x f(x)$ & Gradient of function $f$ with respect to $x$ \\
$f^*(\cdot)$ & Fenchel conjugate of function $f$ \\
$\Id$ & Identity operator \\
$\alpha, \beta$ & Step-size parameters in iterative algorithms \\
\bottomrule
\end{tabular}
\end{table}

\section{A unified framework}\label{sec:unified}
The emergence (and abundance) of lifted neural network training approaches raises the question if we can develop a unified framework with which we can express different families of neural network architectures in a compact form, and express all conventional and lifted training approaches in one framework. To achieve this, we consider neural networks $\net$ of the form
\begin{subequations}\label{eq:nn}
\begin{align}
\net(\obs) & \defeq \GK\Gvar + \GBiasd\label{eq:nnpred} \\
\GM\Gvar &= \GV \Gaux \label{eq:nnpred_const1}\\
\Gaux &= \actfct\left( \GLin \Gvar + \GBias \right)\label{eq:nnpred_const2}
\end{align}
\end{subequations}
where $\obs \in \RR^M$ denotes the network input, and $\net(\obs) \in \RR^N$ denotes the network output. The variables $\Gvar$ and $\Gaux$ represent collections of hidden/auxiliary variables over all layers in the neural network architecture, while $\GBiasd$ and $\GBias$ are either fixed or learnable bias parameters, and $\GK$, $\GM$, $\GV$ and $\GLin$ are operators that model the network architecture and model parameters. The function $\actfct$ denotes a nonlinear activation function that acts on $\GLin \Gvar + \GBias$.
We will show how perceptrons, shallow neural networks, feed-forward neural networks, residual neural networks and proximal neural networks all fit into this framework in the upcoming subsections.
The network parameters are commonly found by minimising an empirical risk over a given training dataset $\{(\obs_i, \var_i)\}_{i\in \mathbb{I}}$, with $\mathbb{I}$ denoting the set of training data indices. 
Following the structure of \cref{eq:nn}, we can express a general training problem that consists of three key terms. The first term, $\loss(\GK\Gvar_i + \GBiasd)$, is the data fidelity term, where $\loss$ is a loss function (such as mean squared error or cross-entropy) that penalises the deviation of the network output (prediction) $\net(\obs)=\GK\Gvar_i + \GBiasd$ from the corresponding ground truth data $\var_i$. The remaining two terms, $\mathcal{C}(\GM\Gvar_i, \GV\Gaux_i)$ and $\mathcal{D}_{\actfct}(\Gaux_i, \GLin\Gvar_i + \GBias)$, relate to the constraints introduced in \cref{eq:nnpred_const1} and \cref{eq:nnpred_const2}. These functions determine how strictly the relationships between the network's internal variables are enforced. With these terms we then define the training objective as the minimisation of the functional
\begin{align}
\uniloss \defeq \mathbb{E}_{i}\left[ \loss(\GK\Gvar_i + \GBiasd) + \mathcal{C}(\GM\Gvar_i, \GV\Gaux_i) + \mathcal{D}_{\actfct}(\Gaux_i, \GLin\Gvar_i + \GBias) \right] \, , \label{eq:loss}
\end{align}
where $\mathbb{E}_{i}$ denotes the expectation over the training samples.
The terms $\mathcal{C}(\GM\Gvar_i, \GV\Gaux_i)$ and $\mathcal{D}_{\actfct}(\Gaux_i, \GLin\Gvar_i + \GBias)$  can model hard constraints, where any deviation is forbidden, or soft constraints, where deviations are penalised. For instance, in many lifted training schemes, these functions are quadratic penalties (e.g., $\mathcal{C}(\ab, \ab') = \frac{\mu}{2} \|\ab - \ab'\|^2_2$), which allows the constraints to be violated at a certain cost (controlled by some hyperparameter $\mu > 0$).

\begin{remark}[Conventional Training]\label{rem:conventinal-tranining}
We want to emphasise the case where the constraints of the network architecture in \cref{eq:nn} are strictly enforced. This corresponds to conventional neural network training. In our framework, this is achieved when $\mathcal{C}$ and $\mathcal{D}_{\actfct}$ are characteristic functions that enforce equality. Specifically, let $$\mathcal{C}(\ab, \ab') = \chi_{\{\mathbf{0}\}}(\ab - \ab') \defeq \begin{cases} 0 & \text{if } \ab = \ab' \\ +\infty & \text{if } \ab \neq \ab' \end{cases}$$ and similarly, let $\mathcal{D}_{\actfct}(\ub, \vb) = \chi_{\{\mathbf{0}\}}(\ub - \actfct(\vb))$. The minimisation of $\uniloss$ in \cref{eq:loss} then becomes the constrained optimisation problem
\begin{align}
\begin{split}
\minimize{\{\Gvar_i, \Gaux_i\}_i,\, \theta} \quad & \mathbb{E}_{i}\left[ \loss(\GK\Gvar_i + \GBiasd) \right] \\
\text{subject to} \quad & \GM\Gvar_i = \GV\Gaux_i \\
& \Gaux_i = \actfct\left( \GLin \Gvar_i + \GBias \right)
\end{split} \, , \label{eq:constrained_loss}
\end{align}
where $\{\Gvar_i\}_i, \{\Gaux_i\}_i$ denote the collection of  auxiliary and hidden variables associated with the entire training dataset, $\theta$ represents the collection of all learnable parameters within the operators $\GK, \GM, \GV, \GLin$ and biases $\GBias, \GBiasd$. The formulation in \cref{eq:constrained_loss} is an explicit representation of the standard training procedure for a neural network $\net$ as defined in \cref{eq:nn} (see, for example, \cite[Appendix A]{wang2023liftedtraining} for a similar derivation for feed-forward networks). Since the constraints must hold, the variables $\Gvar_i$ and $\Gaux_i$ are uniquely determined by the network architecture and its parameters for a given input $\obs_i$. Consequently, the problem boils down to minimising the loss with respect to the network's output, that is,
$$\uniloss = \mathbb{E}_i [ \loss(\net(\obs_i))] \, .$$
\remarkend
\end{remark}

Examples for training losses and different penalisations to enable lifted neural network training are presented in the upcoming sections. In the following subsections, we describe a few examples of network architectures that match the framework \cref{eq:nn}.

\subsection{Single-layer Perceptrons}

A perceptron is the simplest form of a neural network, first introduced as a concept in a 1957 technical report~\cite{rosenblatt1957} and more formally in a 1958 paper by Frank Rosenblatt~\cite{rosenblatt1958}. Rosenblatt's original model was a probabilistic, multi-layered system intended to model information storage in the brain. However, the term ``perceptron'' is now more commonly associated with the simplified, single-layer linear classifier that was the subject of intense analysis and critique \cite{minsky1969}. For an input $\obs \in \RR^{M}$, this simplified model is expressed as
\begin{align}
\net(\obs) = \sigma(\langle w, \obs \rangle + \Gbias) \, ,\label{eq:perceptron1}
\end{align}

\noindent where $\sigma \colon \RR\to\RR$ is the Heaviside step function, $w \in \RR^M$ are the connection weights, and $\Gbias \in \RR$ is the bias term. The bias is mathematically equivalent to the negated value of the threshold that must be overcome for the perceptron to activate.

This model, which uses a hard thresholding function, received a lot of attention due to its mathematical tractability and the famous \emph{Perceptron Convergence Theorem}~\cite{rosenblatt1962,novikoff1962,block1962}. Rosenblatt himself explored more complex multi-layer architectures in his later work~\cite{rosenblatt1962}; but the simplified version is what is traditionally associated with a single-layer perceptron.

\noindent We can introduce a scalar $\zeta \in \RR$, solely for notational compatibility with the unified framework, and define the two-layer vector
\[
\Gvar =
\begin{pmatrix}
y \\ \zeta
\end{pmatrix} \in \RR^{M + 1}
\quad\text{and set}\quad
\GLin = \begin{pmatrix}
w^\top & 0
\end{pmatrix} \in \RR^{1 \times (M + 1)}
\]
and $\GBias = \Gbias$. Then $\Gaux \in \RR$ is just a scalar, and we can pick $\GV = 1$, $\GK = \GM = \begin{pmatrix}
    0_{1 \times M} & 1
\end{pmatrix} \in \RR^{1 \times (M + 1)}$ and $\GBiasd = 0$ to bring~\cref{eq:perceptron1} into the form of~\cref{eq:nn}.
The activation function is historically chosen as the Heaviside step function, but the perceptron model can be generalised to arbitrary activation functions and higher-dimensional outputs.
More generally, we can consider perceptrons with a vector-valued output, of the form
\begin{align}
    \net(\obs) = \sigma( \Glin \obs + \Gbias ) \, ,\label{eq:preceptron2}
\end{align}
where $\Glin \in \RR^{N \times M}$ is a matrix and $b \in \RR^{N}$ a vector, and $\sigma \colon \RR^N \to \RR^N$ is now an activation function that acts on an $N$-dimensional vector.

\noindent Like before, we can introduce a vector ${\zeta} \in \RR^N$ and define
\[
\Gvar = \begin{pmatrix}
    y \\ {\zeta}
\end{pmatrix} \in \RR^{M + N}
\quad\text{and set} \quad
\GLin = \begin{pmatrix}
    \Glin & 0_{N \times N}
\end{pmatrix} \in \RR^{N \times (M + N)}
\]

\noindent and $\GBias = b$, where $0_{N \times N} \in \RR^{N \times N}$ denotes the matrix with $N$ rows and $N$ columns for which all entries are zero. This way, $\Gaux \in \RR^{N}$ is an $N$-dimensional vector and we can choose $\GV = I_{N}$ and $\GK = \GM = \begin{pmatrix}
    0_{N \times M} & I_{N}
\end{pmatrix} \in \RR^{N \times (M + N)}$, where $I_{N} \in \RR^{N \times N}$ denotes the identity matrix with $N$ rows and columns, respectively.

\subsection{Shallow Neural Networks}
A shallow neural network, commonly referred to as a two-layer or single-hidden-layer neural network, represents a critical step in the evolution of neural architectures. It directly addresses the primary limitation of the single-layer perceptron: the inability to model non-linearly separable functions. This limitation was highlighted by Minsky and Papert, who demonstrated that a single-layer perceptron cannot solve the simple XOR problem \cite{minsky1969}.

The introduction of a hidden layer of neurons with non-polynomial activation functions gives the network the ability to form much more complex decision boundaries. The theoretical power of this architecture is captured by the \emph{Universal Approximation Theorem}. This seminal theorem, established in works by Cybenko \cite{cybenko1989} and Hornik et al. \cite{hornik1989}, states that a shallow neural network with a finite number of neurons can approximate any continuous function on compact subsets of $\RR^M$ to any desired degree of accuracy. Hornik later showed that this is a property of the multilayer feedforward architecture itself, rather than the specific choice of a non-polynomial activation function \cite{hornik1991}. Further work by Barron provided explicit bounds on the approximation error in terms of the number of nodes \cite{barron1993}.

Next, we want to demonstrate how shallow networks fit the framework \cref{eq:nn}. We specifically focus on shallow networks of the form
\begin{align}
    \net(\obs) = \sum_{j = 1}^J c_j \, \sigma_j(w_j \, \obs + \Gbias_j) \, ,\label{eq:shallownn}
\end{align}
where $\obs \in \mathbb{R}$ and, for every $j\in \{1, \ldots, J\}$, $(c_j, w_j, b_j) \in \mathbb{R}^3$. By introducing auxiliary variables $(\aux_1, \ldots, \aux_J) \in \mathbb{R}^J$ we can write these shallow networks in constrained form as
\begin{equation*}
\net(\obs) = \sum_{j = 1}^J c_j \, \aux_j \, ,\qquad
\aux_j = \sigma_j(w_j \, \obs + \Gbias_j) \,,\,\, \forall \, j \in \{1, \ldots, J\} \,.
\end{equation*}
If we define $\Gvar = \begin{pmatrix}
    \obs & \aux_1 & \cdots & \aux_J
\end{pmatrix}^\top \in \RR^{J + 1}$, we can bring \cref{eq:shallownn} into the form \cref{eq:nn} by setting
\begin{align*}
    \GLin = \begin{pmatrix}
            w_1 & 0 & 0 & \cdots & 0\\
            w_2 & 0 & 0 & \cdots & 0\\
            \vdots & \vdots & \vdots & & \vdots\\
            w_\nolayers & 0 & 0 & \cdots & 0
        \end{pmatrix} \in \RR^{J \times (J + 1)} , \enskip \GM  = \begin{pmatrix}
            0 & 1 & 0 & \cdots & 0\\
            \vdots & \ddots& \ddots & & \vdots\\
            \vdots & & \ddots& 1 & 0\\
            0 & 0 & \cdots & 0 & 1
        \end{pmatrix} \in \RR^{J \times (J + 1)} \, ,
\end{align*}
$\GBias = \begin{pmatrix}
    \Gbias_1 & \Gbias_2 & \cdots & \Gbias_J
\end{pmatrix}^\top \in \RR^{\nolayers}$, $\GBiasd = \boldsymbol{0}$, $\actfct = \begin{pmatrix}
    \sigma_1 & \sigma_2 & \cdots & \sigma_J
\end{pmatrix}^\top$, $\GV = I_{J \times J}$, and $\GK = \begin{pmatrix}
    0 & c_1 & c_2 & \ldots & c_J
\end{pmatrix} \in \RR^{1 \times (J + 1)}$.

\subsection{Multi-layer Perceptrons (MLP)}\label{sec:mlp}

An MLP is a natural extension of the single-layer perceptron and the shallow neural network, consisting of an input layer, an output layer, and one or more hidden layers in between \cite{goodfellow2016}. The key characteristic of an MLP is that information flows strictly forward from one layer to the next, without cycles. This architecture allows the network to learn hierarchical features; each successive layer learns more abstract and complex representations based on the output of the previous layer. MLPs are neural networks $\net : \RR^M \to \RR^N$ of the form
\begin{align}
\begin{split}
    \net(\obs) &= K\aux_J + d \, ,\\
    \aux_1 &= \sigma_1(\Glin_1 \obs + \Gbias_1) \, ,\\
    \aux_j &= \sigma_j(\Glin_j \aux_{j - 1} + \Gbias_j) \quad \text{for all} \quad j \in \{2, \ldots, J\} \, ,
\end{split}\label{eq:mlp}
\end{align}
for weight matrices $\Glin_j \in \RR^{N_j \times N_{j - 1}}$ with $N_0 = M$, bias vectors $\Gbias_j \in \RR^{N_j}$ and $K \in \RR^{N \times N_J}$ as well as $d \in \RR^N$, where $J$ denotes the number of hidden layers. 
We can define $\Gvar = \begin{pmatrix}
    \obs & \aux_1 & \cdots & \aux_J
\end{pmatrix}^\top \in \RR^{\overline{N}}$, where $\overline{N} = \sum_{j = 0}^J N_j$. Then \eqref{eq:mlp} is of the form \eqref{eq:nn} for
\begin{align*}
    \GLin = \begin{pmatrix}
        \Glin_1 & 0_{N_1 \times N_1} & 0_{N_1 \times N_2} & \cdots & 0_{N_1 \times N_{{J} - 1}} & 0_{N_1 \times N_J}\\
        0_{N_2 \times M} & \Glin_2 & 0_{N_2 \times N_2} & \cdots & 0_{N_2 \times N_{{J} - 1}} & 0_{N_2 \times N_J}\\
        \vdots & & \ddots & & \vdots & \vdots\\
        0_{N_J \times M} & \cdots & & & \Glin_J & 0_{N_J \times N_J}
    \end{pmatrix} \in \RR^{(\overline{N} - M) \times \overline{N}} \, ,
\end{align*}
$\GBias = \begin{pmatrix}
    \Gbias_1 & \Gbias_2 & \cdots & \Gbias_J
\end{pmatrix}^\top \in \RR^{\overline{N} - M}$, $\mathbf{\sigma} = \begin{pmatrix}
    \sigma_1 & \sigma_2 & \cdots & \sigma_J
\end{pmatrix}^\top$, $\GV = I_{(\overline{N} - M) \times (\overline{N} - M)}$,
\begin{align*}
    \GM = \begin{pmatrix}
        0_{N_1 \times M} & I_{N_1 \times N_1} & 0_{N_1 \times N_2} & 0_{N_1 \times N_3} & \cdots & 0_{N_1 \times N_J}\\
        0_{N_2 \times M} & 0_{N_2 \times N_1} & I_{N_2 \times N_2} & 0_{N_2 \times N_3} & \cdots & 0_{N_2 \times N_J}\\
        \vdots & \vdots & & \ddots & & \vdots\\
        0_{N_J \times M} & 0_{N_J \times N_1} & \cdots & & & I_{N_J \times N_J}
    \end{pmatrix} \in \RR^{(\overline{N} - M) \times \overline{N}} \, ,
\end{align*}
and
\begin{align*}
    \GK = \begin{pmatrix}
        0_{N \times M} & 0_{N \times {N_1}} & \cdots & 0_{N \times {N}_{{J} - 1}} & K
    \end{pmatrix} \in \RR^{{N} \times \overline{N}} \, ,
\end{align*}
as well as $\GBiasd = d \in \RR^N$. This demonstrates how a standard multi-layer perceptron, with its sequential layer-wise structure, can be expressed within the proposed unified framework.

\subsection{Residual Networks}
Residual Networks, or ResNets, were introduced by He et al. in 2015 in \cite{he2016deep}. Their development was motivated by the observation that simply stacking more layers in a conventional deep neural network often led to a degradation and subsequent saturation of the training accuracy. This was not caused by overfitting, but rather by the difficulty of optimising very deep networks, a challenge closely related to the vanishing gradient problem \cite{he2016deep,glorot2010understanding}.

The core innovation of ResNets is the ``skip connection'', which allows the network to learn residual functions. Instead of forcing a network to learn an underlying mapping $\Meas(\var)$, the network is reformulated to learn a residual mapping $F(\var) \defeq \Meas(\var) - \var$. The original mapping is then recast as $F(\var) + \var$. This is implemented by adding the input of a block, $\var$, directly to its output. The insight of \cite{he2016deep} is that in practice it often is easier to learn networks that push the residual $F(\var)$ to zero than it is to approximate an identity mapping $H(\var) = \var$.

Interestingly, the iterative structure of ResNets can be interpreted as a forward Euler discretisation of an Ordinary Differential Equation (ODE) \cite{e2017dynamical,haber2017stable}. This perspective connects deep residual learning to the field of dynamical systems and has led to further architectural developments, such as Neural ODEs (NODE)~\cite{chen2018neural} or Runge--Kutta inspired ODENets~\cite{benning2019}. In this work, we consider residual neural networks $\net:\RR^M \rightarrow \RR^N$ of the form
\begin{align}
    \begin{split}
        \net(\obs) &= K\aux_J + d \, ,\\
        \aux_1 &= \obs + h_1 V_1 \sigma_1(\Glin_1 \obs + \Gbias_1) \, ,\\
        \aux_{j} &= \aux_{j - 1} + h_j V_j \sigma_j(\Glin_j \aux_{j - 1} + \Gbias_j) \quad \text{for all} \quad j \in \{2, \ldots, \nolayers\} \, ,
    \end{split}\label{eq:resnet}
\end{align}
for weight matrices $\Glin_j \in \RR^{N_j \times M}$, $K \in \RR^{N \times M}$, $V_j \in \RR^{M \times N_j}$, and bias vectors $\Gbias_j \in \RR^{N_j}$ as well as $d \in \RR^N$. Here, the scalars $h_j > 0$ are step-size parameters arising from the discretisation of an ODE (cf.~\cite{benning2019}). For simplicity, this step-size parameter could also be absorbed into each individual $V_j$.

Similar to the previous examples, we set $\Gvar = \begin{pmatrix}
    \obs & \aux_1 & \aux_2 & \cdots & \aux_J
\end{pmatrix}^\top \in \RR^{(J + 1) M}$. Then \eqref{eq:resnet} is of the form \eqref{eq:nn} for
\begin{align*}
    \GLin = \begin{pmatrix}
        \Glin_1 & 0_{N_1 \times M} & 0_{N_1 \times M} & \cdots & 0_{N_1 \times M} & 0_{N_1 \times M}\\
        0_{N_2 \times M} & \Glin_2 & 0_{N_2 \times M} & \cdots & 0_{N_2 \times M} & 0_{N_2 \times M}\\
        \vdots & & \ddots & & \vdots & \vdots\\
        0_{N_J \times M} & \cdots & & & \Glin_J & 0_{N_J \times M}
    \end{pmatrix} \in \RR^{\overline{N} \times M (J + 1)} \, ,
\end{align*}
for $\overline{N} = \sum_{j = 1}^J N_j$, $\GBias = \begin{pmatrix}
    \Gbias_1 & \Gbias_2 & \cdots & \Gbias_J
\end{pmatrix}^\top \in \RR^{\overline{N} - M}$, $\mathbf{\sigma} = \begin{pmatrix}
    \sigma_1 & \sigma_2 & \cdots & \sigma_J
\end{pmatrix}^\top$,
\begin{align*}
    \GM = \begin{pmatrix}
        -I_{M \times M} & I_{M \times {M}} & 0_{M \times M} & 0_{M \times M} & \cdots & 0_{M \times M}\\
        0_{M \times M} & -I_{M \times M} & I_{M \times M} & 0_{M \times M} & \cdots & 0_{M \times M}\\
        \vdots & \vdots & & \ddots & & \\
        0_{M \times M} & 0_{M \times M} & \cdots & & -I_{M \times M} & I_{M \times M}
    \end{pmatrix} \in \RR^{M{J} \times M{(J+1)}} \, ,
\end{align*}
the matrix
\begin{align*}
    \GV = \begin{pmatrix}
        h_1 V_1 & 0_{M \times N_1} & 0_{M \times N_1} & \cdots & 0_{M \times N_1} \\
        0_{M \times N_2} & h_2 V_2 & 0_{M \times N_2} & \cdots & 0_{M \times N_2} \\
        \vdots & & \ddots &  &  \\
        \vdots & &  & \ddots &  \\
        0_{M \times N_J} & \cdots & \cdots &  0_{M \times N_J} & h_J V_J
    \end{pmatrix} \in \RR^{MJ \times (\overline{N} - M)} \, ,
\end{align*}
and
\begin{align*}
    \GK = \begin{pmatrix}
        0_{N \times M} & 0_{N \times M} & \cdots & 0_{N \times M} & K
    \end{pmatrix} \in \RR^{N \times (J + 1)M} \, ,
\end{align*}
as well as $\GBiasd = d \in \RR^N$. This formulation shows how the additive nature of residual connections, which defines the ResNet architecture, can be captured within the proposed unified framework and justifies the need for the $\GM$ and $\GV$ to capture a wide range of architectures. Note that the analogy to ODEs enables different discretisations as suggested in \cite{benning2019}, which will lead to different designs for the involved block operators.

\subsection{Proximal Denoising Networks}

In the last decade, a new category of neural networks has appeared, whose architectures are inspired by optimisation algorithms. These networks are often referred to as \textit{unfolded networks}, \textit{unrolled methods} or \textit{proximal networks} \cite{gregor2010learning, adler2017solving, adler2018learned, jiu2021deep, monga2021algorithm, le2022faster, le2023pnn, mardani2018neural, wang2025unrolling}. They are designed by unrolling a fixed number of iterations of an optimisation algorithm built to solve problems of the form of~\eqref{eq:variational_reg}.
They have shown to reconstruct images with similar quality compared to traditional networks for performing certain imaging tasks (e.g., image restoration, image denoising, edge detection, etc.), while relying on architectures with fewer parameters. 

Since these networks are reminiscent of optimisation algorithms, a large variety of architectures can be derived. In this section, we focus on two specific architectures that are based on the philosophy of solving least-squares problems with $\ell_1$ regularisation. In particular, the networks provided below are obtained by developing proximal algorithms for solving an inverse problem (see, e.g.,~\cite{combettes2011proximal} for an introduction on proximal methods). In particular, we assume that we have some measurements $\obs \in \RR^M$ obtained as an observation of an unknown signal through a measurement operator $\Meas\colon \RR^N \to \RR^M$, then the objective is to find an estimate of the original signal.

\smallskip\noindent\textbf{Synthesis formulation with LISTA -- }
The \emph{Learned Iterative Soft-Thresholding Algorithm} (\emph{LISTA}) has been introduced in \cite{gregor2010learning} with the objective to
\begin{equation}\label{pb:pnn:synthesis}
\text{find }
\var_\lambda = \Lin \dual_\lambda
\quad \text{such that}\quad
\dual_\lambda = \argmind{\dual\in \RR^\codedim} \, \frac12 \| \Meas \Lin \dual - \obs \|^2 + \lambda \| \dual \|_1 ,
\end{equation}
where $L \colon \RR^\codedim \to \RR^N$ is a dictionary, and $\lambda>0$ a regularisation parameter.
In this context, $\Meas \colon \RR^N \to \RR^M$ and $\obs \in \RR^M$ are the measurement operator and observation related to the inverse problem~\eqref{eq:inverse_problem}.
Problem~\eqref{pb:pnn:synthesis} is known as the synthesis problem~\cite{elad2007analysis}, where $\dual$ are the sparse codes, and the signal can be synthesised as $\var = \Lin \dual$.

Problem~\cref{pb:pnn:synthesis} can be solved with the forward-backward algorithm~\cite{combettes2005signal, lions1979splitting} (also known as \emph{Iterative Soft-Thresholding Algorithm} (\emph{ISTA})), given by
\begin{equation}\label{algo:ista}
(\forall j \in \mathbb{N})\quad
\dual_{j+1} = \soft{\gamma\lambda} \Big( \underset{\eqdef W}{\underbrace{(I - \gamma \Lin^* \Meas^* \Meas \Lin)}} u_j + \underset{\eqdef b}{\gamma \underbrace{\Lin^*\Meas^* \obs}} \Big)
\end{equation}
where $\soft{\gamma\lambda}$ is the soft-thresholding operator (proximity operator of the $\ell_1$-norm) defined as
\[
\soft{\gamma\lambda}(z) = \operatorname{sign}(z)\max(|z|-\gamma\lambda,0).
\]
The operations are applied component-wise when $z$ is vector-valued. Here, $\gamma\lambda > 0$ is the thresholding parameter and $\gamma>0$ is a step-size parameter that satisfies $\gamma < 1/\| \Meas \Lin \|^2$, where $\| \cdot \|$ denotes the operator norm.

The signal is recovered at convergence by defining $x_\lambda = \Lin \dual_\lambda$, where $\dual_\lambda$ is the limit point of the sequence $(u_j)_{j\in \mathbb{N}}$.
In~\eqref{algo:ista}, $W$ constitutes a structured weight matrix and $b$ corresponds to a structured bias term.

LISTA constructs a neural network architecture by unrolling algorithm~\cref{algo:ista} over a fixed number of iterations $\nolayers\in \mathbb{N}$, where the learnable parameters can include the sparsity basis $\Lin$, the regularisation parameter $\lambda$, and the step-size $\gamma$. More generally, one can consider having varying parameters $(\Lin_{j})_{1 \le j \le \nolayers}$ and $(\lambda_j, \gamma_j)_{1 \le j \le \nolayers-1}$ over the $\nolayers$ iterates/layers.

Reformulating the LISTA network architecture with respect to the unified framework, we obtain the unrolled network $\net:\RR^M \to \RR^N$ of the form
\begin{equation}
\begin{split}
    \net(\obs) &= \Lin_{\nolayers} \dual_{\nolayers-1}  \, ,\\
    \dual_1 &= \Lin_1^* \Meas^* y\, ,\\
    \dual_j &= \soft{\gamma\lambda} \left( (I_{\codedim} - \gamma \Lin_j^* \Meas^* \Meas \Lin_j)\dual_{j-1} + \gamma \Lin_j^* \Meas^* \obs \right)
\end{split} \label{eq:lista}
\end{equation}
for $j = 2, \ldots, \nolayers-1$.
If we define
$$\Gvar = \begin{pmatrix}
    \obs \\ \dual_1  \\ \vdots \\ \dual_{\nolayers-1}
\end{pmatrix} \in \RR^{M+(\nolayers-1)\codedim},$$
then~\eqref{eq:lista} fits the unified framework~\cref{eq:nn} with $\GBiasd = \boldsymbol{0}$, $\GV = I_{(\nolayers-1)\codedim \times (\nolayers-1)\codedim}$,
\begin{equation*}
\GLin =
\begin{pmatrix}
    \Lin_1^* \Meas^* & 0_{\codedim \times \codedim} & 0_{\codedim \times \codedim} & \cdots & 0_{\codedim \times \codedim} & 0_{\codedim \times \codedim}\\
    0_{\codedim \times M} & \Glin_{2} & 0_{\codedim \times \codedim} & \cdots & 0_{\codedim\times \codedim} & 0_{\codedim \times \codedim}\\
    \vdots & & \ddots & & \vdots & \vdots\\
    0_{\codedim \times M} & \cdots & & & \Glin_{{\nolayers-1}} & 0_{\codedim \times \codedim}
\end{pmatrix} \in \RR^{(\nolayers-1)\codedim \times (M+ (\nolayers-1)\codedim )} \, ,
\end{equation*}
where $\Glin_{j} = I_{\codedim} - \gamma_{j} \Lin_j^*H^*H\Lin_j$,
and
\begin{equation*}
\GBias =\begin{pmatrix}
0_{\codedim} \\ \Gbias_2 \\ \vdots \\ \Gbias_{\nolayers-1}
\end{pmatrix} \in \RR^{(\nolayers-1) \codedim}
\end{equation*}
where $\Gbias_j = \gamma \Lin_j^*H^* y$ for all $j \in \{2, \ldots, \nolayers-1\}$.
The activation functions are given by $\mathbf{\sigma} = \begin{pmatrix} \operatorname{Id} & \sigma_2 & \cdots & \sigma_{\nolayers-1} \end{pmatrix}$ with
\begin{equation*}
\sigma_j(z)= \soft{\gamma_j \lambda}(z)=\text{sign}(z)\max(|z|-\gamma_j\lambda,0)
\end{equation*}

\noindent for all $j \in \{2, \ldots, \nolayers-1\}$. The middle layer relation is given by
\begin{equation*}
    \GM = \begin{pmatrix}
        0_{\codedim \times M} & I_{\codedim} & 0_{\codedim \times \codedim} & 0_{\codedim \times \codedim} & \cdots & 0_{\codedim \times \codedim}\\
        0_{\codedim \times M} & 0_{\codedim \times \codedim} & I_{\codedim} & 0_{\codedim \times \codedim} & \cdots & 0_{\codedim \times \codedim}\\
        \vdots & \vdots & & \ddots & & \vdots\\
        0_{\codedim \times M} & 0_{\codedim \times \codedim} & \cdots & & & I_{\codedim}
    \end{pmatrix} \in \RR^{(\nolayers-1)\codedim \times (M+(\nolayers-1)\codedim)} \, ,
\end{equation*}

\noindent and the output mapping is defined by
\begin{equation*}
\GK = \begin{pmatrix}
    0_{N \times M} &  0_{N \times \codedim} & \dots & 0_{N \times \codedim} & \Lin_{\nolayers}
    \end{pmatrix} \in \RR^{N \times (M+(\nolayers-1)\codedim) }.
\end{equation*}

\smallskip\noindent\textbf{Analysis formulations with primal-dual algorithms --} Another approach focuses on analysing the minimisation problem
\begin{equation}\label{pb:pnn:analysis}
    \minimize{x\in C} \frac12 \| \Meas \var - \obs \|^2 + \lambda \|L \var \|_1,
\end{equation}

\noindent where $C \subset \RR^N$ represents a convex set for which it is easy to project onto (e.g., $C = [0,+\infty)$ or $C = [0,1]$). Then, problem~\cref{pb:pnn:analysis} can be solved with the Condat-V\~u primal-dual algorithm~\cite{condat2013primal, vu2013splitting} that reads
\begin{equation}\label{algo:pdcv}
(\forall j \in \NN)\quad
\begin{cases}
    \dual_{j+1} = \text{proj}_{[-\lambda,+\lambda]^{\codedim}} \big( \dual_j + \gamma \Lin \var_j \big) \\
    \var_{j+1} = \text{proj}_{C} \Big( \widetilde{\Meas}_{\tau} x_j + \tau \Meas^* \obs - \tau \Lin^* (2 \dual_{j+1} - u_j) \Big).
\end{cases}
\end{equation}
where $\gamma,\tau>0$ are step-sizes and $\widetilde{\Meas}_{\tau} = (I_{N} - \tau \Meas^*\Meas)$. As suggested in \cite{jiu2021deep}, this iterative scheme can be used to construct a neural network architecture by unrolling~\cref{algo:pdcv} over a fixed number of iterations $\nolayers \in \mathbb{N}$. Learnable components include the sparsity basis $\Lin$, the regularisation parameter $\lambda$, and the step-sizes $\gamma$ and $\tau$.
Similar to the LISTA network, one can have varying parameters $\{\Lin_j, \lambda_j, \gamma_j, \tau_j\}_{0 \le j \le J-1}$ over the $J$ iterations. Reformulating the network architecture with respect to the unified framework, we obtain
\begin{align*}
\net(\obs) &= x_{\nolayers} \\
\dual_0
    &=  0_S \\
\var_0
    &= \Meas^* \obs \\
\dual_{1}
    &= \text{proj}_{[-\lambda , +\lambda]^{\codedim}} \big( \gamma_1 \Lin_1 \var_0 \big) \\
\dual_{j}
    &= \text{proj}_{[-\lambda , +\lambda]^{\codedim}} \big( \dual_{j-1}
    + \gamma_{j} \Lin_j x_{j-1} \big)
    \quad   \text{for } j \in \{2, \ldots, \nolayers\} \\
\var_{j}
    &= \text{proj}_{C}
    \Big( \widetilde{\Meas}_{\tau_{j}} \var_{j-1}
    - \tau_{j} \Lin_j^* (2 \dual_{j} - \dual_{j-1}) + \tau_{j} \Meas^* \obs \Big)
    \quad\text{for } j \in \{1, \ldots, \nolayers\}.
\end{align*}
This network also fits within our unified formulation~\eqref{eq:nn} with
\begin{equation*}
\Gvar = \begin{pmatrix}
    \obs \\ 
    \var_0 \\ \dual_1 \\ \var_1 \\  \vdots \\ \dual_{\nolayers} \\ \var_{\nolayers}
\end{pmatrix} \in R^{M + \nolayers (\codedim + N)} \,,
\end{equation*}
defining $\GLin\in \RR^{(N + \nolayers (\codedim+N)) \times (M+ N + \nolayers(\codedim+N) )}$ as
\begin{equation*}
\GLin =
\begin{pmatrix}
    \Meas^* 
    & 0_{\codedim\times N}
    & 0_{\codedim\times \codedim}
    & 0_{N\times N}
    &  0_{N \times \codedim}
    & \dots & \dots & \dots
    & 0_{\codedim\times \codedim}
    & 0_{\codedim\times N} \\
    0_{\codedim\times M}
    &  \gamma_1 \Lin_1
    & 0_{\codedim\times \codedim}
    & 0_{\codedim\times N}
    &  0_{\codedim \times \codedim}
    & \dots & \dots & \dots
    & 0_{\codedim\times \codedim}
    & 0_{\codedim\times N} \\
    0_{N \times M}
    & \widetilde{\Meas}_{\tau_{1}}
    & - \tau_1 \Lin_1^*
    & 0_{N\times N}
    &  0_{N \times \codedim}
    & \dots & \dots & \dots
    & 0_{N\times \codedim}
    & 0_{N\times N} \\
    0_{\codedim\times M}
    &  0_{\codedim\times N}
    & I_{\codedim}
    & \gamma_2 \Lin_2
    &  0_{\codedim \times \codedim}
    & \dots & \dots & \dots
    & 0_{\codedim\times \codedim}
    & 0_{\codedim\times N} \\
    0_{N \times M}
    & 0_{N\times N}
    & \tau_2 \Lin_2^*
    & \widetilde{\Meas}_{\tau_2}
    &  - 2 \tau_2 \Lin_2^*
    & \dots & \dots& \dots
    & 0_{N\times \codedim}
    & 0_{N\times N} \\
    \vdots & \vdots & \vdots & \ddots & \ddots & \ddots & \ddots & \ddots & \vdots& \vdots \\
    0_{\codedim \times M}
    & 0_{\codedim \times N}
    & 0_{\codedim \times \codedim}
    & \dots & \dots & \dots & \dots
    & I_{\codedim}
    & \gamma_J L_J
    & 0_{N\times N}\\
    0_{N \times M}
    & 0_{N \times N}
    & 0_{N \times \codedim}
    & \dots & \dots & \dots& \dots
    & \tau_J L_J^*
    & \widetilde{\Meas}_{\tau_J}
    &  - 2 \tau_J \Lin_J^*
\end{pmatrix},
\end{equation*}
and
\begin{equation*}
    \GBias =\begin{pmatrix}
    0_N \\ 0_{\codedim} \\ \tau_1 \Meas^* \obs \\ 0_{\codedim} \\ \tau_2 \Meas^* \obs  \\ \vdots \\ 0_{\codedim} \\ \tau_J \Meas^* \obs \end{pmatrix} \in \RR^{N + \nolayers (\codedim+N)}.
\end{equation*}
The activation function is now given by alternating between the two projections:
$\mathbf{\sigma} = \begin{pmatrix} I_{N} & \text{proj}_{[-\lambda,+\lambda]^{\codedim}} & \text{proj}_{C} & \text{proj}_{[-\lambda,+\lambda]^{\codedim}} & \cdots & \text{proj}_{C} \end{pmatrix}$.
The middle layers relation~\eqref{eq:nnpred_const1} is given by
$\GM = \begin{pmatrix} 0_{(N + \nolayers(\codedim+N)) \times M} & I_{N + \nolayers(\codedim+N)} \end{pmatrix}$ and $\GV = I_{\nolayers(\codedim+N) \times \nolayers(\codedim+N)}$.
Finally, the output mapping \eqref{eq:nnpred} is given by $\GK=\begin{pmatrix}
0_{N \times (M + N + \nolayers(\codedim+N))}  & I_{N}
\end{pmatrix} $ and $\GBiasd = \boldsymbol{0}$.

\section{Distributed optimisation of neural networks}\label{sec:mac}

The immense computational and memory demands of training modern deep neural networks, driven by ever-increasing model and dataset sizes \cite{langer2020distributed, ben2019demystifying}, have made distributed optimisation an indispensable component of deep learning. A rich and diverse field of strategies has emerged to parallelise the training process across multiple computing devices. While some paradigms, such as federated learning, focus on training across siloed, private data on edge devices~\cite{li2020federated}, the strategies used in large-scale scientific computing can be broadly categorised into two main philosophies:
\begin{itemize}
\item The first, and most common, is system-level parallelism, which focuses on distributing the computational workload of a standard training algorithm, typically \emph{stochastic gradient descent} (SGD). This category includes well-established techniques such as ($i$) data parallelism, where the model is replicated and the data is partitioned~\cite{sergeev2018horovod, li2014scaling}; ($ii$) model parallelism, where the model itself is partitioned across devices, either between layers (pipeline parallelism)~\cite{huang2019gpipe, narayanan2019pipedream} or within layers (tensor parallelism)~\cite{shoeybi2019megatron}; and ($iii$) sophisticated hybrid approaches that partition model states, gradients, and optimiser states to enable training at unprecedented scales~\cite{rajbhandari2020zero}. These methods primarily address system-level bottlenecks of computation, memory, and communication, without altering the underlying mathematical formulation of the nested objective function.
\item The second, alternative philosophy is algorithmic parallelism, which reformulates the optimisation problem itself to expose a parallel structure. Instead of optimising a single, deeply nested function, this approach transforms it into a relaxed problem over a set of simpler, decoupled functions that can be solved more easily in a distributed manner. In the following subsection, we will explore a prime example of this algorithmic approach---MAC-QP~\cite{carreira2014distributed}. Furthermore we demonstrate how it provides a powerful mechanism for decoupling the complex dependencies inherent in deep network architectures and how it fits into the framework \cref{eq:loss}, for a wide range of network architectures as those presented in \Cref{sec:unified}.
\end{itemize}

\subsection{MAC Reformulated in the Unified Framework}\label{sec:distributed_mac}

The MAC-QP, first proposed for training multi-layer perceptrons~\cite{carreira2014distributed}, is a perfect example of the algorithmic parallelism philosophy. Instead of directly minimising the constrained objective function \cref{eq:constrained_loss}, MAC-QP reformulates the problem by introducing auxiliary variables for the output of each layer and subsequent penalisation of constraints with quadratic penalties. This transforms the original nested problem into a higher-dimensional but more loosely coupled optimisation problem.

This formulation is a special case of the general loss function introduced in our unified framework in \cref{eq:loss}. Similarly to conventional training, the architectural constraint $\GM\Gvar_i = \GV\Gaux_i$ is enforced strictly, but the non-linear activation constraint is relaxed. In particular, we choose
\begin{align*}
    \mathcal{C}(\ab, \ab') = \chi_{\{\boldsymbol{0}\}}(\ab - \ab') \quad \text{and} \quad
    \mathcal{D}_{\mathbf{\sigma}}(\ub, \vb) = \frac{\mu}{2} \|\ub - \mathbf{\sigma}(\vb)\|^2_2 \, ,
\end{align*}

\noindent where $\chi_{\{\boldsymbol{0}\}}$ is the characteristic function over the zero-equality constraint (as defined in \Cref{rem:conventinal-tranining}), and $\mu > 0$ is a scalar penalty parameter (note that in~\cite{carreira2014distributed} we could choose different $\mu$'s for each layer, which would correspond to a suitable diagonally weighted Euclidean norm here, but for simplicity we focus on the scalar case).

Substituting these into the general loss \cref{eq:loss} yields the optimisation problem
\begin{align}\label{train:macqp-gen}
\minimize{\{\Gvar_i, \Gaux_i\}_i, \theta} \quad &\mathbb{E}_{i}\left[ \loss(\GK\Gvar_i + \GBiasd) + \frac{\mu}{2} \|\Gaux_i - \actfct(\GLin\Gvar_i + \GBias)\|^2_2 \right] \nonumber \\
\text{subject to} \quad & \GM\Gvar_i = \GV\Gaux_i \, ,
\end{align}
where $i \in \mathbb{I}$, and $\theta$ denotes the collection of learnable parameters within the operators $\GK$, $\GM$, $\GV$, $\GLin$ and biases $\GBias$ and $\GBiasd$. In the following example, we consider this setting in the context of an MLP architecture defined in \cref{eq:mlp}.

\begin{example}[MAQ-QP for MLP]
The operators involved in an MLP are defined in Setion~\ref{sec:mlp}. In particular, operator $\GV$ is the identity matrix. The resulting hard constraint $\GM\Gvar_i = \Gaux_i$ in \eqref{train:macqp-gen} allows us to substitute $\Gaux_i$ with $\GM\Gvar_i$ in the loss function above. The MAC-QP loss for training an MLP can therefore be expressed solely in terms of the variables $\Gvar_i$ and the network parameters $\theta$, i.e.,
\begin{equation}
\uniloss = \mathbb{E}_{i}\left[ \loss(\GK\Gvar_i + \GBiasd) + \frac{\mu}{2} \|\GM\Gvar_i - \actfct(\GLin\Gvar_i + \GBias)\|^2_2 \right] \, . \label{eq:mac_loss}
\end{equation}
Note that given the definitions of $\GM$, $\GK$ and $\GBiasd$ for an MLP, this is equivalent to
\begin{align*}
\uniloss = \mathbb{E}_{i}\left[ \loss(K (\aux_{i})_{ \nolayers} + d) + \frac{\mu}{2} \sum_{j=1}^{\nolayers} \|(\aux_{i})_{j} - \sigma(\Glin_j (\aux_{i})_{j - 1} + \Gbias_j)\|^2_2 \right] \, .
\end{align*}
Here $(\aux_{i})_{j}$ denotes the auxiliary variable $\aux_{j}$ associated with the $i$-th data sample, and $(\aux_{i})_{0} = \obs_i$ for all $i$ (see \Cref{sec:mlp}).

For quadratic loss functions $\mathcal{L}$ and zero bias terms, this is precisely the relaxed learning objective proposed in \cite{carreira2014distributed}, now expressed in our unified notation. The key advantage of this formulation is that the optimisation over the auxiliary variables $\{\Gvar_i\}_i$ decouples layer-wise meaning that we have at most two terms in the objective depend on an individual $(\aux_{i})_{j}$. Furthermore, for fixed $\{\Gvar_i\}_i$, the optimisation over the parameters in $\GLin, \GBias, \GK, \GBiasd$ also decouples and decomposes into smaller, independent problems for each layer, enabling efficient parallelisation. 

\remarkend
\end{example}

While we have used the MLP as the primary example, it is important to note that the MAC-QP training principle can easily be extended to the other architectures described in \Cref{sec:unified} and it motivates the development of the lifted training framework which we introduce in the following section.

\section{Lifted training of neural networks}\label{sec:lifted_training}

Lifted training of neural networks represents a paradigm shift from conventional end-to-end training via back-propagation. The core idea is to reformulate the highly non-convex and deeply nested optimisation problem by ``lifting'' it into a higher-dimensional space~\cite{askari2018lifted, carreira2014distributed}. This is achieved by introducing auxiliary variables that explicitly represent the activations of each layer, thereby breaking the sequential dependencies inherent in the network's forward pass~\cite{askari2018lifted, wang2023liftedtraining}. The original hard constraints defining the network architecture are then relaxed and replaced with penalty terms in the objective function. This transformation yields a larger but structured optimisation problem that is amenable to block optimisation techniques like block-coordinate descent~\cite{askari2018lifted, gu2020fenchel}.

This approach is built upon the MAC-QP approach introduced in \Cref{sec:mac}, which uses lifting in combination with simple quadratic penalties to enable parallel training and was found to produce good weight initialisations for standard network training~\cite{negiar2017optml}. The field has since evolved to include more principled formulations based on Fenchel duality~\cite{gu2020fenchel}, Bregman distances~\cite{wang2023liftedtraining}, and contrastive objectives~\cite{zach2019contrastive} designed to overcome performance limitations of earlier methods. Alternative lifting approaches have explored lifting as a mechanism to build adversarially robust models by design \cite{hoier2020lifted} or as a novel type of network layer inspired by convex optimisation \cite{ochs2018lifting}.

\subsection{The Classical Lifted Approach in the Unified Framework}

The classical lifted approach, as described in \cite{askari2018lifted}, offers a distinct alternative to the MAC-QP formulation. Its core idea stems from the observation that many activation functions, such as the ReLU, can be expressed as the solution to a simple convex optimisation problem. For instance,
\[\text{ReLU}(v) = \argmind{u \ge 0} \|u-v\|_2^2.\]
Rather than penalising the deviation of an auxiliary variable from the \textit{activated} output of an affine transformation, this method penalises the deviation from the \textit{pre-activation} affine output. The effect of the activation function is then enforced through a hard constraint on the auxiliary variable.

This formulation can be expressed in our unified framework \cref{eq:loss} by making a different choice for the penalty functions. The architectural constraint is still enforced strictly, but the non-linear activation constraint is handled by a combination of a quadratic penalty as in the MAC-QP framework and a regularisation function acting on the auxiliary variables. Specifically, we choose
\begin{align*}
    \mathcal{C}(\ab, \ab') = \chi_{\{\boldsymbol{0}\}}(\ab - \ab') \quad \text{and} \quad
    \mathcal{D}_{\mathbf{\sigma}}(\ub, \vb) = \mathcal{R}(\ub) + \frac{\mu}{2} \|\ub - \vb\|^2_2 \, ,
\end{align*}
\noindent where $\mathcal{R}$ denotes some regularisation function. For a ReLU activation, $\mathcal{R}$ is the characteristic function over the non-negative orthant, i.e., $\mathcal{R}(\ub) = \chi_{\geq \boldsymbol{0}}(\ub)$, which is zero if all elements of $\ub$ are non-negative, and $+\infty$ otherwise.
Below we show how this framework applies to the particular case of the MLP described in Section~\ref{sec:mlp}.

\begin{example}[Classical Lifted for MLP]

We consider an
MLP where $\GV$ is given by an identity operator, we can again replace $\Gaux_i$ with $\GM\Gvar_i$, and the optimisation problem becomes
\begin{align*}
\minimize{\{\Gvar_i\}_i, \theta} \quad &\mathbb{E}_{i}\left[ \loss(\GK\Gvar_i + \GBiasd) + \frac{\mu}{2} \|\GM\Gvar_i - (\GLin\Gvar_i + \GBias)\|^2_2 + \mathcal{R}(\GM \Gvar_i) \right] \, ,
\intertext{respectively,}
\minimize{\{\Gvar_i\}_i, \theta} \quad &\mathbb{E}_{i}\left[ \loss(\GK\Gvar_i + \GBiasd) + \frac{\mu}{2} \|\GM\Gvar_i - (\GLin\Gvar_i + \GBias)\|^2_2 \right] \\
\text{subject to} \quad & (\GM\Gvar_i)_j \succeq 0 \quad \text{for all}\,\, i, j \,,
\end{align*}
with $\mathcal{R}(\ub) = \chi_{\geq \boldsymbol{0}}(\ub)$. For an MLP, the latter is equivalent to
\begin{align*}
\minimize{\{\Gvar_i\}_i, \theta} \quad &\mathbb{E}_{i}\left[ \loss(K (\aux_{i})_{\nolayers} + d) + \frac{\mu}{2} \sum_{j=1}^{\nolayers} \|(\aux_{i})_{j} - (\Glin_j (\aux_{i})_{j - 1} + \Gbias_j)\|^2_2 \right] \, , \\
\text{subject to} \quad &(\aux_{i})_{j} \succeq 0 \quad \text{for all} \,\,j \in \{1, \ldots, \nolayers\} \,\,\text{and all}\,\, i,
\end{align*}
where, if $(\aux_{i})_{j}$is vector-valued, the inequality is to be understood point-wise for each entry of the vector.
\remarkend
\end{example}
However, this classical formulation has a significant drawback; as pointed out in~\cite{wang2023liftedtraining,zach2019contrastive} the optimality conditions reveal a critical flaw: The optimality condition with respect to the parameters $\theta$ requires the gradient of the penalty term to be zero. This implies that at convergence we must have $\GM\Gvar_i = \GLin\Gvar_i + \GBias$, which effectively means that we learn an affine-linear neural network and not a nonlinear function approximator.

Despite this limitation, the approach has proven valuable as a pre-training or weight initialisation scheme, providing a good starting point for subsequent fine-tuning with standard back-propagation \cite{negiar2017optml,zach2019contrastive}.

\subsection{Fenchel Lifted Networks Reformulated in the Unified Framework}\label{sec:fenchel}

Fenchel Lifted Networks \cite{gu2020fenchel} advance the lifted paradigm by replacing the simple quadratic penalties of MAC-QP \cite{carreira2014distributed} or the original lifted network training approach \cite{askari2018lifted} with a more principled formulation derived from convex analysis. Instead of relaxing the activation constraint with a quadratic ad-hoc penalty, this approach uses Fenchel duality to represent the non-linear activation $\Gaux = \boldsymbol{\sigma}(\vb)$ as an equivalent biconvex constraint of the form $B_{\boldsymbol{\sigma}}(\Gaux, \vb) \le 0$~\cite{gu2020fenchel}. 

Within our unified framework, this relaxation can be interpreted as a specific choice for the penalty function $\mathcal{D}_{\boldsymbol{\sigma}}$. The architectural constraint is again enforced strictly, while the non-linear activation constraint is penalised using the biconvex function $B_{\boldsymbol{\sigma}}$ weighted by a hyperparameter $\mu > 0$, i.e.,
\begin{align*}
\mathcal{C}(\ab, \ab') = \chi_{\{\mathbf{0}\}}(\ab - \ab') \quad \text{and} \quad
\mathcal{D}_{\boldsymbol{\sigma}}(\ub, \vb) = \mu B_{\boldsymbol{\sigma}}(\ub, \vb) \,.
\end{align*}

\noindent The function $B_{\boldsymbol{\sigma}}$ is constructed via Fenchel duality to satisfy that $B_{\boldsymbol{\sigma}}(\ub, \vb) \le 0$ is equivalent to $\ub = \boldsymbol{\sigma}(\vb)$. Note that for the ReLU activation function, $B_{\boldsymbol{\sigma}}$ is then given by
\begin{align*}
    B_{\text{ReLU}}(\ub, \vb) &= \begin{cases} \frac12 \| \ub \|^2 + \frac12 \| \max(\vb, 0) \|^2 - \langle \vb,\ub \rangle &\text{if } \ub \succeq 0  \\ \infty & \text{otherwise} \end{cases} \, .
\end{align*}
\begin{example}[Fenchel Lifting for MLP]

Substituting these choices into the general loss \cref{eq:loss} and again considering an MLP where $\GV$ is given by the identity and $\Gaux_i = \GM\Gvar_i$ for all samples $i \in \mathbb{I}$, the training objective becomes
\begin{equation*}
\uniloss(\{\Gvar_i\}_i, \theta) = \mathbb{E}_{i}\left[ \loss(\GK\Gvar_i + \GBiasd) + \mu B_{\actfct}(\GM\Gvar_i, \GLin\Gvar_i + \GBias) \right] \, ,
\end{equation*}

\noindent and the optimisation problem can be expressed as
\begin{align*}
    \minimize{\{\Gvar_i\}_i, \theta} \quad &\mathbb{E}_{i}\left[ \loss(K (\aux_{i})_ {\nolayers} + d) + \mu \sum_{j=1}^{\nolayers} B_{\sigma_j}\left((\aux_{i})_{j}, \Glin_j (\aux_{i})_{j - 1} + \Gbias_j\right) \right] \, ,
\end{align*}

\end{example}

\noindent if we substitute the components of $\GK$, $\GM$, $\GV$, $\GLin$, $\GBias$, $\GBiasd$ and $\actfct$ as defined in \Cref{sec:mlp}. Here, $\theta$ again denotes the collection of all learnable parameters. This formulation retains the block-separable structure of MAC-QP and the original lifted formulation, making it highly amenable to parallelisation using block-coordinate descent~\cite{gu2020fenchel}.
The primary advantage of the Fenchel approach is that its block optimisation problems are convex and easier to solve. This is because, unlike formulations based on quadratic penalties~\cite{wang2023liftedtraining}, it does not require differentiating the activation functions, which is particularly beneficial for non-differentiable activations, which we discuss in  \Cref{sec:lifted_bregman_unified}.

\subsection{Lifted Contrastive Learning in the Unified Framework}

In~\cite{zach2019contrastive} the authors discuss limitations of the classical lifted formulation~\cite{askari2018lifted}, in particular its bias towards learning the linear regime of the activation function. To address this limitation, they introduce a contrastive training objective, defined by the difference between two energy functions. In this section, we follow the original formulation in~\cite{zach2019contrastive}
and consider MLP networks with ReLU nonlinearity, apart from the last layer where the activation $\actfct_J$ is taken as identity. Given a single input $\obs_i$ to the network, the feed-forward computations can be approximated by minimising the following convex objective:
\begin{align*}
    & E(\Gvar_i|\theta) \;= \;\frac{1}{2}\|(\aux_{i})_{1} - (\Glin_1 \obs_i + \Gbias_1)\|^2_2 + \frac{\mu_j}{2} \sum_{j=1}^{\nolayers} \|(\aux_{i})_{j} - (\Glin_j (u_{i})_{j - 1} + \Gbias_j)\|^2_2 \,, \\
    &\text{subject to} \quad (\aux_{i})_{j} \succeq 0 \quad \forall j \in \{1, \ldots, \nolayers\}, \forall i \in \mathbb{I}.
\end{align*}%
This objective function concerns only the auxiliary variables and is ``label'' free since it does not dependent on the ground-truth $\var_i$. At this stage, this energy is therefore not influenced by training targets and is referred to as the \textit{free energy}. In other words, for fixed $\theta$, the computations of hidden activations of the network via a forward pass can be determined as minimisers of $E(\Gvar_i|\theta)$. We call $\check{\Gvar}_i := \argmind{\Gvar_i} E(\Gvar_i|\theta)$ the free (energy) solution.
For a single training sample, the classical lifted training scheme can therefore be reformulated as a min-min problem, i.e. minimising an energy $\mathcal{E}_0(\theta)$ over network parameters $\theta$, which is itself a minimised energy with respect to $\Gvar_i$,
\begin{align*}
\mathcal{E}_0(\theta) \; := \; \; & \minimize{\Gvar_i} \quad  E(\Gvar_i|\theta) + \loss((\aux_{i})_{\nolayers}) \, , \\
& \text{subject to} \quad (\aux_{i})_{j} \succeq 0 \quad \text{for all} \,\,j \in \{1, \ldots, \nolayers\} \,.
\end{align*}
In particular, the authors consider $\loss$ to take the form of an indicator function $\chi_{\boldsymbol{0}}(\cdot)$. Under this assumption, the inner minimisation objective with respect to $\Gvar_i$ can be expressed as
\begin{multline*}
\widehat{E}_{x_i}(\Gvar_i|\theta)  =  \frac{1}{2}\|(\aux_{i})_{1} - (\Glin_1 \obs_i + \Gbias_1)\|^2_2  +
 \frac{\mu_j}{2} \sum_{j=1}^{\nolayers-1} \|(\aux_{i})_{j} - (\Glin_j (\aux_{i})_{j - 1} + \Gbias_j)\|^2_2  \\
+ \frac{\mu_J}{2} \|\var_i - (\Glin_{\nolayers} (\aux_{i})_{\nolayers - 1} + \Gbias_{\nolayers})\|^2_2 \,,
\end{multline*}
subject to hard constraints $(\aux_{i})_{j} \succeq 0$ for all $j$ in $\{1, \ldots, \nolayers-1\}$. Notice that this can be viewed as $E(\Gvar_i|\theta)$ with its last term now replaced by a quadratic loss function. This is due to the indicator function such that the network output $(\aux_{i})_{J}$ is clamped to the training target $\var_i$. For this reason, $\widehat{E}_{x_i}(\Gvar_i|\theta)$ is referred to as the \textit{clamped} energy.
We call $\hat{\Gvar}_i := \argmind{\Gvar_i} \widehat{E}_{x_i}(\Gvar_i|\theta)$ the clamped solution.
Networks trained using the classical lifted scheme via minimising the clamped energy have been shown to have a strong tendency to yield linear networks despite the hard constraints on activation functions. To mitigate this drawback, the authors introduce a contrastive variant of the classical lifted training scheme that minimises the following objective,
\begin{align*}
    \mathcal{E}_1(\theta)\; := \; \minimize{\Gvar_i} \widehat{E}_{x_i}(\Gvar_i | \theta) - \minimize{\Gvar_i} E(\Gvar_i|\theta) = \; \widehat{E}_{x_i}(\hat{\Gvar}_i |\theta) - E(\check{\Gvar}_i|\theta)
\end{align*}
which the authors term contrastive loss. The contrastive loss is constructed by subtracting the energy of the free solution from the energy of the clamped solution. Since $\widehat{E}_{\var_i}(\Gvar_i |\theta)$ further adds the clamp constraint $(\aux_{i})_{\nolayers} = \var_i$ to $E(\Gvar_i|\theta)$, therefore $\widehat{E}_{x_i}(\hat{\Gvar}_i | \theta) \geq E(\check{\Gvar}_i|\theta)$. Hence the contrastive objective is always non-negative. Using the unified framework, we can conveniently generalise the training objective for learnable parameters $\theta$ to write out as
\begin{align*}
\mathcal{E}_1(\theta) \; : = \; \;
& \minimize{\{\Gvar_i\}_i}  \;  \mathbb{E}_{i}\left[ \chi_{\{\var_i\}}\left(\Kb \Gvar_i + \GBiasd\right) + \frac{\mu}{2} \|\Mb\Gvar_i - (\Wb\Gvar_i + \GBias)\|^2_2 \right] \\
& \quad \quad - \minimize{\{\Gvar_i\}_i} \; \mathbb{E}_{i}\left[ \frac{\mu}{2} \|\Mb\Gvar_i - (\Wb\Gvar_i + \GBias)\|^2_2 \right] \,, \\
& \text{subject to} \quad  (\GM\Gvar_i)_j \succeq 0 \quad \text{for all}\,\, i, j \,.
\end{align*}

\subsection{Lifted Bregman Training in the Unified Framework}\label{sec:lifted_bregman_unified}

The lifted Bregman training framework, introduced in~\cite{wang2023liftedtraining}, generalises the previously discussed MAC-QP and original lifted method by employing a more sophisticated penalty function based on (generalised) Bregman distances~\cite{bregman1967relaxation,kiwiel1997proximal}. This approach not only provides a rigorous mathematical foundation but also presents a significant practical advantage: it eliminates the need to differentiate the activation functions during training if paired with a suitable optimisation strategy, thereby simplifying the resulting optimisation procedure and its gradient computations.

The framework is specifically designed for activation functions $\boldsymbol{\sigma}$ that are proximal maps.
This class of functions is extensive (cf.~\cite{zhang2017convergence,li2019lifted,combettes2020deep,hasannasab2020parseval}) and includes many common activations like the ReLU (cf. \cite{nair2010rectified}), soft-thresholding, the hyperbolic tangent and even the softmax function, each corresponding to a specific choice of $\Psi$. Here we present three examples of activation functions and their corresponding potentials $\Psi$ (see \cite[Example 1]{wang2023liftedtraining}). The first example is the classical ReLU activation function:
\begin{align*}
    \Psi(w) := \begin{cases} 0 & w \in [0, \infty)^n \\ \infty & \text{otherwise} \end{cases} \quad &\implies \quad \sigma(z)_j = \max( 0, z_j ) \, , \; \forall j \in \{1, \ldots, n \} \, .
\end{align*}
The second example we present is the soft-thresholding activation function where the $\Psi$ is a positive multiple of the one-norm: 
\begin{align*}
    \Psi(w) := \alpha \| w \|_1 \quad &\implies \quad  \sigma(z)_j = \begin{cases} z_j - \alpha & z_j > \alpha \\ 0 & |z_j| \leq \alpha \\ z_j + \alpha & z_j < -\alpha \end{cases} \, , \; \forall j \in \{1, \ldots, n \} \, ,
\end{align*}
for $\alpha > 0$.
And the last example we show is the hyperbolic tangent with $\Psi$ being:
\begin{align*}
    \Psi(w) := \begin{cases} w \tanh^{-1}(w) + \frac12 \left( \log(1 - w^2) - w^2 \right) & |w| < 1 \\ \infty & \text{otherwise} \end{cases}  \implies \sigma(z) = \tanh(z) \, .
\end{align*}

The core of the lifted Bregman approach is to replace the quadratic penalties in MAC-QP or the classical lifted approach with a tailored Bregman penalty function, denoted by $\mathcal{B}_{\Psi}$. Within our unified framework, this corresponds to the following choice for the penalty function $\mathcal{D}_{\boldsymbol{\sigma}}$:
\begin{align*}
    \mathcal{C}(\ab, \ab') = \chi_{\{\mathbf{0}\}}(\ab - \ab') \quad \text{and} \quad
    \mathcal{D}_{\boldsymbol{\sigma}}(\ub, \vb) = \mu \mathcal{B}_{\Psi}(\ub, \vb) \, ,
\end{align*}
where $\mu > 0$ is a penalty parameter. The Bregman penalty $\mathcal{B}_{\Psi}$ is defined as the Bregman distance generated by the potential function $\Phi(\cdot) = \frac{1}{2}\|\cdot\|^2_2 + \Psi(\cdot)$, which yields
\begin{align*}
    \mathcal{B}_{\Psi}(\ub, \vb) &\defeq D_{\Phi}^{\vb}(\ub, \boldsymbol{\sigma}(\vb)) = \Phi(\ub) - \Phi(\boldsymbol{\sigma}(\vb)) - \langle \vb, \ub - \boldsymbol{\sigma}(\vb) \rangle \\
    &= \frac{1}{2} \|\ub - \boldsymbol{\sigma}(\vb)\|^2_2 + D_{\Psi}^{\vb - \boldsymbol{\sigma}(\vb)}(\ub, \boldsymbol{\sigma}(\vb))  \\
    &= \frac{1}{2} \|\ub - \boldsymbol{\sigma}(\vb)\|^2_2 + \Psi(\ub) - \Psi(\boldsymbol{\sigma}(\vb)) - \langle \vb - \boldsymbol{\sigma}(\vb), \ub - \boldsymbol{\sigma}(\vb) \rangle \, .
\end{align*}

\noindent This decomposition is particularly insightful. It reveals that the Bregman penalty contains the standard quadratic penalty term $\frac{1}{2} \|\ub - \boldsymbol{\sigma}(\vb)\|^2_2$ used in MAC-QP, but augments it with a term that depends on the function $\Psi$ defining the activation. This additional term incorporates prior knowledge about the activation function directly into the penalty, making it more tailored to the specific network architecture.

The most significant advantage of this formulation lies in its gradient. The partial derivative of the Bregman penalty with respect to its second argument $\mathbf{v}$ is remarkably simple \cite[Lemma 8]{wang2023liftedtraining} and reads
\begin{align*}
    \nabla_{\mathbf{v}} \mathcal{B}_{\Psi}(\mathbf{u}, \mathbf{v}) = \mathbf{\boldsymbol{\sigma}}(\mathbf{v}) - \mathbf{u} \, .
\end{align*}

\noindent Crucially, this gradient does not depend on the derivative of the activation function $\boldsymbol{{\sigma}}$. In particular, using the lifted Bregman formulation allows
for training networks with non-smooth or even non-differentiable activation functions using gradient-based methods, without resorting to subgradient calculus and overcoming issues like vanishing gradients.

\begin{example}[Lifted Bregman for MLP]
For an MLP, the resulting optimisation problem is
\begin{align*}
    \minimize{\{\Gvar_i\}_i, \theta} \quad \mathbb{E}_{i}\left[ \loss(K (\aux_{i})_{J} + d) + \mu \sum_{j=1}^{J} \mathcal{B}_{\Psi_j}\left((\aux_{i})_{j}, \Glin_j (\aux_{i})_{j - 1} + \Gbias_j\right) \right] \, .
\end{align*}

\noindent This objective is optimised over the network parameters $\theta$ and the auxiliary variables $\{\Gvar_i\}_i$, typically using non-smooth first-order methods that can exploit the specific structure of the Bregman penalty~\cite{wang2023liftedtraining}.
\end{example}

\begin{remark}[Connection to Fenchel Lifted Networks]
The lifted Bregman framework is closely connected to the Fenchel lifted network approach discussed in \Cref{sec:fenchel}. This immediately becomes obvious if we rewrite the Bregman distance
\begin{align*}
    D_{\Phi}^{\mathbf{v}}(\mathbf{u}, \boldsymbol{\actfct}(\mathbf{v})) &= \Phi(\mathbf{u}) - \Phi(\boldsymbol{\actfct}(\mathbf{v})) - \langle \mathbf{v}, \mathbf{u} - \boldsymbol{\actfct}(\mathbf{v}) \rangle
\intertext{to}
    D_{\Phi}^{\mathbf{v}}(\mathbf{u}, \boldsymbol{\actfct}(\mathbf{v})) &= \Phi(\mathbf{u}) + \Phi^\ast(\mathbf{v}) - \langle \mathbf{v}, \mathbf{u} \rangle \, ,
\end{align*}

\noindent where the previous equality follows from the definition of the convex conjugate, i.e., $\Phi^\ast(\mathbf{v}) = \sup_{\mathbf{w}} \langle \mathbf{w}, \mathbf{v}\rangle - \Phi(\mathbf{w})$, and the fact that the supremum is attained at $\mathbf{w} = \boldsymbol{\actfct}(\mathbf{v}) = \text{prox}_{\Psi}(\mathbf{v})$. Hence, the Bregman distance is also a difference of the Fenchel-Young inequality, and is by definition non-negative. More importantly, $D_{\Phi}^{\mathbf{v}}(\mathbf{u}, \boldsymbol{\actfct}(\mathbf{v})) \leq 0$ is equivalent to $\mathbf{v} \in \partial \Phi(\mathbf{u})$ or $\mathbf{u} \in \partial \Phi^\ast(\mathbf{v}) = \left\{ \boldsymbol{\actfct}(\Gaux) \right\}$, respectively; consequently, $\mathbf{u} = \boldsymbol{\actfct}(\mathbf{v})$.

This equivalence reveals that the Bregman penalty $\mathcal{B}_{\Psi}(\mathbf{u}, \mathbf{v})$ is precisely the biconvex function $B_{\boldsymbol{\actfct}}(\mathbf{u}, \mathbf{v})$ utilised in Fenchel Lifted Networks, for the broad and important class of activation functions $\boldsymbol{\actfct}$ that can be expressed as proximal maps. The primary distinction between the two approaches is therefore one of perspective and construction. The Fenchel framework is presented more generally, relying on the existence of a suitable biconvex representation for the activation constraint. The Bregman framework, on the other hand, provides a systematic and constructive method for deriving this specific biconvex penalty for any activation function that is a proximal map.

Beyond this shared foundation, there are subtle but significant differences in their methodology and scope. The Fenchel-lifted approach typically deduces the biconvex function $B_{\boldsymbol{\actfct}}$ on a case-by-case basis, for instance, by analysing the optimality conditions for a specific activation like the ReLU. While effective, deriving these functions for other activations is not always straightforward. In contrast, the Bregman framework is inherently constructive; once an activation $\boldsymbol{\actfct}$ is identified as a proximal map $\text{prox}_{\Psi}$, the corresponding penalty function $\mathcal{B}_{\Psi}$ is generated automatically from the potential $\Psi$. Furthermore, the theoretical analysis underpinning Fenchel lifted networks often requires the activation function to be strictly monotone (which excludes cases such as the ReLU). The Bregman framework relaxes this requirement considerably, as its theoretical guarantees hold for the wider class of proximal activation functions that are monotone but not necessarily strictly monotone.

\remarkend
\end{remark}

\section{Implementation}\label{sec:implementation}

The practical minimisation of \cref{eq:loss} can be achieved through an inner- and outer-optimisation strategy based on \emph{implicit} stochastic gradient method (ISGM), or simply deterministically.

\subsection{Implicit Stochastic Gradient Method for MLPs}
The minimisation of the loss function \cref{eq:loss} can be framed as a stochastic optimisation problem. While conventional approaches would employ an explicit SGD method, a direct descendant of the classical Robbins--Monro procedure, such methods are known to be numerically unstable and critically sensitive to the choice of hyperparameters, particularly the learning rate~\cite{toulis2014statistical, toulis2015scalable}. Alternatively, one can employ ISGM. This approach, also known as implicit SGD or the proximal Robbins--Monro method, formulates the parameter update as a stochastic fixed-point equation~\cite{toulis2017asymptotic, toulis2020proximal}. As with most implicit methods, the principal advantage of this formulation is its enhanced numerical stability, which is demonstrated by its robustness to the choice of learning rate and initial conditions~\cite{toulis2014statistical, toulis2017asymptotic}. Crucially, this stability does not compromise asymptotic performance, as the method achieves the same optimal statistical efficiency as standard SGD in the limit~\cite{toulis2014statistical, toulis2017asymptotic, tran2016towards}. The ISGM iteration that we introduce in~\cref{eq:isgd} is precisely such an update, which requires the solution of an inner optimisation problem at each step, a task we address in subsequent sections. We assume that the architectural constraint in \cref{eq:loss} is strictly enforced, i.e., $\mathcal{C}(\GM\Gvar_i, \GV\Gaux_i) = 0$.

For an MLP where $\GV = I$, this constraint $\GM\Gvar_i = \Gaux_i$ allows us to eliminate $\Gaux_i$ and to work directly with $\Gvar_i$. Recall that $\Gvar_i$ is the concatenation of the input $\obs_i$ and the auxiliary variables $\Gaux_i = ( (\aux_{i})_1, (\aux_{i})_2, \ldots, (\aux_{i})_{J} )^\top$, i.e., $\Gvar_i = (\obs_i, (\aux_{i})_{1}, (\aux_{i})_{2}, \ldots, (\aux_{i})_{J})^\top$. In the optimisation process, we only optimise over the auxiliary variables $\Gaux_i$, as the input $\obs_i$ is fixed.

Suppose we have $\nosamples$ samples, i.e., $\mathbb{I} = \{1, \ldots, \nosamples\}$ and split these into $\nobatches$ batches $\{ S_p \}_{p = 1}^\nobatches$ such that $\bigcup_{p = 1}^\nobatches S_p = \mathbb{I}$ and $\bigcap_{p = 1}^\nobatches S_p = \emptyset$. For a batch $S_p$ and samples $\{ (\obs_i, \var_i)\}_{i \in S_p}$, we define
\begin{equation}\label{eq:batch_loss}
\uniloss^p\left(\theta, \{ \Gaux_i \}_{i \in S_p } \right) \defeq \sum_{i \in S_p} \left[ \mathcal{L}_{\var_i}(\GK\Gvar_i + \GBiasd) + \mathcal{D}_{\mathbf{\sigma}}(\GM\Gvar_i, \GLin\Gvar_i + \GBias) \right] \, ,
\end{equation}

\noindent such that $\uniloss(\theta, \{ \Gaux_i \}_{i = 1}^s ) = \frac{1}{s} \sum_{p = 1}^\nobatches \uniloss^p(\theta, \{ \Gaux_i \}_{i \in S_p} )$. Then, the corresponding implicit stochastic gradient iteration reads
\begin{equation}
    (\theta^{k+1}, \{ \Gaux_i^{k+1} \}_{i \in S_p}) = \argmind{\theta, \{ \Gaux_i \}_{i \in S_p}} \left\{ \uniloss^p\left(\theta, \{ \Gaux_i \}_{i \in S_p} \right) + \frac{1}{2\tau_k} \|\theta - \theta^{k}\|^2_2 \right\} \, ,\label{eq:isgd}
\end{equation}

\noindent where $\tau_k$is a (possibly iteration dependent) step-size parameter or learning rate. As mentioned earlier, in contrast to conventional SGD, each iterate \cref{eq:isgd} constitutes an implicit optimisation problem that needs to be solved with another optimisation procedure. The advantage of this approach is that we do not need to store all variables $\{\Gaux_i \}_{i = 1}^s$ in memory at all times, but only $\max_p |S_p|$ many, where $|S_p|$ denotes the cardinality of batch $S_p$. This is extremely helpful, given that each $\Gaux_i$ contains all auxiliary variables per sample, and memory requirements easily grow for deep networks with lots of layers and many auxiliary variables. In the following, we briefly discuss how \cref{eq:isgd} can be solved with conventional, deterministic algorithms.

\subsection{First-order Algorithms for Inner ISGM Problems}\label{sec:first_order_algorithms}
Each problem \cref{eq:isgd} constitutes a deterministic and continuous but non-convex optimisation problem, for which we can employ a vast array of optimisation algorithms. In the following, we assume that either the entire function \eqref{eq:batch_loss} is continuously differentiable, or consists of a continuously differentiable and a proximal part, so that first-order optimisation methods can be applied. Other algorithms can be implemented; discussing all possible optimisation algorithms that are suitable for the computation of \cref{eq:isgd} is beyond the scope of this work.

\subsubsection{The Differentiable Case}\label{sec:diff-case}
When the loss function $\loss$ and the penalty function $\mathcal{D}_{\boldsymbol{\actfct}}$ are both continuously differentiable, the entire batch loss objective $\uniloss^p$ in \cref{eq:batch_loss} becomes differentiable. This allows for the application of standard gradient-based optimisation methods, such as gradient descent or more advanced variants like Adam~\cite{kingma2014adam}, to solve the inner ISGM problem \cref{eq:isgd}. These methods require the computation of the gradients of $\uniloss^p$ with respect to the learnable parameters $\theta$ and the auxiliary variables $\{\Gaux_i\}_{i \in S_p}$. 

The gradient with respect to the auxiliary variables $\Gaux_j$ for a sample $j \in S_p$ is given by
\begin{align*}
\nabla_{\Gaux_j} \uniloss^p = \GK_{\text{aux}}^\top \nabla \mathcal{L}_{\var_j}(\GK\Gvar_j + \GBiasd) &+ \GM_{\text{aux}}^\top \nabla_1 \mathcal{D}_{\boldsymbol{\actfct}}(\GM\Gvar_j, \GLin\Gvar_j + \GBias) \\
&+ \GLin_{\text{aux}}^\top \nabla_2 \mathcal{D}_{\boldsymbol{\actfct}}(\GM\Gvar_j, \GLin\Gvar_j + \GBias) \, ,
\end{align*}
\noindent where $\GK_{\text{aux}}$, $\GM_{\text{aux}}$, and $\GLin_{\text{aux}}$ denote the sub-blocks of the operators $\GK$, $\GM$, and $\GLin$ that correspond to the auxiliary variables (excluding the input), and $\nabla_1 \mathcal{D}_{\actfct}$ and $\nabla_2 \mathcal{D}_{\actfct}$ denote the partial gradients of $\mathcal{D}_{\boldsymbol{\actfct}}$ with respect to its first and second arguments, respectively.

Similarly, the gradients with respect to the learnable parameters $\theta$ (which are the components of the operators $\GK, \GLin$ and biases $\GBiasd, \GBias$) are computed by summing the contributions from each sample in the batch. If we assume that all components of the operators and biases are learnable parameters, then their gradients can be expressed as
\begin{align*}
    \nabla_{\GK} \uniloss^p &= \sum_{i \in S_p} [\nabla \mathcal{L}_{\var_i}(\GK\Gvar_i + \GBiasd)] \Gvar_i^\top \, , \\
    \nabla_{\GBiasd} \uniloss^p &= \sum_{i \in S_p} \nabla \mathcal{L}_{\var_i}(\GK\Gvar_i + \GBiasd) \, , \\
    \nabla_{\GLin} \uniloss^p &= \sum_{i \in S_p} [\nabla_2 \mathcal{D}_{\actfct}(\GM\Gvar_i, \GLin\Gvar_i + \GBias)] \Gvar_i^\top \, , \\
    \nabla_{\GBias} \uniloss^p &= \sum_{i \in S_p} \nabla_2 \mathcal{D}_{\actfct}(\GM\Gvar_i, \GLin\Gvar_i + \GBias) \, .
\end{align*}

\noindent We note that while these expressions are provided for completeness, their manual implementation is usually not required in practice, since modern deep learning libraries such as PyTorch or JAX provide powerful automatic differentiation frameworks that can compute these gradients automatically.

\subsubsection{Proximal Gradient Descent}
In many practical applications, particularly those derived from the lifted training frameworks discussed in \Cref{sec:lifted_training}, the objective function $\uniloss^p$ is not fully differentiable. Instead, it often comprises the sum of a smooth, differentiable function and a non-smooth, convex function for which the proximity operator is efficiently computable. We consider the case where the batch loss can be additively split as
\[
\uniloss^p(\theta, \{\Gaux_i\}_{i \in S_p}) = \sum_{i \in S_p} \left[ \Psi(\GM\Gvar_i) + \diff(\theta, \Gaux_i) \right] \, ,
\]

\noindent where $\diff$ is continuously differentiable with respect to its arguments and $\Psi$ is a proper, lower semi-continuous, convex function. The term $\Psi(\GM\Gvar_i)$ introduces non-smoothness through the composition of the convex function $\Psi$ with the linear operator $\GM$.

While this composite structure prevents the direct application of standard gradient descent, it can be addressed by proximal splitting algorithms. If the proximity operator of the function $\Gaux_j \mapsto \Psi(\GM\Gvar_j)$ is efficiently computable, one can still employ proximal gradient descent. The update rule for a sample $j \in S_p$ would then be 
\begin{align*}
    \Gaux_j^{t+1} = \text{prox}_{\alpha_t (\Psi \circ \GM_{\text{aux}})}\left( \Gaux_j^t - \alpha_t \nabla_{\Gaux_j} \diff(\theta^t, \Gaux_j^t) \right) \, ,
\end{align*}
where $\alpha_t > 0$ is a step-size parameter and $\GM_{\text{aux}}$ denotes the sub-block of $\GM$ corresponding to the auxiliary variables. However, computing the proximity operator of a composition with arbitrary $\GM_{\text{aux}}$ can be challenging. In such cases, more general splitting methods like the Alternating Direction Method of Multipliers (ADMM) (cf. \cite{gabay1983chapter,boyd2011distributed}) or primal-dual algorithms like the Primal-Dual Hybrid Gradient (PDHG) method (cf. \cite{zhu2008efficient,pock2009algorithm,esser2010general,chambolle2011first,chambolle2016introduction}) are utilised, which avoid computing $\text{prox}_{\alpha_t (\Psi \circ \GM_{\text{aux}})}$, instead using the more easily computable $\text{prox}_{\alpha_t (\Psi)}$.

The update for the learnable parameters $\theta$ in this case reduces to a standard gradient step:
\begin{align*}
    \theta^{t+1} = \theta^t - \alpha_t \nabla_\theta \diff(\theta^t, \Gaux_j^t) \, .
\end{align*}

\noindent While convergence guarantees for proximal gradient descent applied to deterministic, non-convex problems exist (see, e.g., \cite{attouch2013convergence, bolte2014proximal}), we want to emphasise how block-coordinate descent variants can be used to solve \cref{eq:isgd} via sequences of convex optimisation problems in the next section.

\subsubsection{Adam Variant}\label{sec:prox-adam}

While the gradient and proximal gradient methods use plain first-order steps, it is often beneficial to employ adaptive and momentum-based schemes to accelerate convergence and stabilise training \cite{nesterov1983method,sutskever2013importance,huang2013accelerated,kingma2014adam,teboulle2018simplified,melchior2019proximal,mukkamala2020convex}. A widely used method of this type is Adam~\cite{kingma2014adam}, which maintains exponential moving averages of both the first moment (gradient) and the second moment (squared gradient) to form a preconditioned update direction. For the learnable parameters $\theta$, given the current iterate $(\theta^t,\Gaux_j^t)$, we first compute the gradient
\[
g^{t}_{\theta} := \nabla_\theta \diff\!\left(\theta^{t},\Gaux^{t}_{j}\right).
\]
The first and second moment estimates are then updated as
\begin{align*}
m^{t+1}_{\theta} &= p_1\, m^{t}_{\theta} + (1-p_1)\, g^{t}_{\theta},\\
q^{t+1}_{\theta} &= p_2\, q^{t}_{\theta} + (1-p_2)\, \left(g^{t}_{\theta} \odot g^{t}_{\theta}\right),
\end{align*}
where $p_1,p_2\in(0,1)$ are fixed decay parameters and $\odot$ denotes the Hadamard product. To correct for initialisation bias, we form
\[
\widehat m^{t+1}_{\theta}=\frac{m^{t+1}_{\theta}}{1-(p_1)^{t+1}},\qquad
\widehat q^{\, t+1}_{\theta}=\frac{q^{t+1}_{\theta}}{1-(p_2)^{t+1}}.
\]
The parameter update then becomes
\[
\theta^{t+1}_{\theta} = \theta^{t}_{\theta} - \alpha_t \,\frac{\widehat m^{t+1}_{\theta}}{\sqrt{\widehat q^{\, t+1}_{\theta}}+\varepsilon},
\]

\noindent where $\alpha_t$ is the step-size and $\varepsilon>0$ is a small constant for numerical stability.

In analogy with the update for $\theta$, we can extend the proximal gradient step for the auxiliary variables $\Gaux_j$ also by incorporating Adam style moment estimates. At iteration $t$, we denote the computed gradient with respect to $\Gaux_j$ by:
\[
g^{t}_{\Gaux} := \nabla_{\Gaux_j} \diff\!\left(\theta^{t},\Gaux^{t}_{j}\right).
\]
The exponential moving averages of the moments are  updated as
\begin{align*}
m^{t+1}_{\Gaux} &= p_1\, m^{t}_{\Gaux} + (1-p_1)\, g^{t}_{\Gaux},\\
q^{t+1}_{\Gaux} &= p_2\, q^{t}_{\Gaux} + (1-p_2)\, \left(g^{t}_{\Gaux} \odot g^{t}_{\Gaux}\right),
\end{align*}
with fixed parameters $p_1,p_2\in(0,1)$ and elementwise product $\odot$. Bias correction then yields
\[
\widehat m^{t+1}_{\Gaux}=\frac{m^{t+1}_{\Gaux}}{1-(p_1)^{t+1}},\qquad
\widehat q^{\, t+1}_{\Gaux}=\frac{q^{t+1}_{\Gaux}}{1-(p_2)^{t+1}}.
\]
Replacing the plain gradient step in the proximal gradient scheme with this Adam search direction, the update rule of $\Gaux_j$ for a sample $j \in S_p$ is then given by
\[
\Gaux^{t+1}_{j} = \operatorname{prox}_{\alpha_t (\Psi \circ \GM_{\text{aux}})}\!\left(
\Gaux^{t}_{j} - \alpha_t \,\frac{\widehat m^{t+1}_{\Gaux}}{\sqrt{\widehat q^{\, t+1}_{\Gaux}}+\varepsilon}
\right),
\]
where $\alpha_t$ is the step-size and $\GM_{\text{aux}}$ denotes the sub-block of $\GM$ corresponding to the auxiliary variables.

In this way, the Adam scheme extends the classical gradient step by adaptively scaling and smoothing the gradient information across iterations.

\subsection{Solving the Inner ISGM Problem via Block-Coordinate Descent}\label{sec:isgm-bcd}

The optimisation problem \cref{eq:isgd} that arises in each step of the implicit stochastic gradient method is a deterministic, non-convex problem. However, for many of the lifted training frameworks discussed in \Cref{sec:lifted_training}, this problem exhibits a favourable structure. Specifically, while it is jointly non-convex in the parameters $\theta$ and the auxiliary variables $\{ \Gaux_i \}_{i \in S_p}$, it is often convex in each block of variables when the other is held fixed. This property makes block-coordinate descent (BCD) an ideal solution strategy.
The BCD approach tackles the minimisation problem in \eqref{eq:isgd} by generating a sequence of iterates $\{(\theta^t, \{ \Gaux_i^t \}_{i \in S_p})\}_t = \{(\theta^t, V^t_p)\}_t$ via the alternating scheme
\begin{align*}
\theta^{t + 1} &\!=\! \argmind{\theta} \left\{ \sum_{i \in S_p} \left[ \loss(\GK\Gvar_i^t + \GBiasd) + \mathcal{D}_{\mathbf{\sigma}}(\GM\Gvar_i^t, \GLin\Gvar_i^t + \GBias) \right] + \frac{1}{2\tau_{k}} \|\theta - \theta^{k}\|^2_2 \right\} , \\
V^{t+1}_p
&\!=\! \argmind{\{ \Gaux_i \}_{i \in S_p}} \left\{ \sum_{i \in S_p} \left[ \loss(\GK^{t + 1}\Gvar_i + \GBiasd^{t + 1}) + \mathcal{D}_{\mathbf{\sigma}}(\GM\Gvar_i, \GLin^{t + 1}\Gvar_i + \GBias^{t + 1}) \right] \right\} .
\end{align*}
In the update for the parameters $\theta$, the subproblem is typically convex and often decouples into smaller, independent optimisation problems for the parameters of each layer. For example, when using quadratic penalties, the objective with respect to the parameters $(\GK, \GBiasd, \GLin, \GBias)$ becomes a sum of quadratic terms. These are essentially (nonlinear) least-squares problems that can be solved efficiently.
For the update of the auxiliary variables, a key advantage is that the problem is separable across the training samples in the batch $S_p$. This means we can solve for each $\Gaux_j$ independently and in parallel. Furthermore, we will see that, depending on the chosen architecture, many of the auxiliary components of $\Gaux_j$ are independent of each other and can be computed in parallel as well.

For the lifted frameworks, the resulting subproblem with respect to a single $\Gaux_j$ is generally convex. This allows us to find a global minimum for each $\Gaux_j$ efficiently using the deterministic first-order algorithms discussed in the previous \Cref{sec:first_order_algorithms}. In the following, we show what ISGM with block-coordinate descent looks like for the MAC-QP approach as introduced in \Cref{sec:mac}, when training MLPs.

\subsection{Solving the Inner ISGM Problem via Linearised Block-Coordinate Descent}

While the block-coordinate descent (BCD) method described in the previous section is powerful, it relies on our ability to solve the subproblems for both the parameters $\theta$ and the auxiliary variables $\{ \Gvar_i \}_{i \in S_p}$ exactly and efficiently at each iteration. Although these subproblems are often convex, finding their exact minimiser can still be computationally demanding, especially if they don't admit a closed-form solution.

An effective alternative is the linearised block-coordinate descent method, also known as block-coordinate gradient descent or alternating linearised minimisation. Instead of performing a full minimisation for each block, this approach simply takes one or more gradient steps with respect to that block. This trades the expensive full optimisation for a cheaper, approximate update, which can lead to faster overall convergence in practice.

The core idea is to replace the minimisation problems in the BCD updates with simple gradient descent steps. The alternating scheme for the sequence of iterates $\{(\theta^t, \{ \Gvar_i^t \}_{i \in S_p})\}_t$ within the inner ISGM loop then becomes
\begin{align*}
\theta^{t + 1} &= \theta^{t} - \alpha_t \nabla_{\theta} \left. \left( \uniloss^p\left(\theta, \{ \Gaux_i^t \}_{i \in S_p } \right) + \frac{1}{2\tau_k} \|\theta - \theta^k\|^2_2 \right) \right|_{\theta=\theta^t} \, , \\
\{ \Gaux_i^{t + 1} \}_{i \in S_p} &= \{ \Gaux_i^{t} \}_{i \in S_p} - \beta_t \nabla_{\{\Gaux_i\}} \left. \uniloss^p\left(\theta^{t+1}, \{ \Gaux_i \}_{i \in S_p } \right) \right|_{\{\Gaux_i\}=\{\Gaux_i^t\}} \, .
\end{align*}
Here, $\alpha_t > 0$ and $\beta_t > 0$ are step-size parameters for the parameter and auxiliary variable updates, respectively. Also note that the iteration index of the quadratic penalty stems from ISGM and therefore is $k$ and not $t$. The step-size parameters can be chosen as fixed constants, determined via a line search for $\alpha$ and $\beta$, or adapted during the optimisation using a learning rate schedule such as cosine annealing.

The gradients required for these updates are precisely those detailed in \Cref{sec:first_order_algorithms}. For instance, the gradient with respect to the parameter block $\theta$ would involve computing the gradients for each component ($\GK, \GBiasd, \GLin, \GBias$) as laid out for the differentiable case in \Cref{sec:diff-case}. Similarly, the update for the auxiliary variables is performed by taking a step in the negative direction of the gradient with respect to each $\Gaux_j$.

This linearised approach preserves the key benefits of BCD, such as the ability to update parameters and auxiliary variables in parallel across layers and samples, while significantly reducing the computational cost of each inner iteration. It is particularly well-suited for scenarios where the objective function is smooth and differentiable, as is the case for the MAC-QP framework applied to MLPs.

\subsection{Example: MLP with MAC-QP}\label{sec:example_mac_qp}

To make the abstract ISGM and BCD framework more concrete, we now detail the specific update steps for the MAC-QP as described in \Cref{sec:mac}, applied to a MLP as described in \Cref{sec:mlp}. For MAC-QP, the ISGM batch loss function \cref{eq:batch_loss} becomes
\begin{align*}
    \uniloss^p(\theta, \{ \Gvar_i \}_{i \in S_p}) = \sum_{i \in S_p} \left[ \mathcal{L}_{\var_i}(\GK \Gvar_i + \GBiasd) + \frac{\mu}{2} \| \GM \Gvar_i - \actfct(\Wb \Gvar_i + \GBias) \|^2_2 \right] \, .
\end{align*}

\noindent Similar to \Cref{sec:mac}, we can specifically write this with auxiliary coordinates and quadratic loss function as 
\begin{align*}
\uniloss^p(\theta, \{ \Gvar_i \}_{i \in S_p})
& = \sum_{i \in S_p} \Big[ \frac12 \| K (\aux_{i})_{ \nolayers} + d - \var_i \|^2  \\ &\qquad + \frac{\mu}{2} \sum_{j = 1}^{\nolayers} \| (\aux_{i})_{j} - \sigma_j(\Glin_j (\aux_{i})_{j-1} + \Gbias_j)\|^2_2 \Big] \, .
\end{align*}

\noindent Hence, the ISGM iteration \cref{eq:isgd} for an MLP with MAC-QP becomes 
\begin{align}
\begin{split}
    (\theta^{k+1}, \{ \Gaux_i^{{k+1}}\}_{i \in S_p}) = \argmind{\theta, \{ \Gaux_i \}_{i \in S_p}} &\Bigg\{ \sum_{i \in S_p} \Big[ \frac12 \| K (\aux_{i})_{J} + d - \var_i \|^2 \vphantom{\sum_{j=1}^{J}}\\
    & + \frac{\mu}{2} \sum_{j=1}^{J} \| (\aux_{i})_{j} - \sigma_j(\Glin_j (\aux_{i})_{j-1} + \Gbias_j)\|^2_2 \Big] \\
    &+ \frac{1}{2\tau_{k}} \|\theta - \theta^{k}\|^2_2 \vphantom{\sum_{j=1}^{J}}\Bigg\} \, ,
\end{split}\label{eq:isgd_mlp}
\end{align}

\noindent with $(\aux_{i})_{0} = \obs_i$. Solving \cref{eq:isgd_mlp} with BCD then requires the solution of the sequence of iterates $\{(\theta^t, \{ \Gaux_i^t \}_{i \in S_p})\}_t$ via the alternating scheme
\begin{align*}
\theta^{t + 1}
&= \argmind{\theta} \Bigg\{ \sum_{i \in S_p} \Big[ \frac12 \| K (\aux_{i})_{J}^t + d - \var_i \|^2 \\
 &\qquad\qquad+ \frac{\mu}{2} \sum_{j=1}^{J} \| (\aux_{i})_{j}^t - \sigma_j(\Glin_j (\aux_{i})_{j-1}^t + \Gbias_j)\|^2_2 \Big] + \frac{1}{2\tau_{k}} \|\theta - \theta^{{k}}\|^2_2 \Bigg\} \, ,\\
(\aux_{i})_{j}^{t + 1}
&= \argmind{\aux_{i, j}} \Bigg\{ \sum_{i \in S_p} \Big[ \frac12 \| K^{t + 1} (\aux_{i})_{J} + d^{t + 1} - \var_i \|^2\\
&\qquad \qquad +\frac{\mu}{2} \sum_{j=1}^{J} \| (\aux_{i})_{j} - \sigma_j(\Glin_j^{t + 1} (\aux_{i})_{j-1} + \Gbias_j^{t + 1})\|^2_2 \Big] \Bigg\} \, ,
\end{align*}

\noindent for $j = 1, \ldots, J$ and $i \in S_p$. The first step updates the parameters $\theta = (K, d, \{\Glin_j, \Gbias_j\}_{j=1}^J)$, while the second step updates the auxiliary variables $\Gvar_i = (\obs_i, (\aux_{i})_{1}, \dots, (\aux_{i})_{J})^\top$ for each sample $i$ in the batch $S_p$. We note that due to the separable structure of the batch loss function, many of these optimisation problems can be solved independently for individual parameters, samples and auxiliary variables, allowing for efficient parallelisation. In particular, for parameters $K$ and $d$, we have
\begin{align*}
\begin{split}
(K^{t + 1}, d^{t + 1}) = \argmind{(K, d)} & \Bigg\{ \sum_{i \in S_p} \left[ \frac12 \| K (\aux_{i})_{J}^t + d - \var_i \|^2 \right].\\
& + \frac{1}{2\tau_{k}} \|K - K^{k}\|^2_2 + \frac{1}{2\tau_{k}} \|d - d^{{k}}\|^2_2 \Bigg\} \, ,
\end{split}
\end{align*}

\noindent where, with slight abuse of notation, we denote both the Euclidean norm and the Frobenius norm by $\|\cdot\|_2$. This is a standard least-squares problem that can be solved efficiently, e.g., via the conjugate gradient method.

\noindent For the linear layers $\Glin_j$ and biases $\Gbias_j$, we have
\begin{align*}
\begin{split}
    (\Glin_j^{t + 1}, \Gbias_j^{t + 1}) = \argmind{(\Glin_j, \Gbias_j)} &\Bigg\{ \sum_{i \in S_p} \left[ \frac{\mu}{2} \| (\aux_{i})_{j}^t - \sigma_j(\Glin_j (\aux_{i})_{j-1}^t + \Gbias_j)\|^2_2 \right] \\
&+ \frac{1}{2\tau_{k}} \|\Glin_j - \Glin_j^{{k}}\|^2_2 + \frac{1}{2\tau_{{k}}} \|\Gbias_j - \Gbias_j^{{k}}\|^2_2 \Bigg\} \, ,
\end{split}
\end{align*}

\noindent for each $j \in \{ 1, \ldots, J\}$. Hence, each weight and bias can be updated independently in parallel. This is a nonlinear least-squares problem, which can be solved efficiently using first-order methods as described in \Cref{sec:first_order_algorithms}, or alternatively with methods such as the Gauss--Newton method or similar techniques.

\noindent For each of the auxiliary variables $((\aux_{i})_{1}, \ldots, (\aux_{i})_{J})$, the update step reads

\[
\begin{split}
(\aux_{i})_{j}^{t + 1} = \argmind{(\aux_{i})_{\nolayers}} \bigg\{ \frac12 \| K^{t + 1} (\aux_{i})_{\nolayers} + d^{t + 1} - \var_i \|^2 \\
+ \frac{\mu}{2} \| (\aux_{i})_{\nolayers} - \sigma_{\nolayers}(\Glin_{\nolayers}^{t + 1} (\aux_{i})_{\nolayers-1} + \Gbias_{\nolayers}^{t + 1})\|^2_2 \bigg\}
\end{split}
\]
\noindent if $j = \nolayers$ and
\[
\begin{split}
 (\aux_{i})_{j}^{t + 1} = \argmind{(\aux_{i})_{j}} \bigg\{ \frac{\mu}{2} \| (\aux_{i})_{j} - \sigma_j(\Glin_j^{t + 1} (\aux_{i})_{j-1}+ \Gbias_j^{t + 1})\|^2_2 \\
+ \frac{\mu}{2} \| (\aux_{i})_{j+1} - \sigma_j(\Glin_{j + 1}^{t + 1} (\aux_{i})_{j}+ \Gbias_{j + 1}^{t + 1})\|^2_2\bigg\}
\end{split}
\]

\noindent if $ j < \nolayers$, respectively. We see that each auxiliary variable $(\aux_{i})_{j}$ is only updated based on its neighboring variables, which allows for all even and all odd auxiliary variables to be updated in parallel.

Alternatively, we can employ the linearised block-coordinate descent strategy to avoid solving the non-linear least-squares problems for the parameters and auxiliary variables. Instead of a full minimisation, we perform gradient descent steps for each block. The update for the parameters $(\Glin_j, \Gbias_j)$ for $j \in \{1, \ldots, J\}$ would then be 
\begin{align*}
    \Glin_j^{t + 1} &= \Glin_j^t - \alpha_t \left( \frac{1}{\tau_{k}}{(\Glin_j^t - \Glin_j^k)} - \mu \sum_{i \in S_p} (\delta_{i})_{j}^t ((\aux_{i})_{j}^t)^\top \right) \, , \\
    \Gbias_j^{t + 1} &= \Gbias_j^t - \alpha_t \left( \frac{1}{\tau_{k}}{(\Gbias_j^t - \Gbias_j^k)} - \mu \sum_{i \in S_p} (\delta_{i})_{j}^t \right) \, ,
\end{align*}

\noindent where $(\delta_{i})_{j}^t = ((\aux_{i})_{j}^t - \sigma_j(\Glin_j^t (\aux_{i})_{j-1}^t + \Gbias_j^t)) \nabla\sigma_j(\Glin_j^t (\aux_{i})_{j-1}^t + \Gbias_j^t)$, and $\alpha_t$ is a step-size. The updates for the final layer parameters $(K, d)$ are simple gradient steps on a quadratic and are not detailed here.

\noindent Similarly, the updates for the auxiliary variables $(\aux_{i})_{j}$ for $j < J$ become
\begin{align*}
    (\aux_{i})_{j}^{t+1} = (\aux_{i})_{j}^t - \beta_t \left( \mu((\aux_{i})_{j}^t - \sigma_j(\cdot)) - \mu (\Glin_{j+1}^{t+1})^\top (\delta_{i})_{j+1}^t \right) \, ,
\end{align*}

\noindent with a similar update for the final auxiliary variable $(\aux_{i})_{\nolayers}$ that also includes the gradient from the data fidelity term. This linearised approach is computationally cheaper per iteration but, unlike the Bregman-based methods discussed later, requires the activation functions $\sigma_j$ to be differentiable.

In \Cref{sec:lifted-bregman-mlp} we will show a more intricate example of how the BCD framework and Adam scheme can be applied to MLP type of networks in combination with the lifted Bregman framework.

\section{Applications to Inverse Problems}\label{sec:applications_inverse_problems}

The unified framework for lifted training provides a powerful lens through which we can approach the solution of inverse problems using deep learning. As discussed in the introduction, solving an inverse problem $\Meas(\var) = \obs$ requires regularisation to overcome ill-posedness. Learning-based methods aim to learn this regularisation from data, often by training a network $\net$ to approximate the inverse mapping, $\net(\obs) \approx \var$. The lifted framework offers two distinct but complementary strategies in this domain.

First, it enables the robust and efficient training of specialised network architectures tailored for inverse problems, such as PNNs or unrolled algorithms. These architectures inherently combine the structure of classical optimisation algorithms with the expressive power of learned parameters. Lifted training, particularly the Bregman approach, is exceptionally well-suited for these networks as their activations are typically proximal maps (e.g., soft-thresholding). We explore this in the context of MLPs with proximal activations in \Cref{sec:lifted-bregman-mlp}, building on recent work~\cite{wang2024lifted}.

Second, the lifted framework can be utilised for the stable inversion of pre-trained neural networks. In scenarios where a network models a forward process, $\net(\var) = \obs$, recovering the input $\var$ from an observation $\obs$ is itself an ill-posed inverse problem. The lifted formulation provides a rigorous variational regularisation strategy for this inversion, as detailed in \Cref{sec:lifted-inversion} following the methodology introduced in~\cite{wang2023liftedinversion}.

While we present these two lifting strategies within the context of inverse problems, where training can be viewed as an inverse problem in the weight space, we want to emphasise that the resulting optimisation framework is general and applicable to a wide array of regression and machine learning tasks.

\subsection{Lifted Inversion of Neural Networks}
\label{sec:lifted-inversion}
The task of inverting a neural network, i.e., determining the input $\var$ that produces a given output $\obs = \net(\var)$, is a fundamental problem in understanding and interpreting deep learning models \cite{linden1989inversion,mahendran2015understanding}. Like most inverse problem, this inversion is usually ill-posed, meaning a solution may not exist, may not be unique, or may not depend continuously on the input data. Consequently, stable inversion requires the use of regularisation techniques, a well-established concept in the field of inverse problems (cf. \cite{engl1996regularization, benning2018modern}).

The lifted Bregman framework, discussed in \Cref{sec:lifted_bregman_unified} for network training, provides a natural and powerful foundation for the regularised inversion of pre-trained neural networks \cite{wang2023liftedinversion}. While training aims to find the optimal parameters $\theta$ for a given dataset, inversion assumes the parameters $\theta$ are fixed and seeks to find the input $\var$ that corresponds to a measured output $\obs^\delta$, a potentially noisy version of the true output $\obs$.

Following \cite{wang2023liftedinversion}, the inversion of an $\nolayers$-layer MLP can be formulated as estimating the auxiliary variables and the input for given weights. We introduce auxiliary variables $\aux_1, \dots, \aux_{\nolayers-1}$ representing the outputs of the hidden layers, and the input to be recovered is $\var = \aux_0$. The goal is to find the set of variables $\{\aux_0, \dots, \aux_{\nolayers-1}\}$ that are consistent with the network architecture and the measured output $\obs^\delta$.

The lifted Bregman approach formulates this as a variational regularisation problem. Instead of enforcing the architectural constraints $\aux_j = \actfct_j(\Glin_j \aux_{j-1} + \Gbias_j)$ strictly, the model penalises deviations using the same Bregman penalties $\mathcal{B}_{\Psi_j}$ that were used for training. The resulting optimisation problem reads
\begin{equation}\label{eq:lifted_inversion}
\begin{aligned}
\minimize{\{\aux_j\}_{j=0}^{\nolayers-1}}
&\Bigg\{\mathcal{B}_{\Psi_\nolayers}(\obs^\delta, \Glin_\nolayers \aux_{\nolayers-1} + \Gbias_\nolayers) \\
&+ \sum_{j=1}^{\nolayers-1} \mathcal{B}_{\Psi_j}(\aux_j, \Glin_j \aux_{j-1} + \Gbias_j) + \reghyp \mathcal{R}(\aux_0) \Bigg\} \, ,
\end{aligned}
\end{equation}

\noindent where $\aux_0 = \var$ is the input we wish to recover. The final term, $\mathcal{R}(\var)$, is a regularisation term that incorporates prior knowledge about the desired input $\var$, with $\reghyp > 0$ controlling the strength of this prior. The first term, $\mathcal{B}_{\Psi_\nolayers}(\obs^\delta, \Glin_\nolayers \aux_{\nolayers-1} + \Gbias_\nolayers)$, acts as the data fidelity term, measuring the discrepancy between the final layer's pre-activation and the measurement $\obs^\delta$ in a way that is linked to the activation function $\actfct_\nolayers = \text{prox}_{\Psi_\nolayers}$.

A key advantage of this formulation is that in some cases it allows for a rigorous convergence analysis, particularly for the single-layer perceptron case ($\nolayers=1$, see \cite{wang2023liftedinversion}). In this scenario, problem \eqref{eq:lifted_inversion} simplifies to
\begin{equation}
\minimize{\var} \left\{ \mathcal{B}_{\Psi}(\obs^\delta, \Glin \var + \Gbias) + \reghyp \mathcal{R}(\var) \right\} \, . \label{eq:lifted_inversion_perceptron}
\end{equation}
This is a convex optimisation problem, for which well-posedness and stability can be established. More importantly, it can be proven that this formulation constitutes a convergent variational regularisation method \cite[Theorem 2]{wang2023liftedinversion}.

Specifically, if the true solution $\var^\dagger$ satisfies a source condition of the form $W^*v^\dagger \in \partial R(\var^\dagger)$ for some $v^\dagger$, and the noise level is bounded such that \[\mathcal{B}_\Psi(\obs^\delta, \Glin \var^\dagger + \Gbias) \le \delta^2,\] then the solution $\var_\reghyp$ to \cref{eq:lifted_inversion_perceptron} satisfies the error estimate
\begin{equation*}
\reghyp D_{\mathcal{R}}^{\text{symm}}(\var_\reghyp, \var^\dagger) \le (1+c)\delta^2 + \frac{\reghyp^2}{c} \|v^\dagger\|^2 + 2c J_\Psi\left(\obs^\delta + \frac{\reghyp}{c}v^\dagger, \obs^\delta - \frac{\reghyp}{c}v^\dagger\right) \, ,
\end{equation*}

\noindent for any $c \in (0, 1]$. Here, $D_{\mathcal{R}}^{\text{symm}}$ is the symmetric Bregman distance associated with the regularisation function $\mathcal{R}$, and $J_\Psi$ is the Burbea-Rao divergence related to the potential $\Psi$ i.e.,
\[
J_\Psi(v,v')
\;:=\;
\frac{1}{2}\Bigl(\Psi(v)+\Psi(v')-2\Psi\!\left(\tfrac{v+v'}{2}\right)\Bigr),
\qquad \forall \: v,v' \in \operatorname{dom}(\Psi).
\]
This estimate guarantees that as the noise $\delta$ goes to zero, the regularised solution $\var_\reghyp$ converges to the true solution $\var^\dagger$ (in the sense of the Bregman distance) for a suitable choice of the regularisation parameter $\reghyp(\delta)$. This provides the first (to our knowledge) rigorous convergence proof for a model-based inversion of a neural network without restrictive assumptions like differentiability of the activation function or tangential cone conditions. For instance, for the ReLU activation, the Burbea-Rao term vanishes under mild conditions, leading to a traditional convergence rate (see \cite[Example 1]{wang2023liftedinversion}).

While a full convergence theory for the joint, non-convex multi-layer problem \eqref{eq:lifted_inversion} is still an open question, the problem can be tackled computationally via an alternating minimisation or block-coordinate descent scheme. This approach sequentially solves for each variable ($\var, \aux_1, \dots, \aux_{\nolayers-1}$) while keeping the others fixed, leveraging the fact that each sub-problem is convex. For the single-layer problem \eqref{eq:lifted_inversion_perceptron}, efficient primal-dual algorithms can be employed \cite{chambolle2011first, wang2023liftedinversion}. Numerical results demonstrate that this approach can effectively invert both single- and multi-layer networks. When evaluated on the same data, the resulting reconstructions are visually superior to those obtained with pre-trained decoders.

\subsection{Lifted Bregman Strategy for Learning MLPs}\label{sec:lifted-bregman-mlp}

While the lifted Bregman strategy applies broadly across different families of neural network architectures~\cite{wang_2023}, in this section we detail an example for training an MLP (see \Cref{eq:mlp}). In particular, we consider the family of MLP networks that utilise the soft-thresholding operator as their activation functions; we have
\begin{align*}
    \soft{\lambda}(\mathbf{v}) \defeq \text{prox}_{\lambda \|\cdot\|_1}(\mathbf{v}) = \arg\min_{\mathbf{u}} \left\{ \frac{1}{2} \|\mathbf{u} - \mathbf{v}\|^2_2 + \lambda \|\mathbf{u}\|_1 \right\} \, .
\end{align*}

\noindent Within our unified framework, training this type of MLPs with the lifted Bregman approach corresponds to the following choice of penalty functions:
\begin{align*}
    \mathcal{C}(\ab, \ab') = \chi_{\{\mathbf{0}\}}(\ab - \ab') \quad \text{and} \quad
    \mathcal{D}_{\mathbf{\sigma}}(\mathbf{u}, \mathbf{v}) = \mu \mathcal{B}_{\Psi}(\mathbf{u}, \mathbf{v}) \, ,
\end{align*}

\noindent where the Bregman penalty $\mathcal{B}_{\Psi}$ is now computed as the Bregman distance generated by the potential function $\Phi(\cdot) = \frac{1}{2}\|\cdot\|^2_2 + \lambda \|\cdot\|_1$. Hence the original problem is lifted into minimising the following objective function:
\begin{align*}
    \minimize{\{\aux_i\}_i, \theta} \quad \mathbb{E}_{i}\left[ \loss(\Lin_J (\aux_{i})_{\nolayers-1}) + \sum_{j=1}^{\nolayers-1} \mu_j \mathcal{B}_{\psi_j}\left((\aux_{i})_{j}, \Glin_{j} (\aux_{i})_{j-1} + \Gbias_j\right) \right] \, .
\end{align*}
This total objective can be split into the sum of a Lipschitz-differentiable function $\diff$ and a proximable function (i.e. a function that has a closed form proximal operator) $\Psi$ using the property of the Bregman penalty function, resulting in the following form 
\begin{equation*}
\uniloss = \Psi(\Gaux_i) + \diff(\theta, \Gaux_i) \,,
\end{equation*}
with $\theta = (K, d, \{\Glin_j, \Gbias_j\}_{j=1}^J)$, where
\begin{align*}
 \diff(\theta, \Gaux_i) = & \quad \mathbb{E}_{i}\left[ \loss(\Lin_J (\aux_{i})_{\nolayers-1}) \right]\\
 & + \mathbb{E}_{i} \Bigg[\sum_{j=1}^{\nolayers-1} \mu_{j} \Big( \frac{1}{2}\|\aux_{j}\|^2 + \Big(\frac{1}{2}\|\cdot \|^2+\psi_j\Big)^* \Big(\Glin_{j} (\aux_{i})_{j-1} + \Gbias_j\Big)\\
&\qquad - \langle \aux_{j}, \Glin_{j} (\aux_{i})_{j-1} + \Gbias_j \rangle \Big) \Bigg],
\end{align*}

\noindent and
\begin{equation*}
\Psi(\Gaux_i) = \mathbb{E}_{i} \left[\sum_{j=1}^{\nolayers-1} \mu_j \| (\aux_{i})_{j} \|_1 \right] \,.
\end{equation*}

\noindent One way to proceed is to use a block-coordinate forward-backward strategy and alternatingly optimise between $\theta$ and $\Gaux_i$ to minimise the training objective deterministically. The iterations can then be summarised as:
\begin{align}
\theta^{t + 1} &= \theta^{t}  - \alpha_{t} \nabla_{\theta} \diff(\theta^{t},\Gaux_i^t) \, ,\\
\Gaux^{t+1}_{i} &= \soft{\beta_{t} \lambda \mu} \big( \Gaux_i^{t} - \beta_{t} \nabla_{\Gaux} \diff(\theta^{t+1},\Gaux_i^{t}) \big) \,.
\end{align}\label{algo:pgd-mlp}
The first step updates the parameters $\theta = (K, d, \{\Glin_j, \Gbias_j\}_{j=1}^J)$ via gradient methods with step-size parameter {$\alpha_{t}$}, while the second step updates the auxiliary variables via proximal gradient
methods with step-size parameter {$\beta_{t}$}.

Since $\diff(\theta, \Gaux_i)$ is differentiable, we could use automatic differentiation and the chain rule to compute its partial gradients with respect to its arguments {$\Gaux_i^{t}$ and $\theta^{t}$. On the other hand the step-sizes $\alpha_{t}$ and $\beta_{t}$} can be computed by either estimating the Lipschitz constants of operator $\Glin_j$, backtracking computations or using diminishing rates.

\section{Numerical Results}\label{sec:numerical_results}

In this section, we present numerical experiments that demonstrate the key features and performance of the lifted training and inversion framework. We first analyse the computational efficiency gained by exploiting the block structure of the unified framework for parallelisation. Subsequently, we evaluate the performance of the lifted Bregman training strategy on several canonical inverse problems (deblurring, denoising, and inpainting) using MLPs with proximal activations. Finally, we demonstrate the application of the lifted framework for the stable inversion of a pre-trained autoencoder.

All experiments are implemented in Python using the PyTorch library and our code is available on \href{https://github.com/model-based-and-data-driven-inversion/unified-framework-for-lifted-inversion.git}{GitHub}. Runtime comparison results are performed on an Apple M1 system using the PyTorch MPS backend for GPU acceleration. All training experiments are conducted on NVIDIA GeForce RTX 2080 Ti. 

\subsection{Unified Framework Implementation Details}

\begin{table}
\caption{Runtime comparison of the forward pass in the unified framework, contrasting vectorisation across layers with sequential evaluation without vectorisation.}
\centering
  \begin{filecontents*}{csv/forwards_final.csv}
Layers,Vmap (ms),No Vmap (ms),Speed-Up
1,0.211,0.091,0.43
2,0.201,0.135,0.67
4,0.201,0.295,1.47
8,0.205,0.671,3.28
16,0.186,1.288,6.92
32,0.185,2.36,12.76
64,0.195,4.797,24.64
128,0.197,10.656,54.16
\end{filecontents*}
\csvreader[
    separator=comma,
    tabular=rrrr, 
    table head=\toprule Layers & Vectorised (ms) & Non-Vectorised (ms) & Speed-Up\\\midrule,
    late after line=\\,
    table foot=\bottomrule
  ]{csv/forwards_final.csv}{}{\csvlinetotablerow}
\label{table:forward}
\end{table}

Earlier in \cref{eq:mlp}, we showed that an MLP can be expressed in both a sequential layer-wise structure and within the unified framework. In the case of the unified framework, we exploit its block structure to vectorise across layers.

We report a performance analysis as a function of the number of layers in \Cref{table:forward} and \Cref{table:back}, respectively. We consider between 1 and 128 layers. The results in \Cref{table:forward} are for the forward pass in the unified framework. The results in \Cref{table:back} are for the back-propagation.

We compare both formulations directly with one another, with identical auxiliary variables as input. By computing the intermediate auxiliary states with both implementations, we found both implementations produced outputs that were numerically identical (bit-for-bit equality in single point precision), confirming their equivalence discussed in \Cref{sec:mlp}. Unlike the sequential layer-wise structure, the block structure of the unified framework allows us to vectorise computation across layers, providing a performance advantage as shown in \Cref{table:back} and \Cref{table:forward} for deep networks, respectively.

\begin{table}
\caption{Runtime comparison of back-propagation in the unified framework, contrasting vectorisation across layers with sequential evaluation without vectorisation.}
\centering
  \begin{filecontents*}{csv/back_final.csv}
Layers,Vmap (ms),No Vmap (ms),Speed-Up
1,1.724,1.503,0.87
2,1.814,1.931,1.06
4,1.850,2.840,1.54
8,1.796,4.319,2.41
16,1.869,38.679,20.7
32,1.727,137.044,79.37
64,1.784,515.212,288.79
128,1.894,2011.176,1062.07
\end{filecontents*}
\csvreader[
    separator=comma,
    tabular=rrrr, 
    table head=\toprule Layers & Vectorised (ms) & Non-Vectorised (ms) & Speed-Up\\\midrule,
    late after line=\\,
    table foot=\bottomrule
  ]{csv/back_final.csv}{}{\csvlinetotablerow}
\label{table:back}
\end{table}

Vectorising layers within a unified framework yields substantial speed-ups during back-propagation through very deep networks (see \Cref{table:back}), with the effect being more pronounced than in the forward pass.

\subsection{Lifted Bregman Training}

In this section, we present numerical results for training MLPs via the lifted Bregman strategy as outlined in \Cref{sec:lifted-bregman-mlp}. We evaluate our training strategy across three canonical inverse problems: denoising, deblurring, and inpainting in a model agnostic manner by formulating each task as \cref{eq:inverse_problem}, differing only in the forward operator $H$ (either identity, convolution, or masking).
\begin{itemize}
\item \textbf{Deblurring:} Images are degraded with a Gaussian blur (kernel size = 5, $\sigma=1$) and mild additive Gaussian noise ($\sigma=0.03$).
\item \textbf{Denoising:} Images are corrupted with additive Gaussian noise of higher intensity ($\sigma=0.15$).
\item \textbf{Inpainting:} A random $30\%$ of pixels is removed from each image.
\end{itemize}

We keep the same architecture and image prior across different tasks. Each task uses 5,000 training images and 500 validation images sampled from the full MNIST dataset~\cite{lecun1998mnist}, reshaped to $28 \times 28$ grayscale inputs. In particular, we consider a 7-layer MLP model \cref{eq:mlp}, with fully-connected linear layers. We fix the hidden dimension at 784 across all layers and use the soft-shrinkage activation function. This choice of activation function is motivated by the LISTA framework, where sparsity of the hidden states is promoted via an $\ell_1$-norm regularisation. We further consider two settings for the shrinkage parameter that controls the activation strength: a stronger regularisation with $\lambda = 2 \times 10^{-1}$ and a weaker one where $\lambda = 2 \times 10^{-2}$.

We choose $\mu_j = 5 \times 10^{-3}$ for all layers to balance the Bregman penalty terms and the loss function $\loss$. We adopt the block coordinate descent (BCD) approach to minimise the training objective, and update $\theta$ and $\Gaux$ via the alternating scheme outlined in \Cref{sec:isgm-bcd}. For each of the alternating minimisation problem, we implement the deterministic first-order algorithms discussed in \Cref{sec:first_order_algorithms}.

A standard option is to follow the proximal gradient descent scheme presented in \Cref{algo:pgd-mlp}. The lifted Bregman training strategy deterministically minimises the Bregman training objective by alternating a gradient step for updating $\theta$, followed by a proximal gradient step for updating $\Gaux$. Alternatively, the gradient updates for $\theta$ and $\Gaux$ can be implemented using different variants of the gradient-based optimisers. As an example, one may employ Nesterov or Heavyball momentum methods to accelerate convergence. One could also consider adaptive and momentum-based schemes to stabilise training, resulting in an Adam-type variant of the proximal gradient descent algorithm, as described in \Cref{sec:prox-adam}.

\begin{figure}
    \centering
    \includegraphics[width=0.49\linewidth]{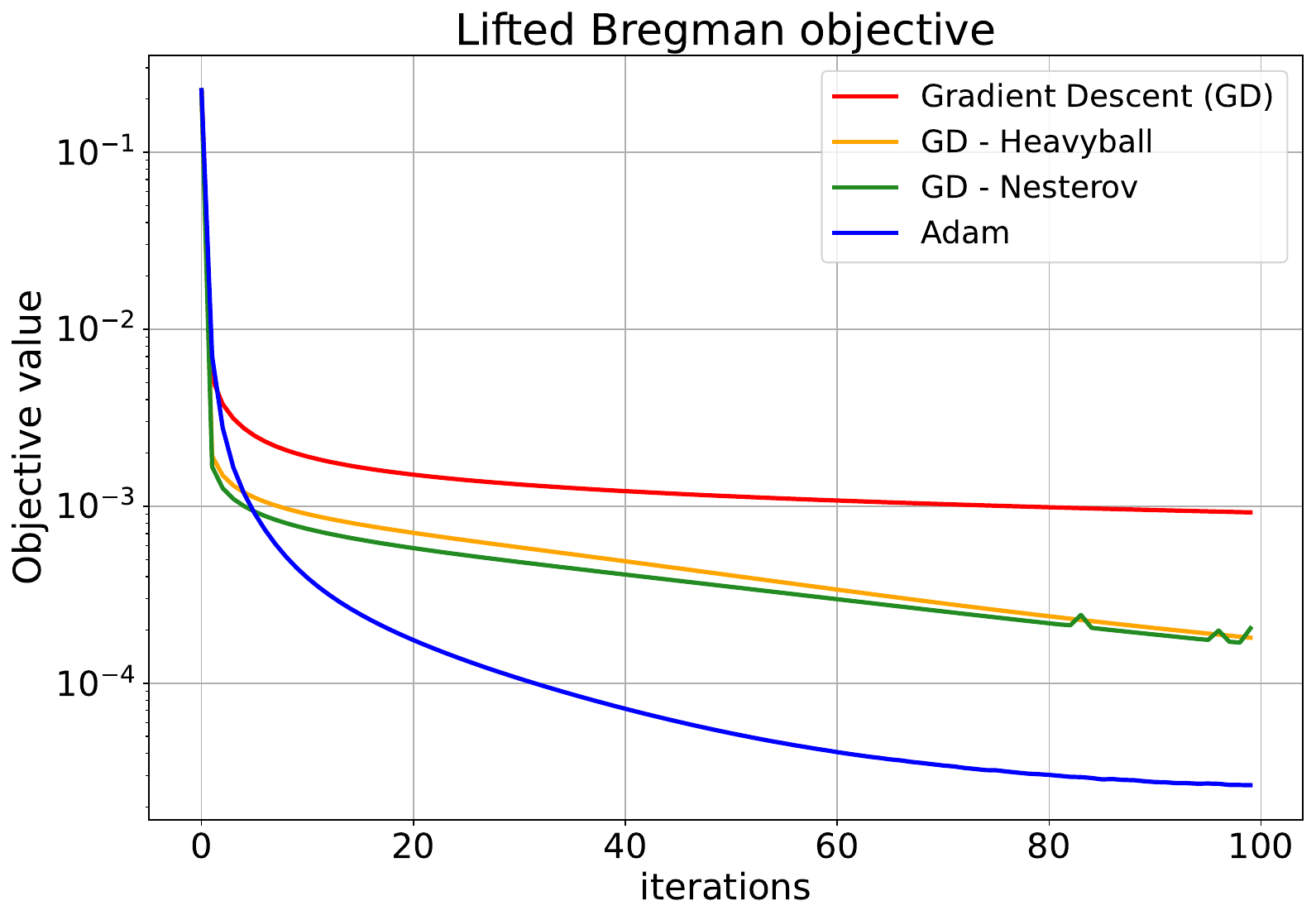}
    \includegraphics[width=0.49\linewidth]{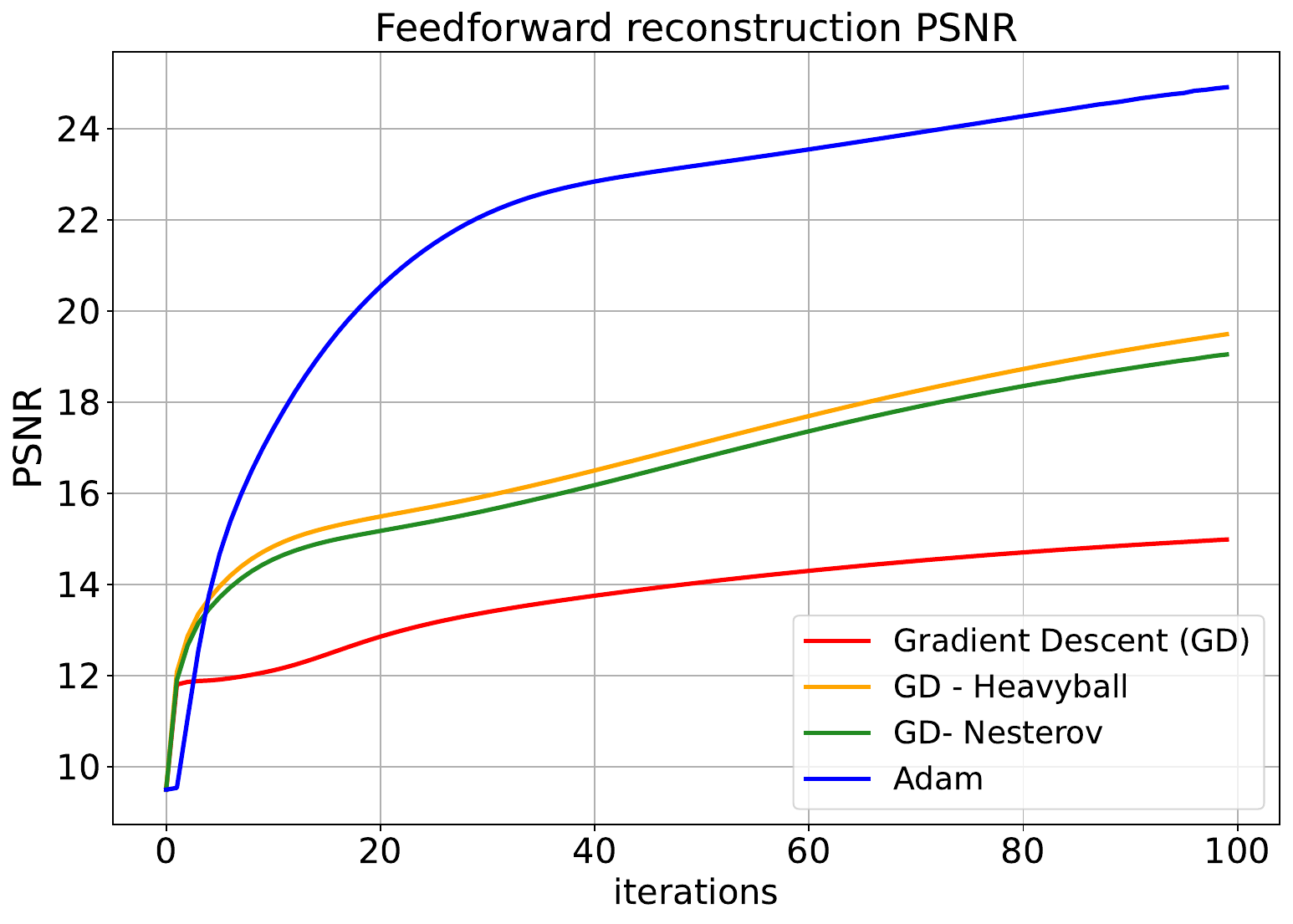}
    \caption{\textbf{Left}: Lifted Bregman Objective decay curves for different optimisers at the gradient step. \textbf{Left}: PSNR curves for the reconstructed images during training. (Both plotted every 100 steps.) }
    \label{fig:LB_different_optimisers}
\end{figure}

To compare the performance of different optimisers, we conduct a small scale denoising experiment where we train the same MLP model on 2,000 noisy images. At each optimisation step, we record the lifted Bregman training objective and the Peak Signal-to-Noise Ratio (PSNR) to evaluate the quality of the reconstructed images. \Cref{fig:LB_different_optimisers} illustrates the effect of the respective acceleration schemes, with Adam consistently demonstrating a more favourable convergence behaviour.

In our experiments, we use Adam optimiser and set the learning rate $\lr$ for both $\theta$ and $\Gaux$ variables at $8 \times 10^{-4}$,  for the deblurring and denoising task. For the inpainting task, we use a smaller learning rate of $1 \times 10^{-4}$. We compare conventional training against lifted Bregman training. In conventional training, the network is trained deterministically via minimising the MSE between the feedforward outputs and the ground truth images, trained for 50,000 optimisation steps also using the Adam optimiser. At each training step, we record the MSE $\ell_2$ reconstruction error and PSNR value.

In both training strategies, the network parameters $\theta$ are initialised identically. For the lifted Bregman training strategy, the auxiliary variables $\Gaux$ can be initialised following different strategies. For instance, one could perform a feed-forward computation with the initial network parameters and save all the intermediate hidden states~\cite{askari2018lifted,zach2019contrastive,wang2023liftedtraining}. Alternatively, one could assign the auxiliary variables with learning targets ``locally'' to begin with. In particular, here we initialise by replicating the noisy input image across all layers. We also implement random initialisation by sampling from a Gaussian distribution but empirically, we found the former initialisation strategy providing improved and more stable results. The intuition is that auxiliary variables are anchored to the data, so initialising them with data-driven warm starts is more effective than drawing from a noise distribution.

\begin{figure}
    \centering
    \includegraphics[width=0.49\linewidth]{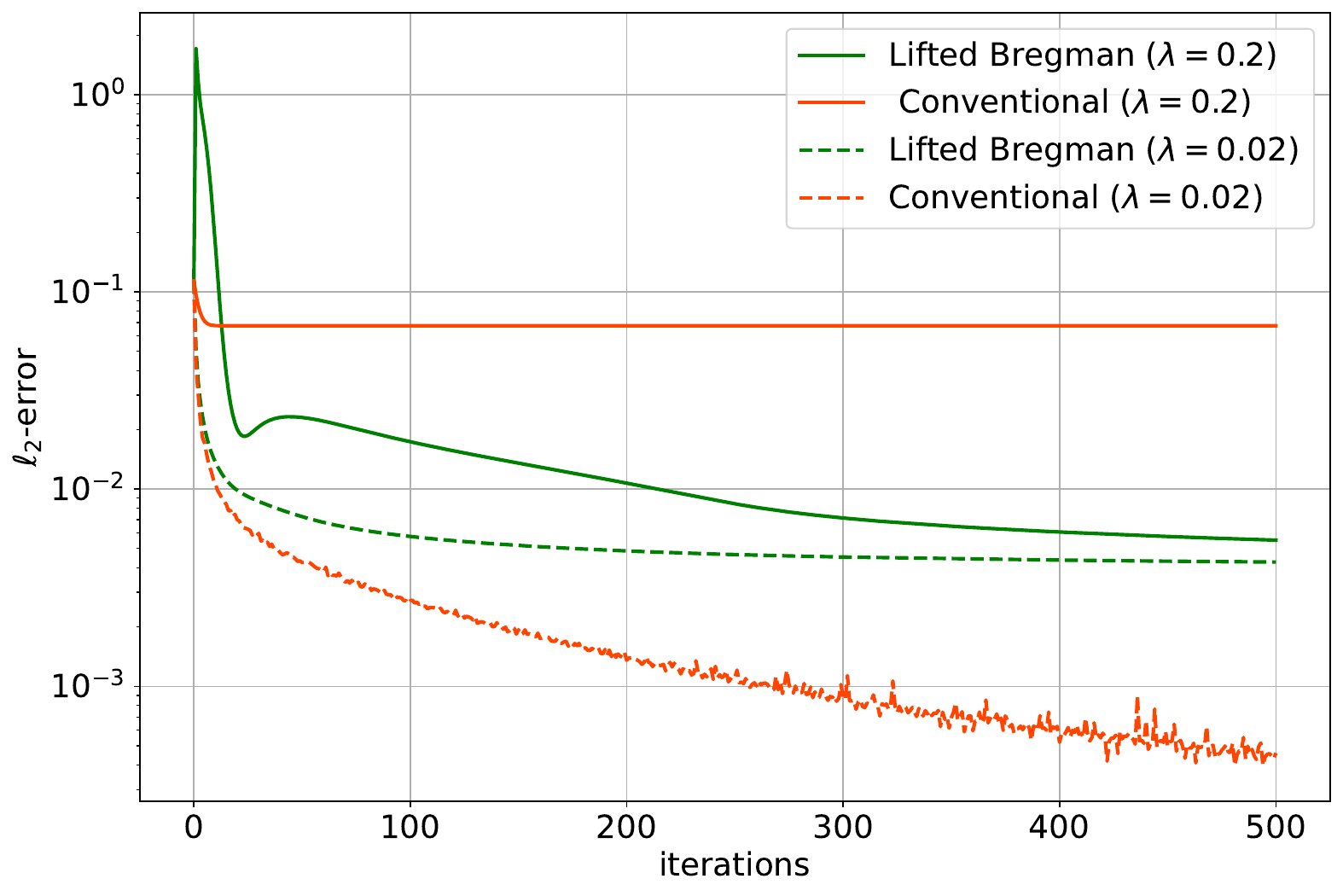}
    \includegraphics[width=0.49\linewidth]{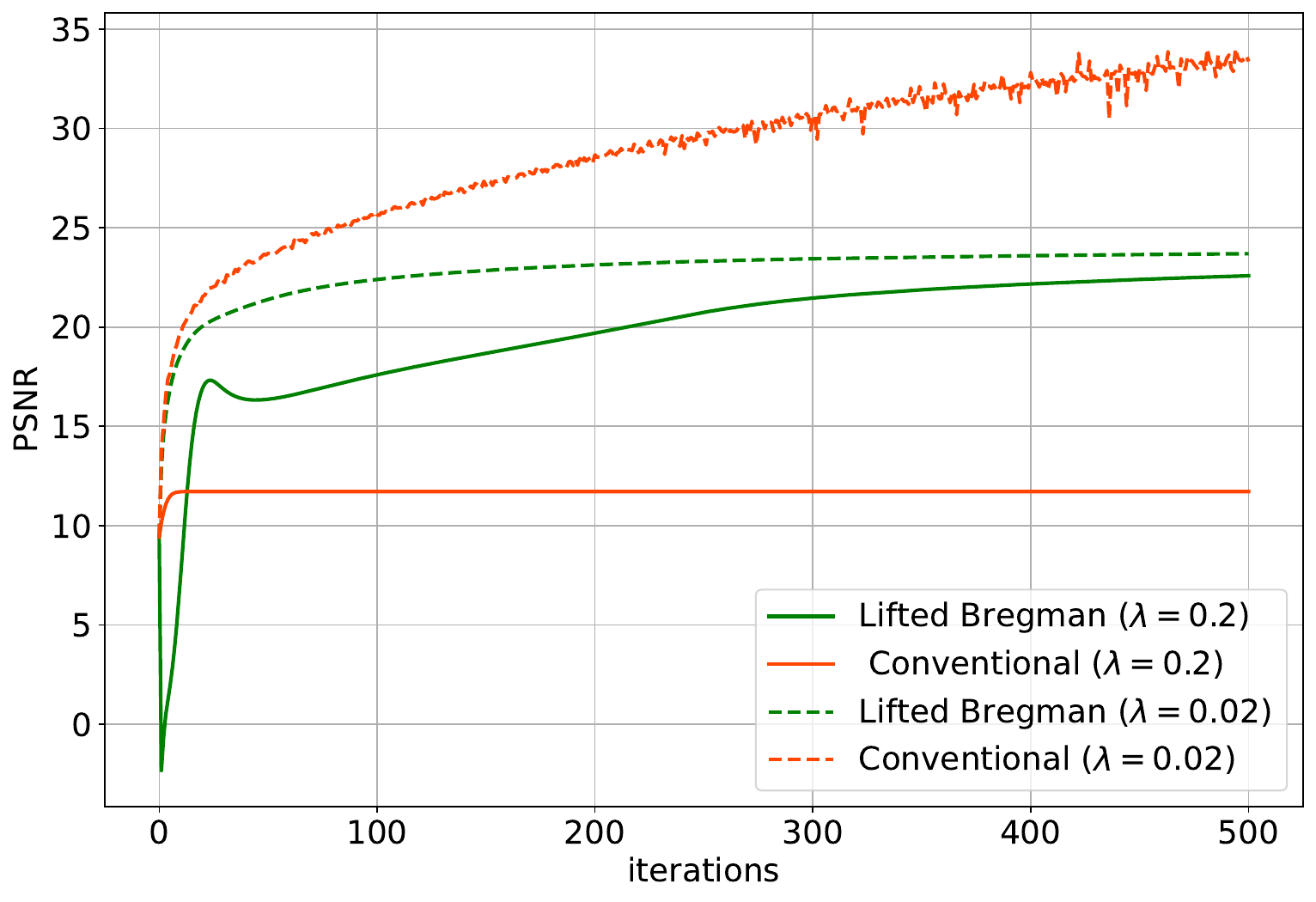}
    \caption{\textbf{(Deblurring $\ell_2$-error and PSNR curves)} Left: $\ell_2$ reconstruction error decay tracked along training steps. Right: PSNR value tracked along training steps. (Results are shown for $\lambda = 0.02$ and $\lambda = 0.2$, respectively.)}
    \label{fig:deblur-comparison}
\end{figure}

\begin{figure}
    \centering
    \includegraphics[width=0.49\linewidth]{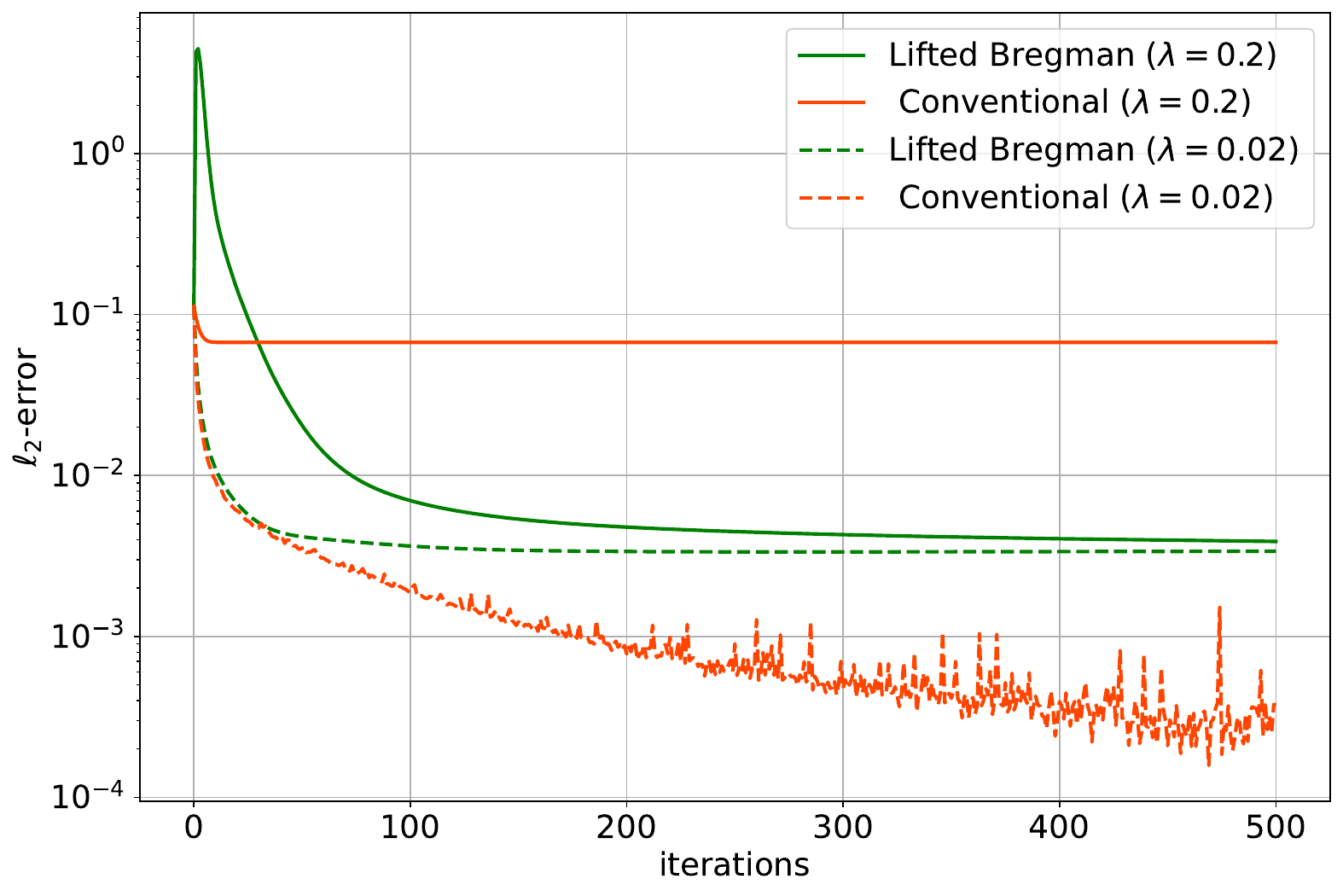}
    \includegraphics[width=0.49\linewidth]{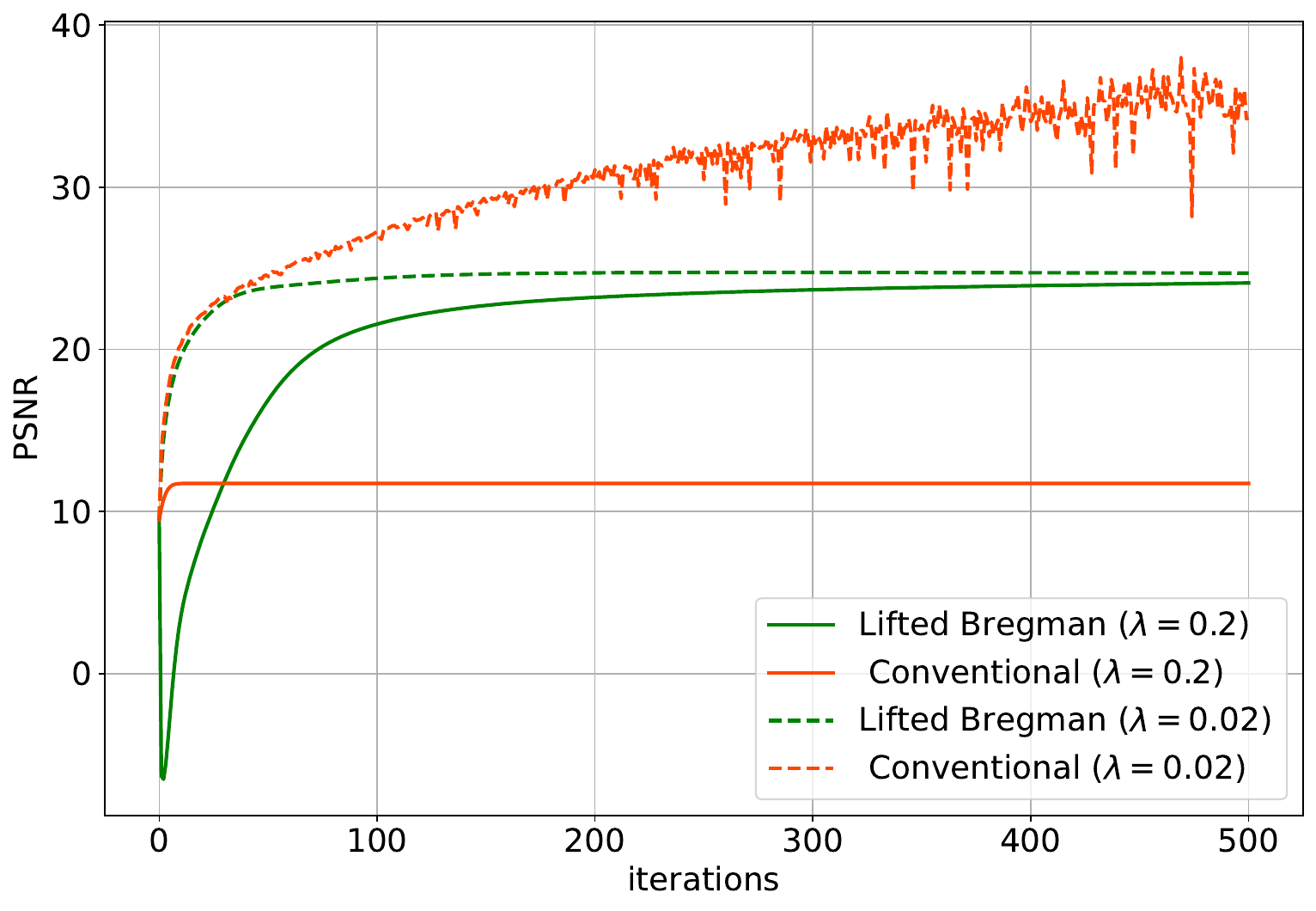}
    \caption{\textbf{(Denoising $\ell_2$-error and PSNR curves)} Left: $\ell_2$ reconstruction error decay tracked along training steps. Right: PSNR value tracked along training steps. (Results are shown for $\lambda = 0.02$ and $\lambda = 0.2$, respectively.) }
    \label{fig:denoise-comparison}
\end{figure}

\begin{figure}
    \centering
    \includegraphics[width=0.49\linewidth]{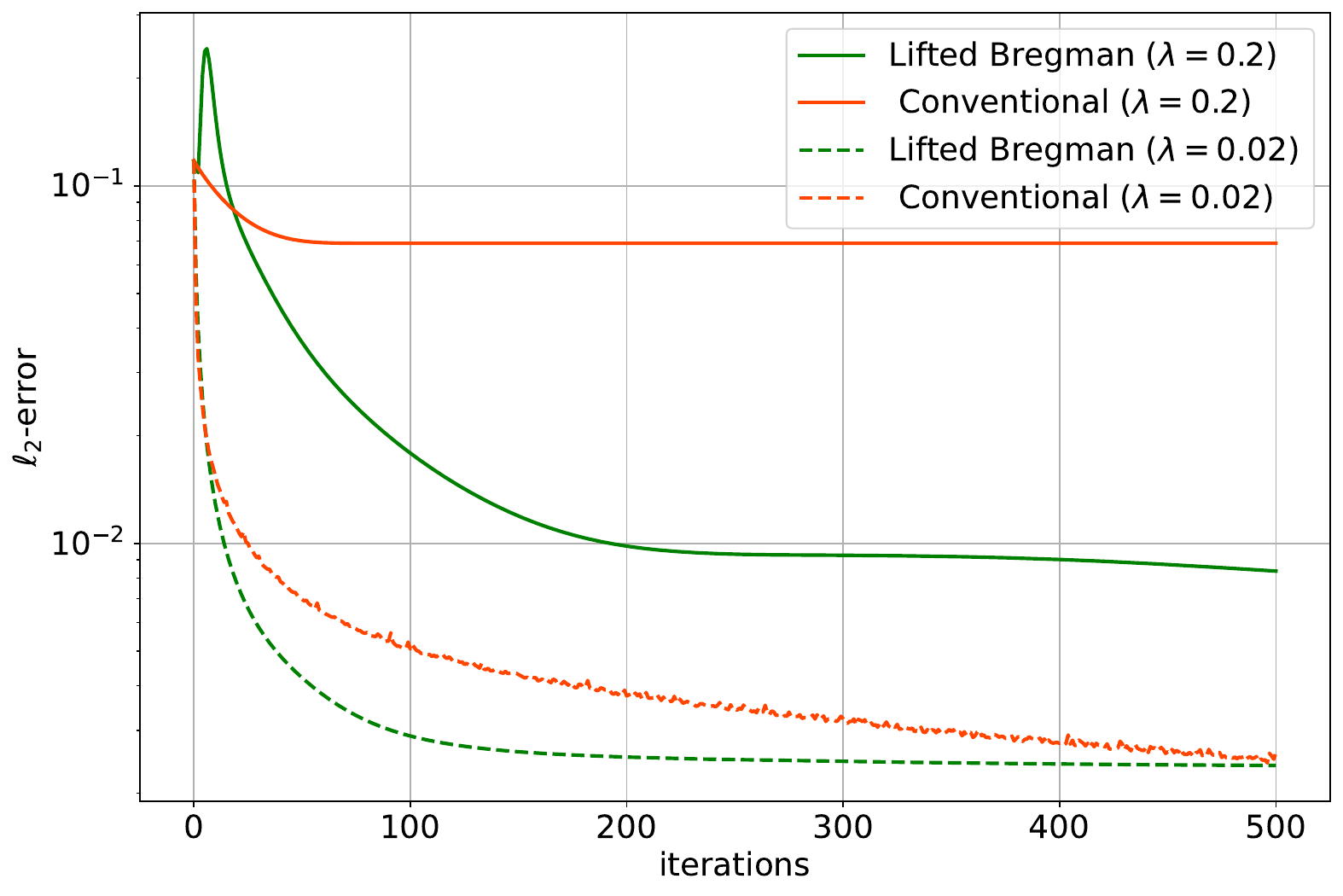}
    \includegraphics[width=0.49\linewidth]{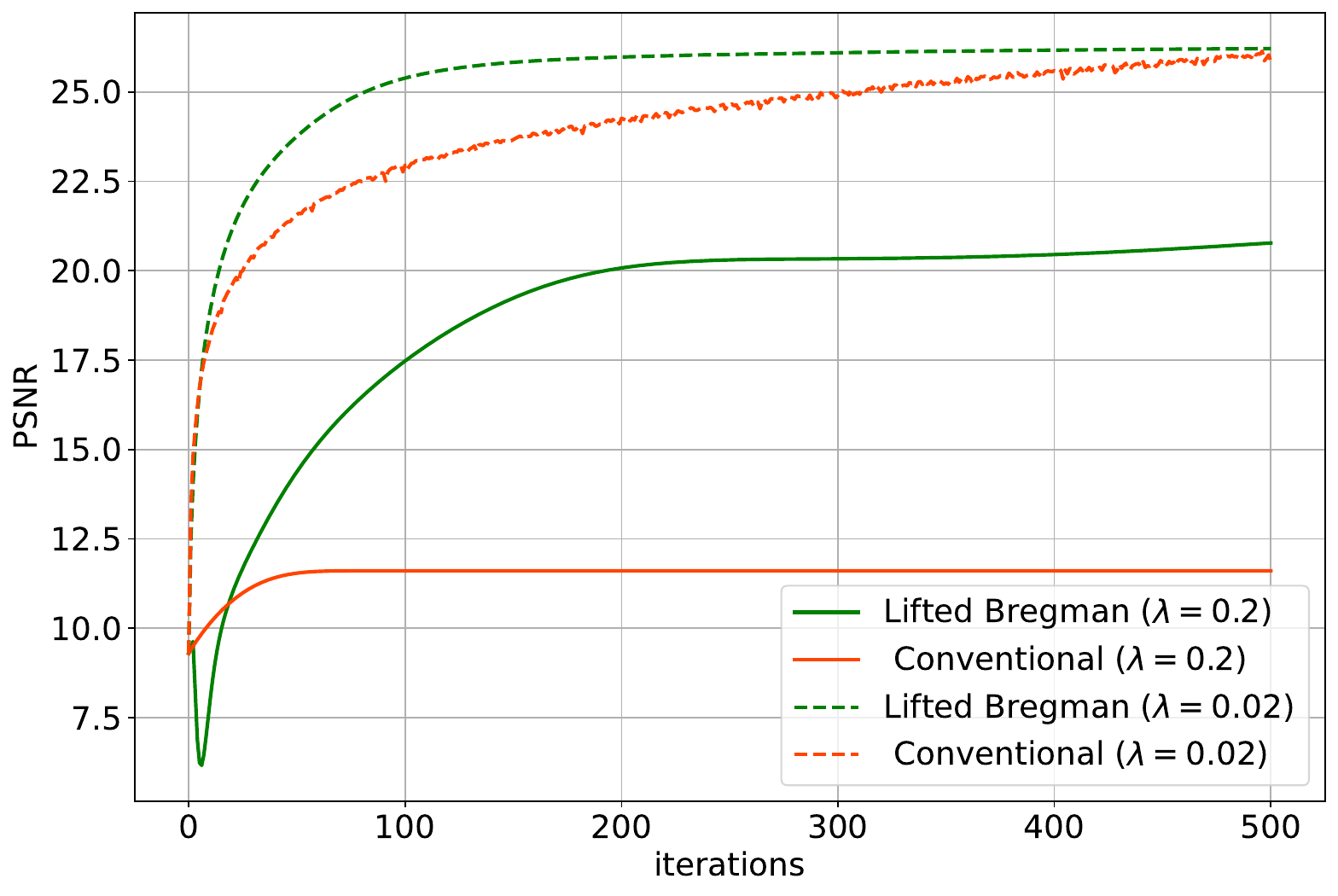}
    \caption{\textbf{(Inpainting $\ell_2$-error and PSNR curves)} Left: $\ell_2$ reconstruction error decay tracked along training steps. Right: PSNR value tracked along training steps. (Results are shown for $\lambda = 0.02$ and $\lambda = 0.2$, respectively.) }
    \label{fig:inpaint-comparison}
\end{figure}

\begin{figure}
\centering

\begin{subfigure}[t]{0.48\textwidth}

\centering
{\hspace{2mm} Training}
\vspace{2mm}

\begin{minipage}{\linewidth}
    \begin{minipage}{0.05\linewidth}
      \centering \rotatebox{90}{$ $}
    \end{minipage}%
    \begin{minipage}{0.95\linewidth}
    \centering {\small Groundtruth}\par
      \includegraphics[width=\linewidth]{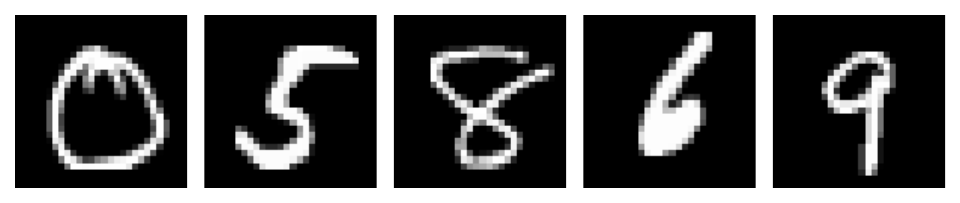}\par\vspace{1mm}
      \vspace{-3mm}
      {\small Observation}\par
      \includegraphics[width=\linewidth]{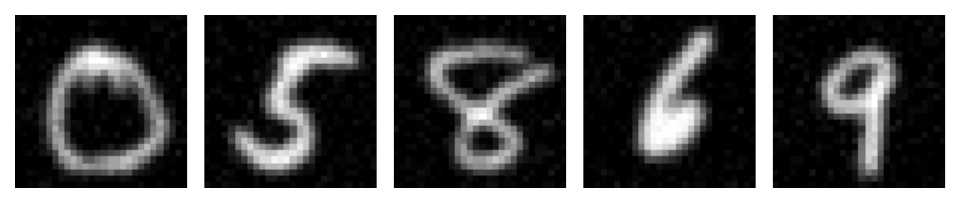}
    \end{minipage}
  \end{minipage}\par\vspace{3mm}

  \begin{minipage}{\linewidth}
    \begin{minipage}{0.05\linewidth}
      \centering \rotatebox{90}{\small $\lambda=0.2$}
    \end{minipage}%
    \begin{minipage}{0.95\linewidth}
    \centering {\small Lifted Bregman}\par
      \includegraphics[width=\linewidth]{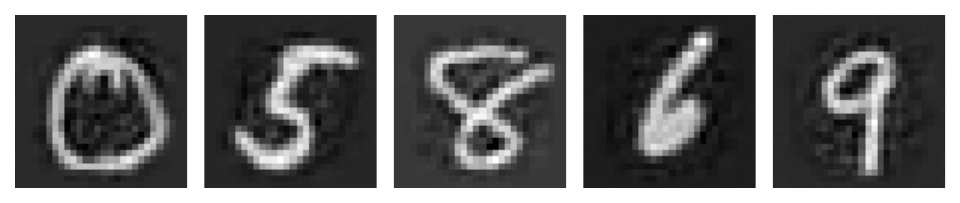}\par\vspace{-1mm}
      \includegraphics[width=\linewidth]{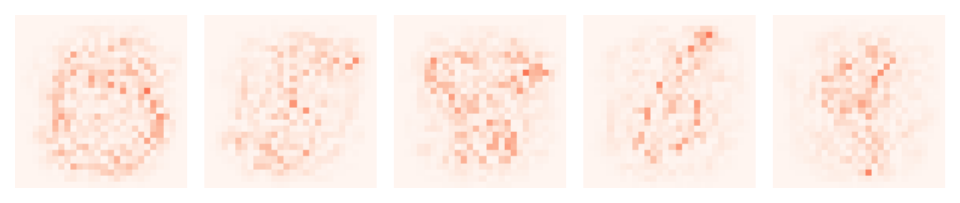}
    \end{minipage}
  \end{minipage}\par\vspace{0.5mm}

  \begin{minipage}{\linewidth}
    \begin{minipage}{0.05\linewidth}
      \centering \rotatebox{90}{\small $\lambda=0.02$}
    \end{minipage}%
    \begin{minipage}{0.95\linewidth}
      \includegraphics[width=\linewidth]{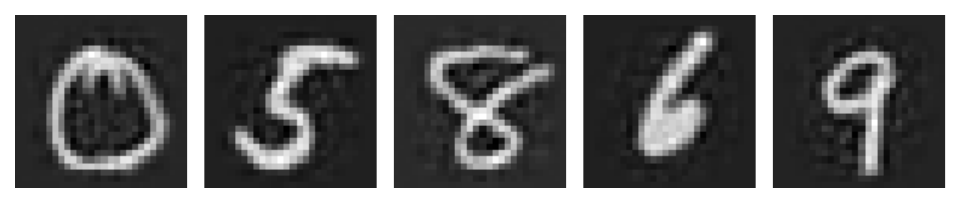}\par\vspace{-1mm}
      \includegraphics[width=\linewidth]{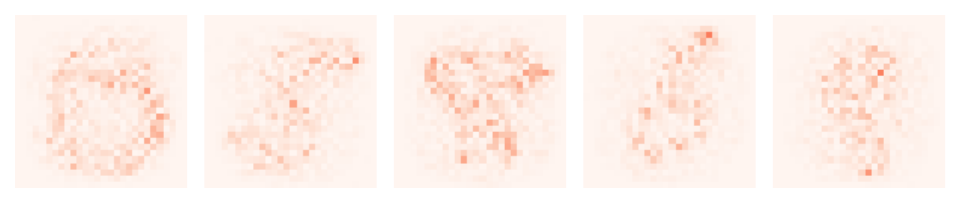}
    \end{minipage}
  \end{minipage}\par\vspace{3mm}

  \begin{minipage}{\linewidth}
    \begin{minipage}{0.05\linewidth}
      \centering \rotatebox{90}{\small $\lambda=0.2$}
    \end{minipage}%
    \begin{minipage}{0.95\linewidth}
    \centering {\small Conventional}\par
      \includegraphics[width=\linewidth]{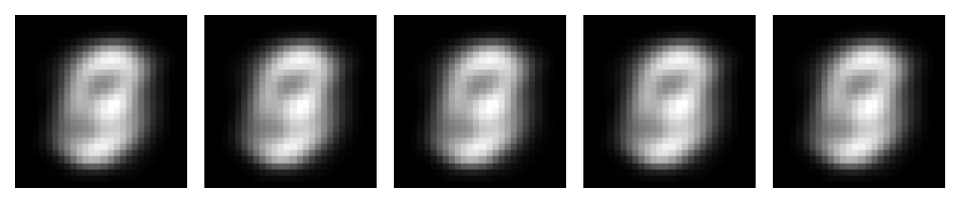}\par\vspace{-1mm}
      \includegraphics[width=\linewidth]{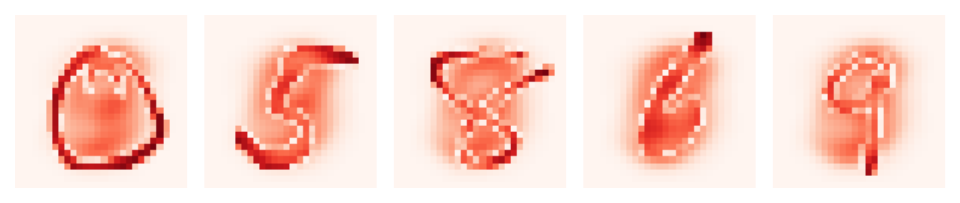}
    \end{minipage}
  \end{minipage}\par\vspace{0.5mm}
  
  \begin{minipage}{\linewidth}
    \begin{minipage}{0.05\linewidth}
      \centering \rotatebox{90}{\small $\lambda=0.02$}
    \end{minipage}%
    \begin{minipage}{0.95\linewidth}
      \includegraphics[width=\linewidth]{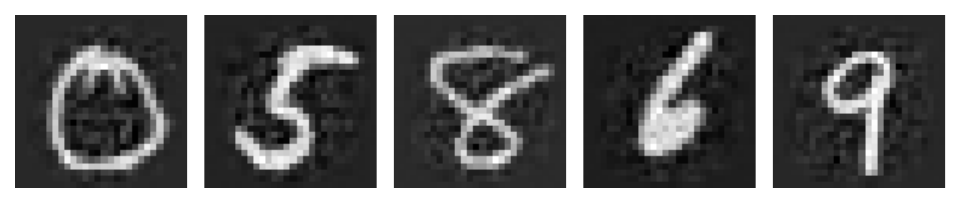}\par\vspace{-1mm}
      \includegraphics[width=\linewidth]{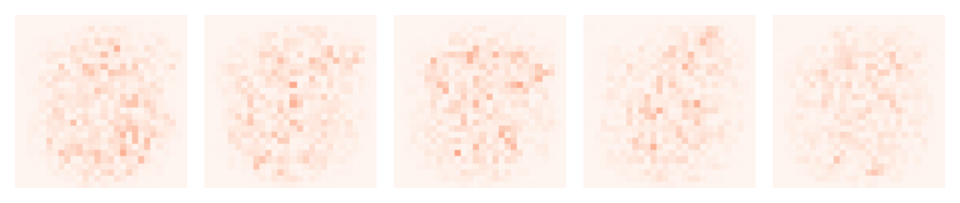}
    \end{minipage}
  \end{minipage}
\end{subfigure}\hfill
\begin{subfigure}[t]{0.48\textwidth}
 
\centering
{\hspace{2mm} Validation}
\vspace{2mm}

\begin{minipage}{\linewidth}
    \begin{minipage}{0.05\linewidth}
      \centering \rotatebox{90}{$ $}
    \end{minipage}%
    \begin{minipage}{0.95\linewidth}
    \centering {\small Groundtruth}\par
      \includegraphics[width=\linewidth]{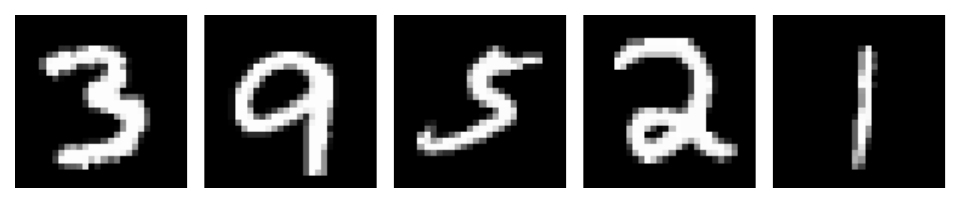}\par\vspace{1mm}
      \vspace{-3mm}
      {\small Observation}\par
      \includegraphics[width=\linewidth]{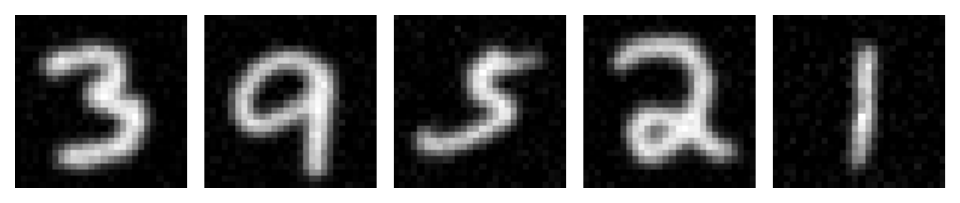}
    \end{minipage}
  \end{minipage}\par\vspace{3mm}

  \begin{minipage}{\linewidth}
    \begin{minipage}{0.05\linewidth}
      \centering \rotatebox{90}{\small $\lambda=0.2$}
    \end{minipage}%
    \begin{minipage}{0.95\linewidth}
    \centering {\small Lifted Bregman}\par
      \includegraphics[width=\linewidth]{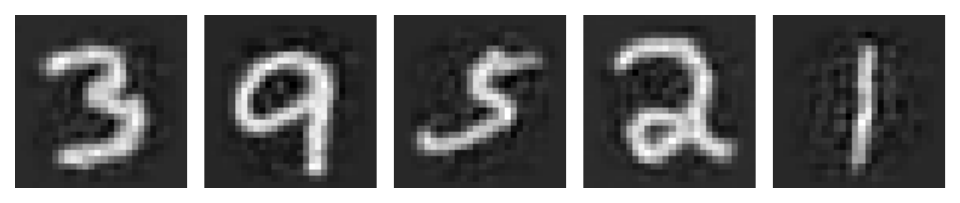}\par\vspace{-1mm}
      \includegraphics[width=\linewidth]{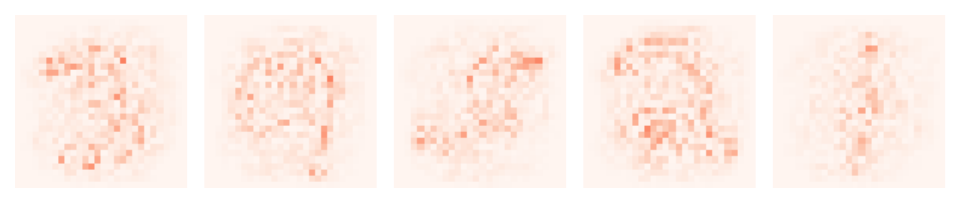}
    \end{minipage}
  \end{minipage}\par\vspace{0.5mm}
  
  \begin{minipage}{\linewidth}
    \begin{minipage}{0.05\linewidth}
      \centering \rotatebox{90}{\small $\lambda=0.02$}
    \end{minipage}%
    \begin{minipage}{0.95\linewidth}
      \includegraphics[width=\linewidth]{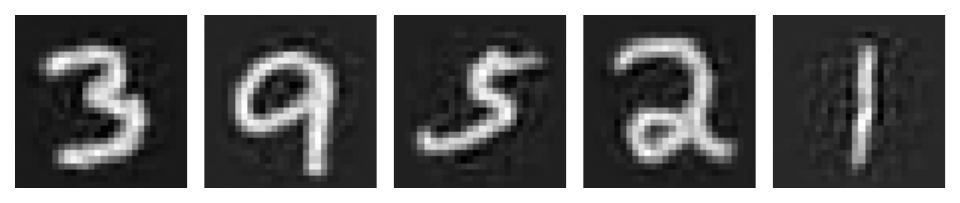}\par\vspace{-1mm}
      \includegraphics[width=\linewidth]{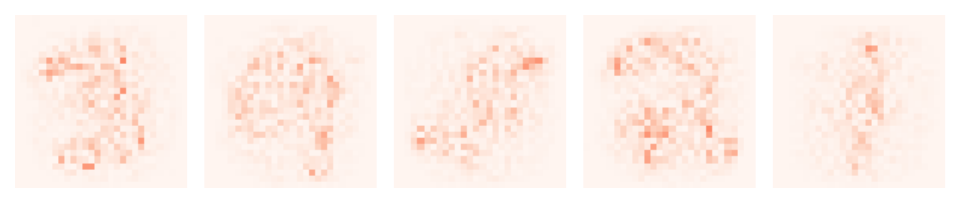}
    \end{minipage}
  \end{minipage}\par\vspace{3mm}

  \begin{minipage}{\linewidth}
    \begin{minipage}{0.05\linewidth}
      \centering \rotatebox{90}{\small $\lambda=0.2$}
    \end{minipage}%
    \begin{minipage}{0.95\linewidth}
    \centering {\small Conventional}\par
      \includegraphics[width=\linewidth]{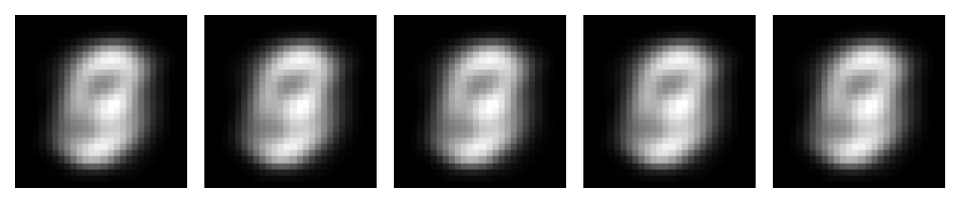}\par\vspace{-1mm}
      \includegraphics[width=\linewidth]{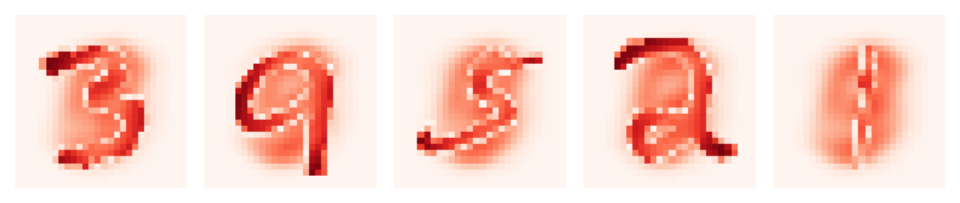}
    \end{minipage}
  \end{minipage}\par\vspace{0.5mm}
  
  \begin{minipage}{\linewidth}
    \begin{minipage}{0.05\linewidth}
      \centering \rotatebox{90}{\small $\lambda=0.02$}
    \end{minipage}%
    \begin{minipage}{0.95\linewidth}
      \includegraphics[width=\linewidth]{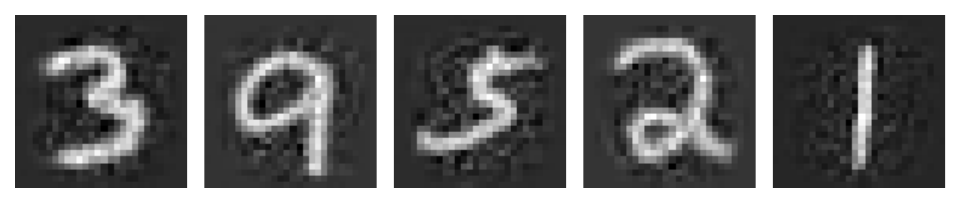}\par\vspace{-1mm}
      \includegraphics[width=\linewidth]{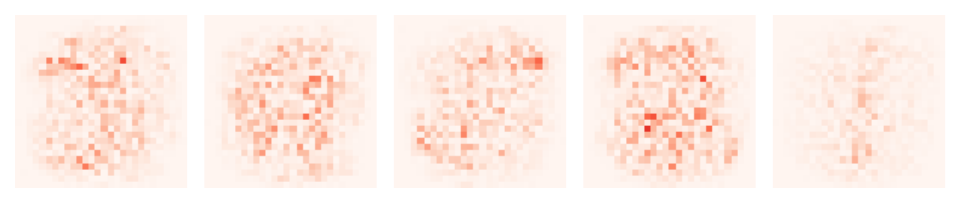}
    \end{minipage}
  \end{minipage}
\end{subfigure}
    \caption{\textbf{(Deblurred image visualisations)} Left: Deblurred sample images from the training dataset and error maps for both strategies. Right: Deblurred sample images from the validation dataset and error maps for both strategies. (Results are shown for $\lambda = 0.02$ and $\lambda = 0.2$, respectively.)}
    \label{fig:deblur-visualisation}
\end{figure}

\begin{figure}
\centering

\begin{subfigure}[t]{0.48\textwidth}

\centering
{\hspace{2mm} Training}
\vspace{2mm}

\begin{minipage}{\linewidth}
    \begin{minipage}{0.05\linewidth}
      \centering \rotatebox{90}{$ $}
    \end{minipage}%
    \begin{minipage}{0.95\linewidth}
    \centering {\small Groundtruth}\par
      \includegraphics[width=\linewidth]{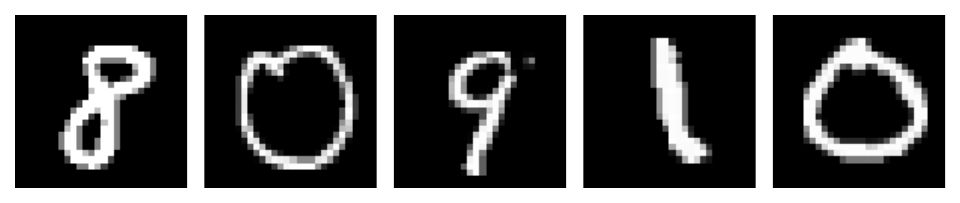}\par\vspace{1mm}
      \vspace{-3mm}
      {\small Observation}\par
      \includegraphics[width=\linewidth]{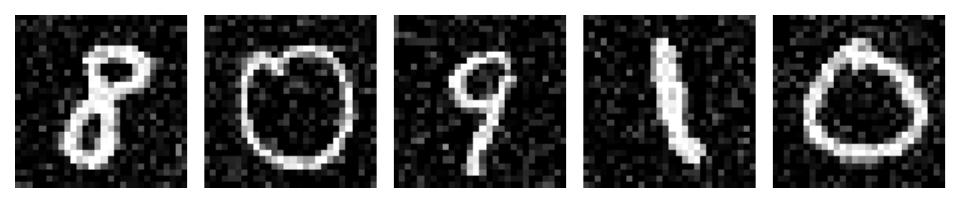}
    \end{minipage}
  \end{minipage}\par\vspace{3mm}

  \begin{minipage}{\linewidth}
    \begin{minipage}{0.05\linewidth}
      \centering \rotatebox{90}{\small $\lambda=0.2$}
    \end{minipage}%
    \begin{minipage}{0.95\linewidth}
    \centering {\small Lifted Bregman}\par
      \includegraphics[width=\linewidth]{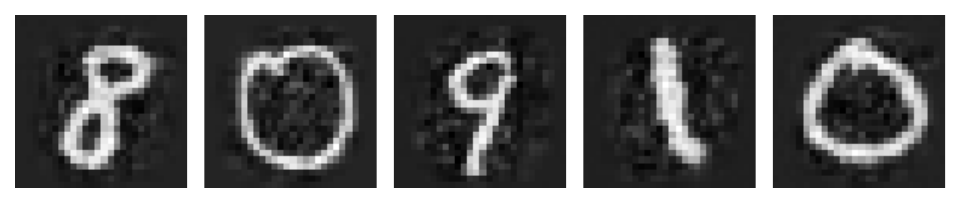}\par\vspace{-1mm}
      \includegraphics[width=\linewidth]{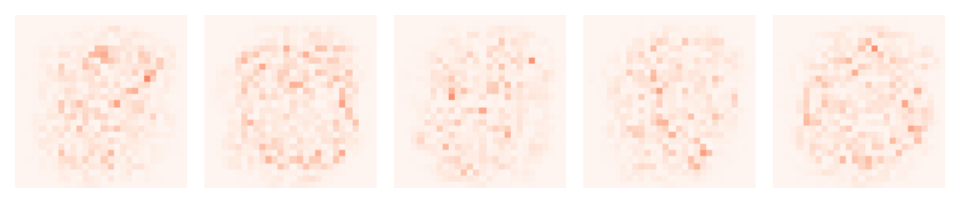}
    \end{minipage}
  \end{minipage}\par\vspace{0.5mm}

  \begin{minipage}{\linewidth}
    \begin{minipage}{0.05\linewidth}
      \centering \rotatebox{90}{\small $\lambda=0.02$}
    \end{minipage}%
    \begin{minipage}{0.95\linewidth}
      \includegraphics[width=\linewidth]{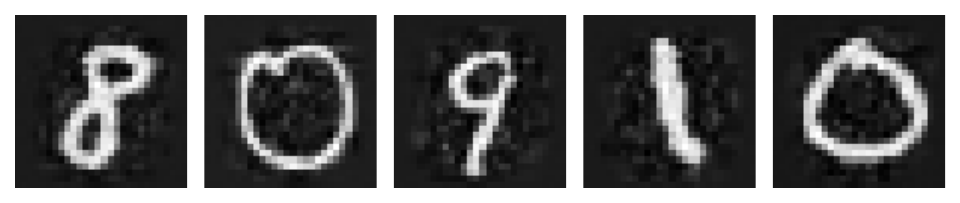}\par\vspace{-1mm}
      \includegraphics[width=\linewidth]{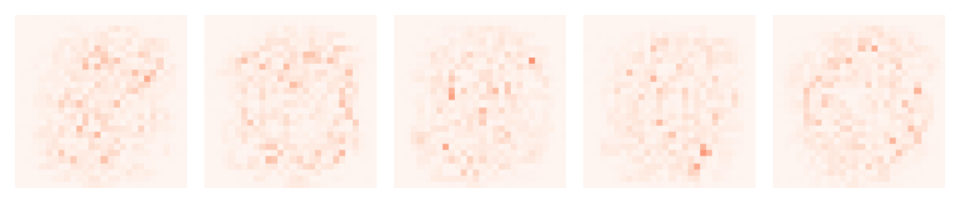}
    \end{minipage}
  \end{minipage}\par\vspace{3mm}

  \begin{minipage}{\linewidth}
    \begin{minipage}{0.05\linewidth}
      \centering \rotatebox{90}{\small $\lambda=0.2$}
    \end{minipage}%
    \begin{minipage}{0.95\linewidth}
    \centering {\small Conventional}\par
      \includegraphics[width=\linewidth]{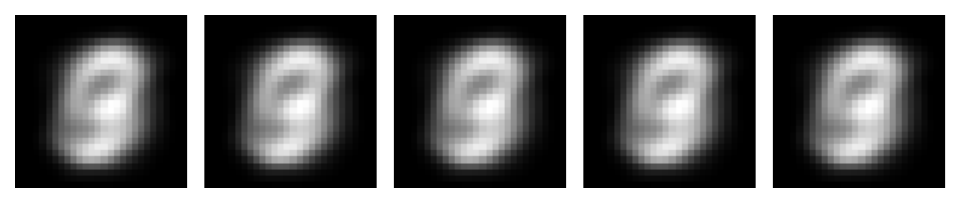}\par\vspace{-1mm}
      \includegraphics[width=\linewidth]{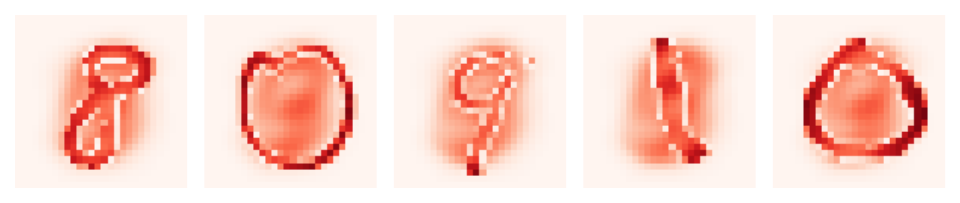}
    \end{minipage}
  \end{minipage}\par\vspace{0.5mm}
  
  \begin{minipage}{\linewidth}
    \begin{minipage}{0.05\linewidth}
      \centering \rotatebox{90}{\small $\lambda=0.02$}
    \end{minipage}%
    \begin{minipage}{0.95\linewidth}
      \includegraphics[width=\linewidth]{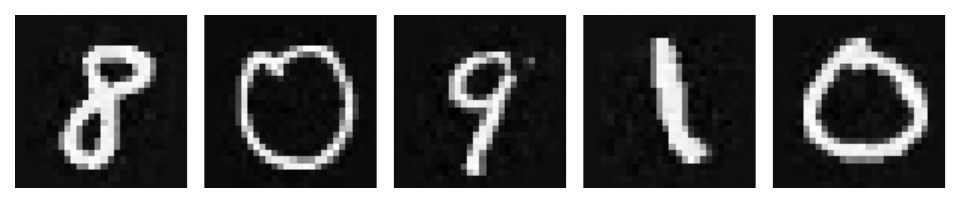}\par\vspace{-1mm}
      \includegraphics[width=\linewidth]{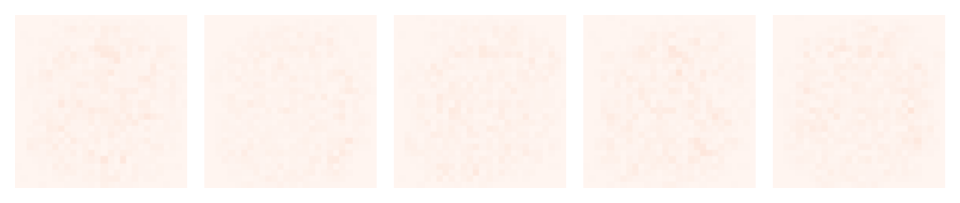}
    \end{minipage}
  \end{minipage}
\end{subfigure}\hfill
\begin{subfigure}[t]{0.48\textwidth}
 
\centering
{\hspace{2mm} Validation}
\vspace{2mm}

\begin{minipage}{\linewidth}
    \begin{minipage}{0.05\linewidth}
      \centering \rotatebox{90}{$ $}
    \end{minipage}%
    \begin{minipage}{0.95\linewidth}
    \centering {\small Groundtruth}\par
      \includegraphics[width=\linewidth]{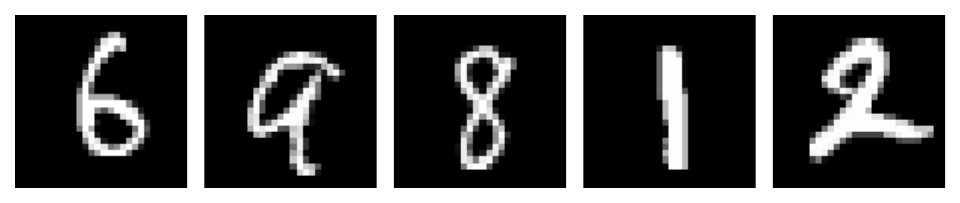}\par\vspace{1mm}
      \vspace{-3mm}
      {\small Observation}\par
      \includegraphics[width=\linewidth]{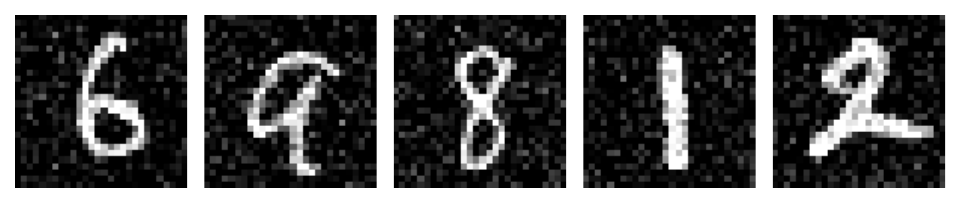}
    \end{minipage}
  \end{minipage}\par\vspace{3mm}

  \begin{minipage}{\linewidth}
    \begin{minipage}{0.05\linewidth}
      \centering \rotatebox{90}{\small $\lambda=0.2$}
    \end{minipage}%
    \begin{minipage}{0.95\linewidth}
    \centering {\small Lifted Bregman}\par
      \includegraphics[width=\linewidth]{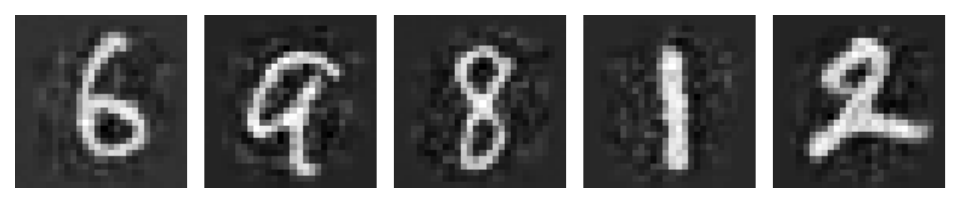}\par\vspace{-1mm}
      \includegraphics[width=\linewidth]{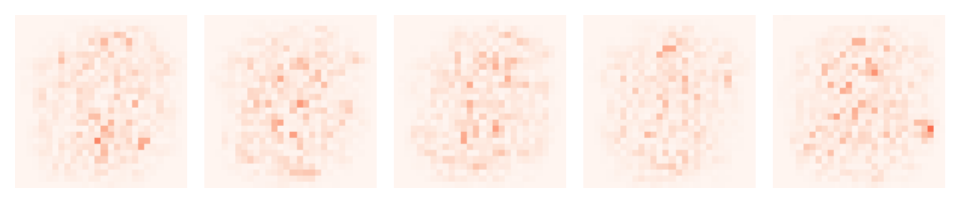}
    \end{minipage}
  \end{minipage}\par\vspace{0.5mm}
  
  \begin{minipage}{\linewidth}
    \begin{minipage}{0.05\linewidth}
      \centering \rotatebox{90}{\small $\lambda=0.02$}
    \end{minipage}%
    \begin{minipage}{0.95\linewidth}
      \includegraphics[width=\linewidth]{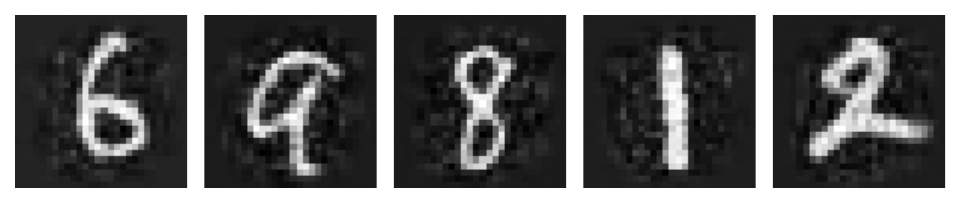}\par\vspace{-1mm}
      \includegraphics[width=\linewidth]{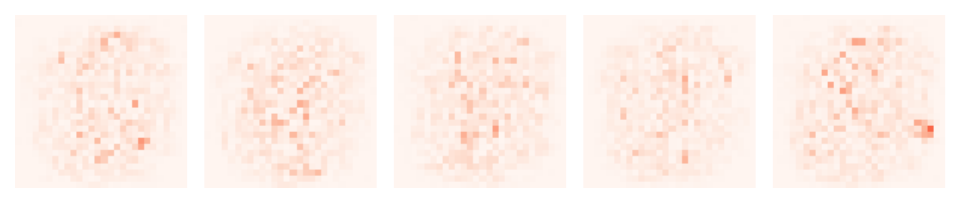}
    \end{minipage}
  \end{minipage}\par\vspace{3mm}

  \begin{minipage}{\linewidth}
    \begin{minipage}{0.05\linewidth}
      \centering \rotatebox{90}{\small $\lambda=0.2$}
    \end{minipage}%
    \begin{minipage}{0.95\linewidth}
    \centering {\small Conventional}\par
      \includegraphics[width=\linewidth]{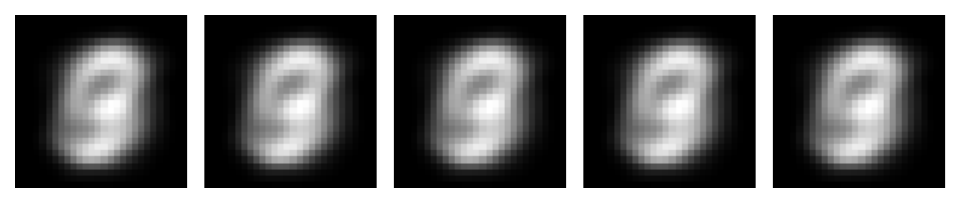}\par\vspace{-1mm}
      \includegraphics[width=\linewidth]{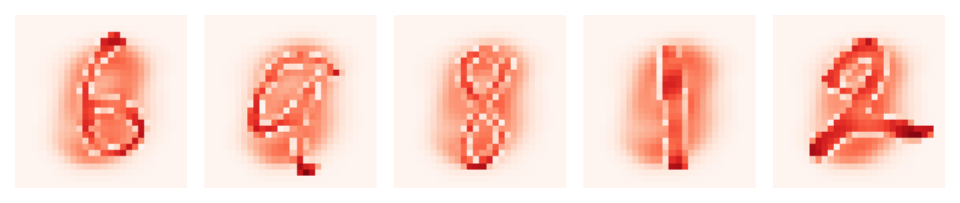}
    \end{minipage}
  \end{minipage}\par\vspace{0.5mm}
  
  \begin{minipage}{\linewidth}
    \begin{minipage}{0.05\linewidth}
      \centering \rotatebox{90}{\small $\lambda=0.02$}
    \end{minipage}%
    \begin{minipage}{0.95\linewidth}
      \includegraphics[width=\linewidth]{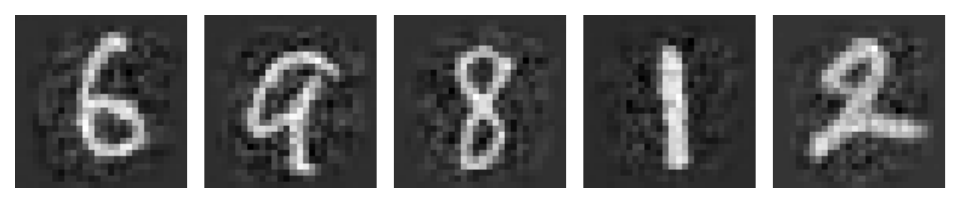}\par\vspace{-1mm}
      \includegraphics[width=\linewidth]{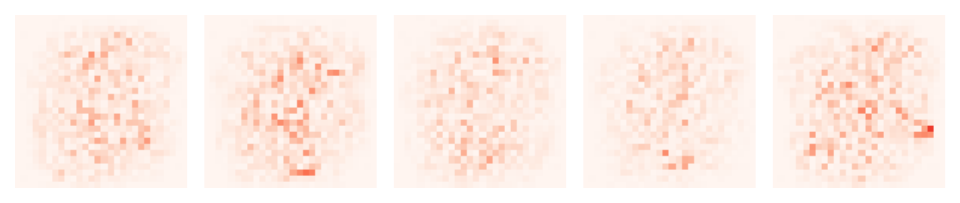}
    \end{minipage}
  \end{minipage}
\end{subfigure}
    \caption{\textbf{(Denoised image visualisations)} Left: Denoised sample images from the training dataset and error maps for both strategies. Right: Denoised sample images from the validation dataset and error maps for both strategies. (Results are shown for $\lambda = 0.02$ and $\lambda = 0.2$, respectively.)}
    \label{fig:denoise-visualisation}
\end{figure}

\begin{figure}
\centering

\begin{subfigure}[t]{0.48\textwidth}
 
\centering
{\hspace{2mm} Training}
\vspace{2mm}

\begin{minipage}{\linewidth}
    \begin{minipage}{0.05\linewidth}
      \centering \rotatebox{90}{$ $}
    \end{minipage}%
    \begin{minipage}{0.95\linewidth}
    \centering {\small Groundtruth}\par
      \includegraphics[width=\linewidth]{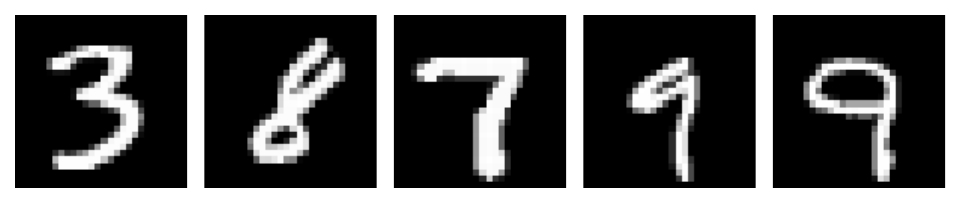}\par\vspace{1mm}
      \vspace{-3mm}
      {\small Observation}\par
      \includegraphics[width=\linewidth]{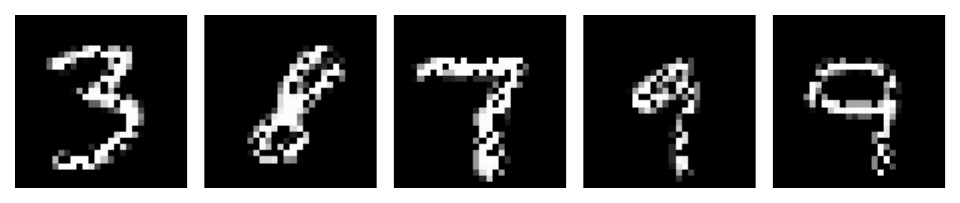}
    \end{minipage}
  \end{minipage}\par\vspace{3mm}

  \begin{minipage}{\linewidth}
    \begin{minipage}{0.05\linewidth}
      \centering \rotatebox{90}{\small $\lambda=0.2$}
    \end{minipage}%
    \begin{minipage}{0.95\linewidth}
    \centering {\small Lifted Bregman}\par
      \includegraphics[width=\linewidth]{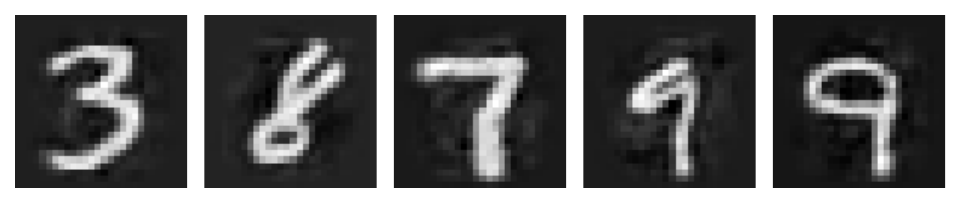}\par\vspace{-1mm}
      \includegraphics[width=\linewidth]{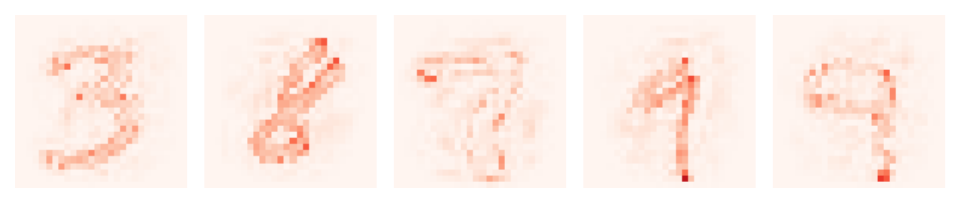}
    \end{minipage}
  \end{minipage}\par\vspace{0.5mm}

  \begin{minipage}{\linewidth}
    \begin{minipage}{0.05\linewidth}
      \centering \rotatebox{90}{\small $\lambda=0.02$}
    \end{minipage}%
    \begin{minipage}{0.95\linewidth}
      \includegraphics[width=\linewidth]{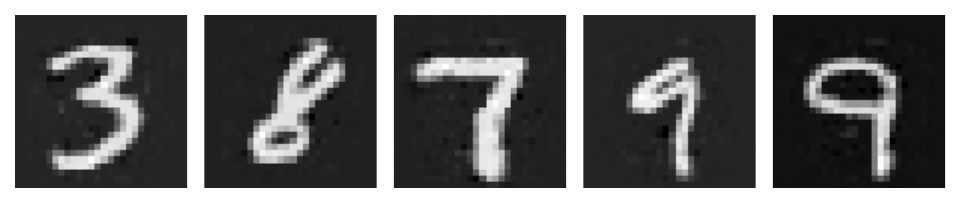}\par\vspace{-1mm}
      \includegraphics[width=\linewidth]{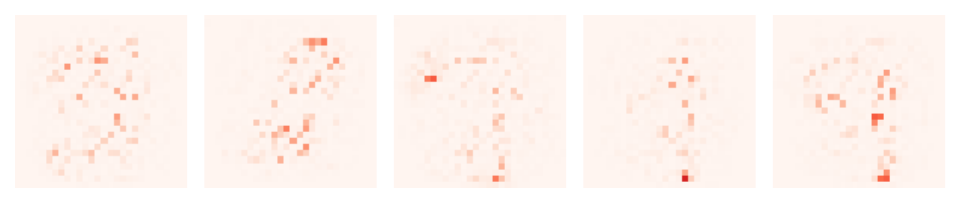}
    \end{minipage}
  \end{minipage}\par\vspace{3mm}

  \begin{minipage}{\linewidth}
    \begin{minipage}{0.05\linewidth}
      \centering \rotatebox{90}{\small $\lambda=0.2$}
    \end{minipage}%
    \begin{minipage}{0.95\linewidth}
    \centering {\small Conventional}\par
      \includegraphics[width=\linewidth]{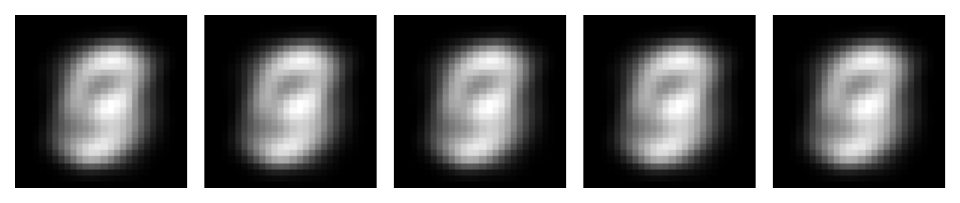}\par\vspace{-1mm}
      \includegraphics[width=\linewidth]{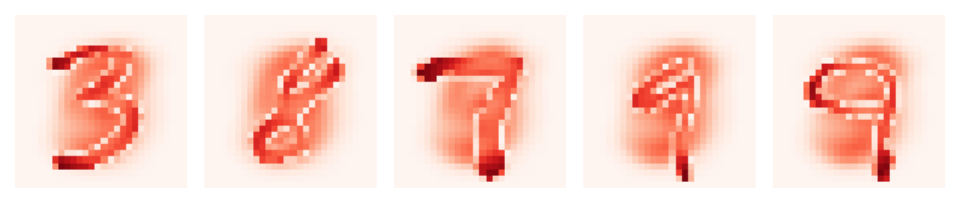}
    \end{minipage}
  \end{minipage}\par\vspace{0.5mm}
  
  \begin{minipage}{\linewidth}
    \begin{minipage}{0.05\linewidth}
      \centering \rotatebox{90}{\small $\lambda=0.02$}
    \end{minipage}%
    \begin{minipage}{0.95\linewidth}
      \includegraphics[width=\linewidth]{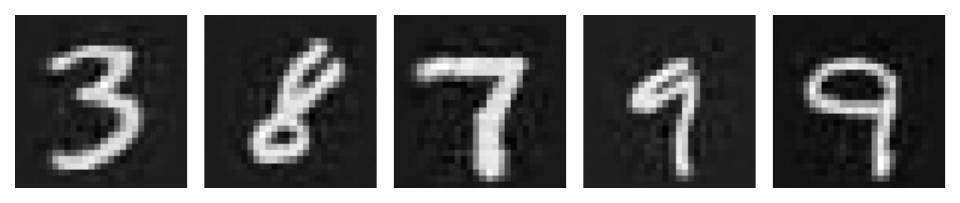}\par\vspace{-1mm}
      \includegraphics[width=\linewidth]{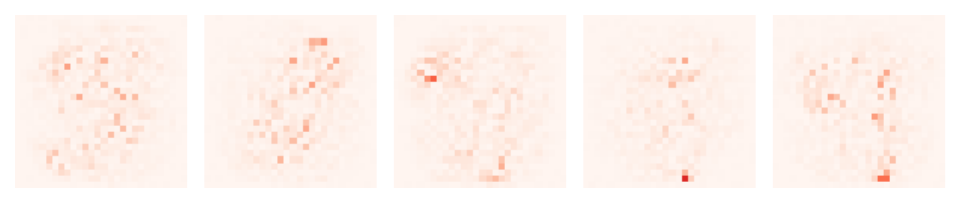}
    \end{minipage}
  \end{minipage}
\end{subfigure}\hfill
\begin{subfigure}[t]{0.48\textwidth} 

\centering
{\hspace{2mm} Validation}
\vspace{2mm}

\begin{minipage}{\linewidth}
    \begin{minipage}{0.05\linewidth}
      \centering \rotatebox{90}{$ $}
    \end{minipage}%
    \begin{minipage}{0.95\linewidth}
    \centering {\small Groundtruth}\par
      \includegraphics[width=\linewidth]{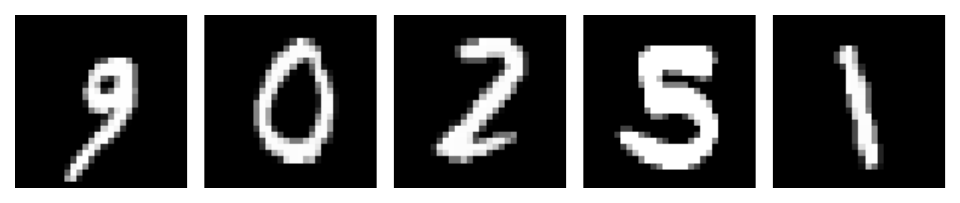}\par\vspace{1mm}
      \vspace{-3mm}
      {\small Observation}\par
      \includegraphics[width=\linewidth]{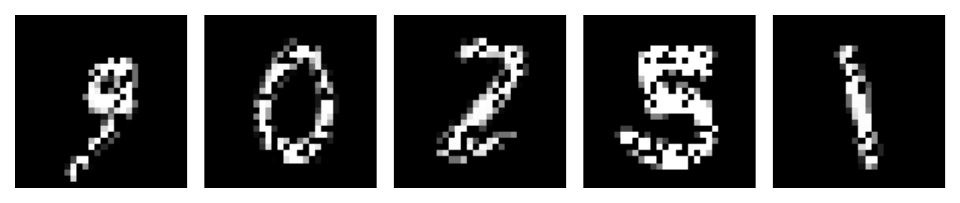}
    \end{minipage}
  \end{minipage}\par\vspace{3mm}

  \begin{minipage}{\linewidth}
    \begin{minipage}{0.05\linewidth}
      \centering \rotatebox{90}{\small $\lambda=0.2$}
    \end{minipage}%
    \begin{minipage}{0.95\linewidth}
    \centering {\small Lifted Bregman}\par
      \includegraphics[width=\linewidth]{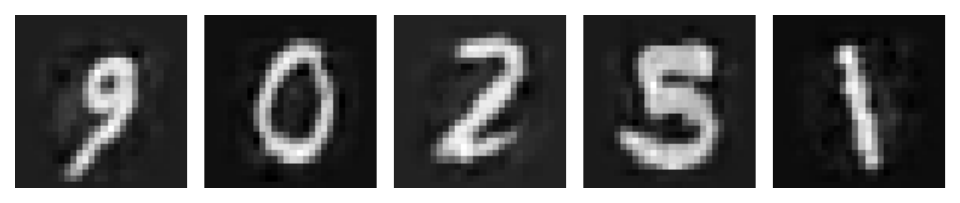}\par\vspace{-1mm}
      \includegraphics[width=\linewidth]{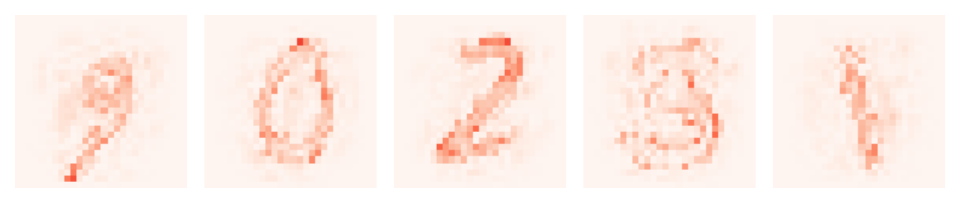}
    \end{minipage}
  \end{minipage}\par\vspace{0.5mm}
  
  \begin{minipage}{\linewidth}
    \begin{minipage}{0.05\linewidth}
      \centering \rotatebox{90}{\small $\lambda=0.02$}
    \end{minipage}%
    \begin{minipage}{0.95\linewidth}
      \includegraphics[width=\linewidth]{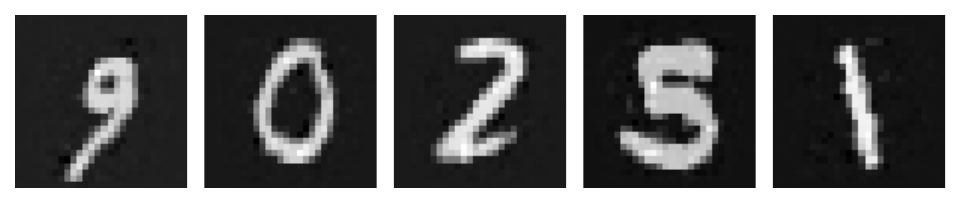}\par\vspace{-1mm}
      \includegraphics[width=\linewidth]{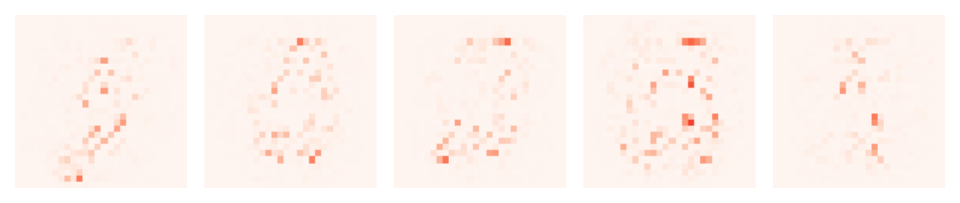}
    \end{minipage}
  \end{minipage}\par\vspace{3mm}

  \begin{minipage}{\linewidth}
    \begin{minipage}{0.05\linewidth}
      \centering \rotatebox{90}{\small $\lambda=0.2$}
    \end{minipage}%
    \begin{minipage}{0.95\linewidth}
    \centering {\small Conventional}\par
      \includegraphics[width=\linewidth]{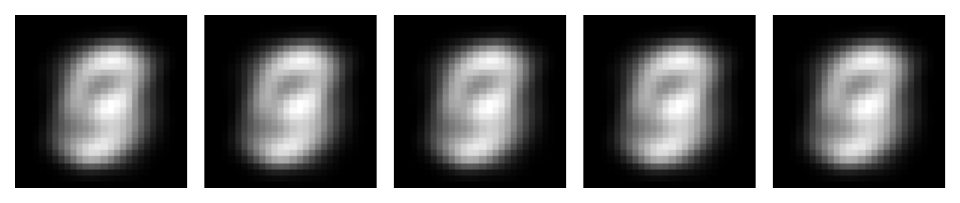}\par\vspace{-1mm}
      \includegraphics[width=\linewidth]{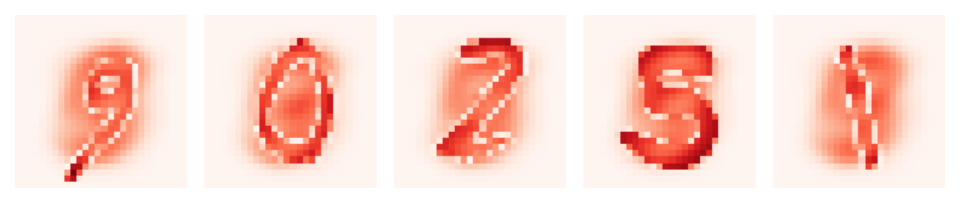}
    \end{minipage}
  \end{minipage}\par\vspace{0.5mm}
  
  \begin{minipage}{\linewidth}
    \begin{minipage}{0.05\linewidth}
      \centering \rotatebox{90}{\small $\lambda=0.02$}
    \end{minipage}%
    \begin{minipage}{0.95\linewidth}
      \includegraphics[width=\linewidth]{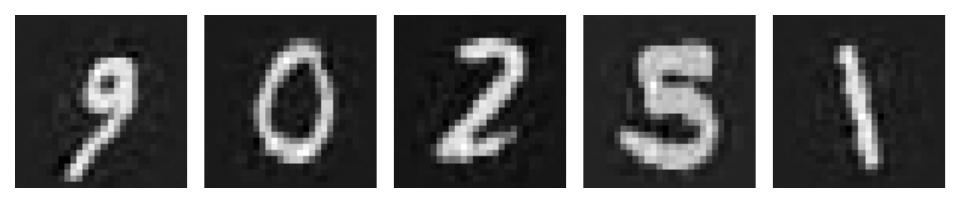}\par\vspace{-1mm}
      \includegraphics[width=\linewidth]{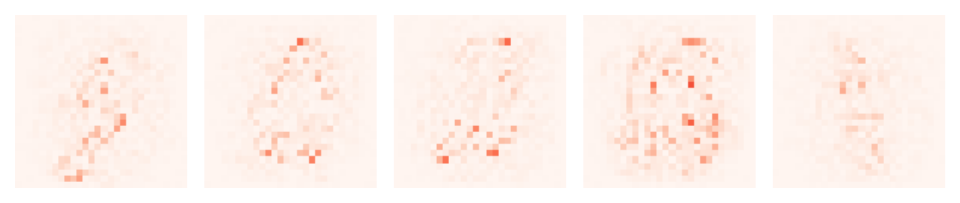}
    \end{minipage}
  \end{minipage}
\end{subfigure}
\caption{\textbf{(Inpainted image visualisations)} Left: Inpainted sample images from the training dataset and error maps for both strategies. Right: Inpainted sample images from the validation dataset and error maps for both strategies. (Results are shown for $\lambda = 0.02$ and $\lambda = 0.2$, respectively.)}
\label{fig:inpaint-visualisation}
\end{figure}

\Cref{fig:deblur-comparison}, \Cref{fig:denoise-comparison}, and \Cref{fig:inpaint-comparison} present the results for the deblurring, denoising, and inpainting tasks, respectively. The plots on the left show the decay of the $\ell_2$ reconstruction error over training steps under the two settings of $\lambda=2 \times 10^{-2}$ and $\lambda=2 \times 10^{-1}$ , while the plots on the right compare the evolution of the PSNR of the reconstructed images across training iterations for both strategies. In \Cref{fig:deblur-visualisation}, \Cref{fig:denoise-visualisation}, and \Cref{fig:inpaint-visualisation}, we further visualise five reconstructed sample images per task, obtained with the trained models under both strategies, along with their respective error maps, for both the training and validation datasets. For validation and testing, we utilise the trained network parameters $\theta$ and perform a standard forward pass to generate the reconstructions. The auxiliary variables $\Gaux$ are not explicitly optimised during inference; they are determined by the learned weights and the network architecture, consistent with conventional network evaluation.

Under lifted Bregman training we are able to recover recognisable digits even with $\lambda = 2 \times 10^{-1}$. This observation contrasts with the conventional setting and suggests that the lifted Bregman framework provides additional stability against the effects of overly aggressive shrinkage. As shown, conventional training methods struggle to learn meaningful reconstructions, highlighting that training this type of architecture poses a particularly challenging scenario for standard approaches. We evaluate conventional training across soft-thresholding parameters $\lambda \in \{2 \times 10^{-1}, 1 \times 10^{-1}, 5 \times 10^{-2}, 4 \times 10^{-2}, 2 \times 10^{-2}\}$ and Adam learning rates $\lr \in \{1 \times 10^{-3},5 \times 10^{-4}, 4 \times 10^{-4}\}$. Under this setting, recognisable digits are only recovered with $\lambda = 2 \times 10^{-2}$, across all tested learning rates. At larger values of $\lambda$, the reconstructions degrade substantially. Among the learning rates at $\lambda = 2 \times 10^{-2}$, $\lr = 5 \times 10^{-4}$ exhibit the most favourable training dynamics, reaching the lowest training loss after 50,000 epochs compared with $\lr = 1 \times 10^{-3}$ and $\lr = 4 \times 10^{-4}$. However, while the model achieves good training performance under this setting for both deblurring and denoising tasks, its validation performance is noticeably poorer compared to lifted Bregman training, indicating signs of overfitting as can be seen in in \Cref{fig:deblur-visualisation}, \Cref{fig:denoise-visualisation}, and \Cref{fig:inpaint-visualisation}.

\subsection{Inversion of Neural Networks}

We demonstrate the multilayer network inversion described in \Cref{sec:lifted-inversion} via the example of inverting an autoencoder $\net(\var)$ trained on the MNIST dataset. The network architecture is defined as
\[
\net(\var)=\text{Dec}\big(\text{Enc}(x,\boldsymbol{\Theta}_{\text{Enc}}),\boldsymbol{\Theta}_{\text{Dec}}\big),
\]

\noindent where $\text{Enc}(\cdot,\boldsymbol{\Theta}_{\text{Enc}})$ and $\text{Dec}(\cdot,\boldsymbol{\Theta}_{\text{Dec}})$ denote the encoder and decoder with parameters $\boldsymbol{\Theta}_{\text{Enc}}$ and $\boldsymbol{\Theta}_{\text{Dec}}$, respectively.

The encoder consists of three layers. The first two are convolutional layers (kernel $4\times4$ and stride $2$), each followed by a ReLU activation function. As the spatial resolution is reduced by a factor of two at each of these layers (from $28\times28$ to $14\times14$ to $7\times7$), the number of feature channels is doubled (from $1$ to $8$ to $16$). The third layer is a fully connected layer with weights $W_3 \in\mathbb{R}^{392\times784}$ and bias $b_3\in\mathbb{R}^{392}$ that maps the flattened feature vector to a 392 dimensional latent code.

The decoder also consists of three layers. The first is a fully connected layer with weights $W_4 \in \mathbb{R}^{784\times392}$ and bias $b_4 \in \mathbb{R}^{784}$ that expands the code and is followed by a ReLU activation function, after which the output is reshaped to $16\times7\times7$. This is followed by two transposed convolutional layers (kernel $4\times4$ and stride $2$) that successively upsample the feature maps, doubling the spatial resolution at each stage (from $7\times7$ to $14\times14$ to $28\times28$) while halving the channel width (from $16$ to $8$ to $1$). A ReLU activation function is also applied after the first transposed convolution.

\begin{figure}
  \centering
  \includegraphics[width=\linewidth]{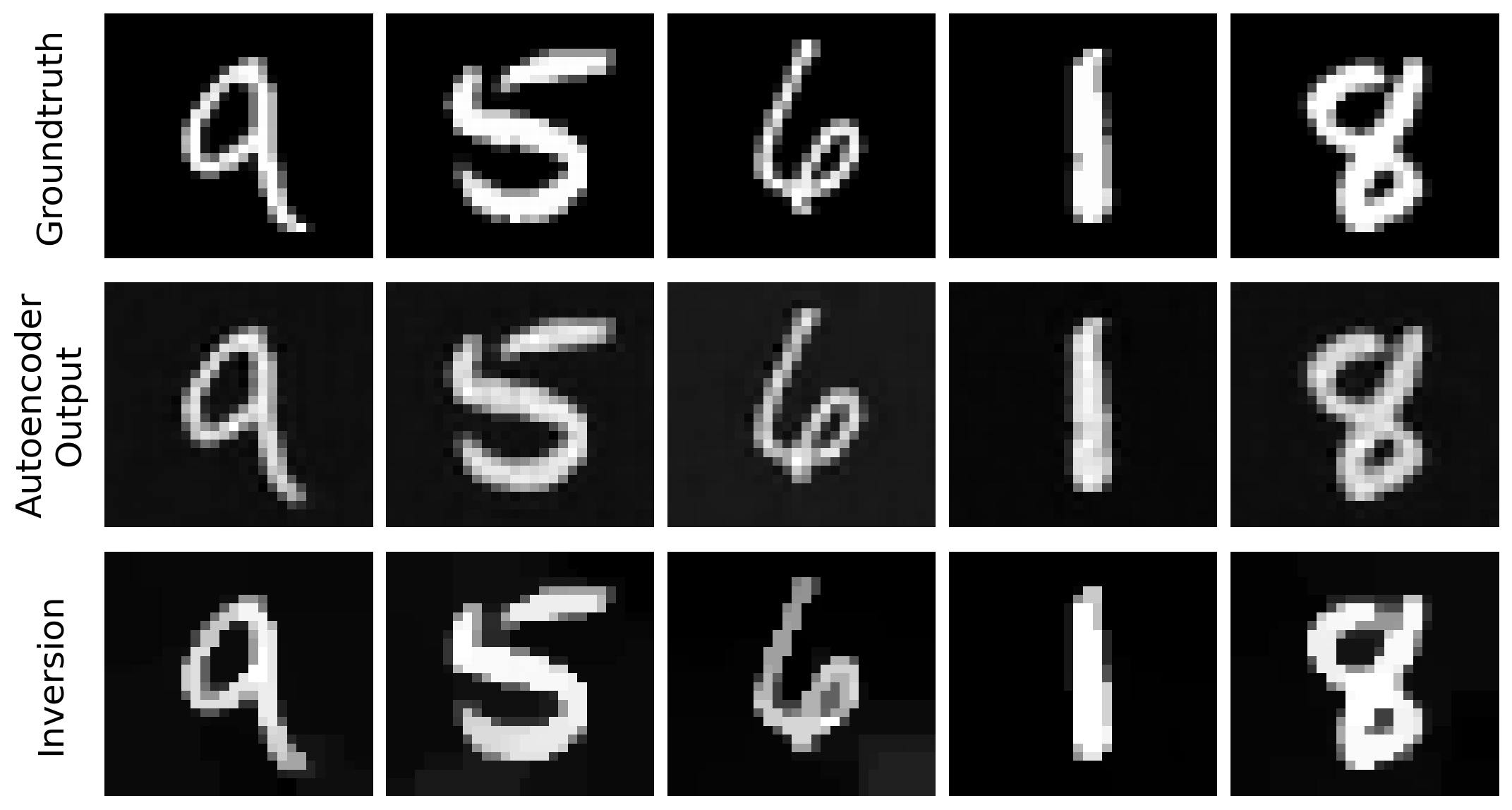}
  \caption{\textbf{(Network inversion image visualisations)} Comparison of the results of the network inversion (third row) with the output of the trained decoder (second row) on five MNIST sample images (first row). }
  \label{fig:Inversion}
\end{figure}

We train this six-layer autoencoder on the MNIST dataset by minimising the mean-squared error (MSE), using the Adam optimiser with learning rate $\lr = 1 \times 10^{-4}$. The output images from the trained network on the validation set are shown in the second row of \Cref{fig:Inversion}.

For the inversion experiment, we perturb the encoder output with zero mean Gaussian noise with standard deviation 0.1 to obtain the observation $y^\delta$, and we seek to reconstruct the corresponding input image $x$ from the relation
\[
\text{Enc}(x,\boldsymbol{\Theta}_{\text{Enc}})=y^\delta \,.
\]
Following the implementation details in \cite{wang2023liftedinversion}, we solve the corresponding convex optimisation problem \cref{eq:lifted_inversion} by a coordinate descent approach. This consists of a PDHG update step when optimising with respect to the input variable $x$, where we alternate between a descent step in the $x$ variable and an ascent step in the dual variable $z$, and proximal gradient updates when optimising for the auxiliary variables $\{\aux_1, \dots, \aux_{\nolayers-1}\}$, as shown in \cref{eq:mlp-x0-pdhg} and \cref{eq:x_l-proximal-gradient-specific-form}, respectively. That is,
\begin{subequations}
  \begin{align}
    x^{k + 1} &= x^k - \tau_{x} \left( \left( \text{prox}_{\Psi} \left(f(x^k, \Theta_{1}) \right) - u_{1}^k \right) \mathcal{J}_f^x(x^k, \Theta_{1}) + \reghyp \nabla^\top z^k \right) \, , \\
    z^{k + 1} &= \text{prox}_{\tau_z R^\ast}\left( z^k + \tau_z \reghyp \nabla  \left(  2 x^{k + 1} -  x^{k} \right) \right) \, .
\end{align}\label{eq:mlp-x0-pdhg}%
\end{subequations}
Moreover, we have
\begin{align}\label{eq:x_l-proximal-gradient-specific-form}
\begin{split}
    u_j^{k + 1} &= \text{prox}_{\kappa_j \Psi_j}\left( \kappa_j \left( u_j^k - \tau_{u_j} \left( \left( \text{prox}_{\Psi_j}\left( f(u_j^k, \Theta_{j + 1}) \right) - u_{j + 1}^k \right)  \right. \right. \right. \\
    &  \qquad \qquad \qquad \left. \left. \left. \mathcal{J}_f^u(u_j^k, \Theta_{j + 1}) - f(u_{j - 1}^{k}, \Theta_j) \right) \right) \right).
\end{split}
\end{align}
where $\kappa_j = \tau_{u_j}/(1 + \tau_{u_j})$, $\mathcal{J}_f^{u}$ and $\mathcal{J}_f^{x}$ denote the Jacobians of $f$ with respect to $u$ and $x$, respectively, and $\text{prox}_{\tau_z R^\ast}$ is the argument itself when the maximum of the Euclidean vector-norm per pixel is bounded by one; otherwise it is the projection onto the unit ball.

We define $R$ as $\| \nabla x \|_{2, 1}$. If we consider a two-dimensional scalar-valued image $x\in \mathbb{R}^{H \times W}$, we can define a finite forward difference discretisation of the gradient operator $\nabla: \mathbb{R}^{H \times W} \rightarrow \mathbb{R}^{H \times W \times 2}$ as
\begin{align*}
    &(\nabla x)_{i,j,1} = \begin{cases}
        x_{i+1,j} - x_{i,j} \;\; \text{if } 1 \leq i < H, \\
        0 \;\; \text{else,}\\
    \end{cases}\\
    &(\nabla x)_{i,j,2} =
    \begin{cases}
        x_{i,j+1} - x_{i,j} \;\; \text{if } 1 \leq j < W, \\
        0 \;\; \text{else.}
    \end{cases}
\end{align*}
This implementation allows for simultaneous updates of all auxiliary variables, unlike the sequential approach in~\cite{wang2023liftedinversion}.

The step-size parameters for the PDHG update step are chosen as $\tau_x = 1.99/\|W_{1}\|_2^{2}$ and $\tau_z = 1/(8\,\reghyp)$ with regularisation parameter $\reghyp = 7 \times 10^{-2}$, which we select empirically based on the visual quality of the inverted images. The step-size parameters for the proximal gradient updates $\tau_{u_j}$ for each layer, are set as $\tau_{u_j}={1.99}/{\|W_{j+1}\|_2^{2}}$. All variables $\{\aux_0, \dots, \aux_{\nolayers-1}\}$ are initialised with zero vectors.

With this approach, we reconstruct the input image $x$ from the noisy encoder observation $y^\delta$, which is visualised in \Cref{fig:Inversion}. The first row shows the ground truth MNIST images, the second row the autoencoder output on the validation dataset and the last row the results from the network inversion for five example images. Here, the PDHG is set to a maximum number of iterations of 1,000 or stops when the improvements on $x$ and $z$ are less than $1 \times 10^{-5}$ in norm and the coordinate descent algorithm stops after 500 iterations.

\begin{figure}
  \centering
  \includegraphics[width=\linewidth]{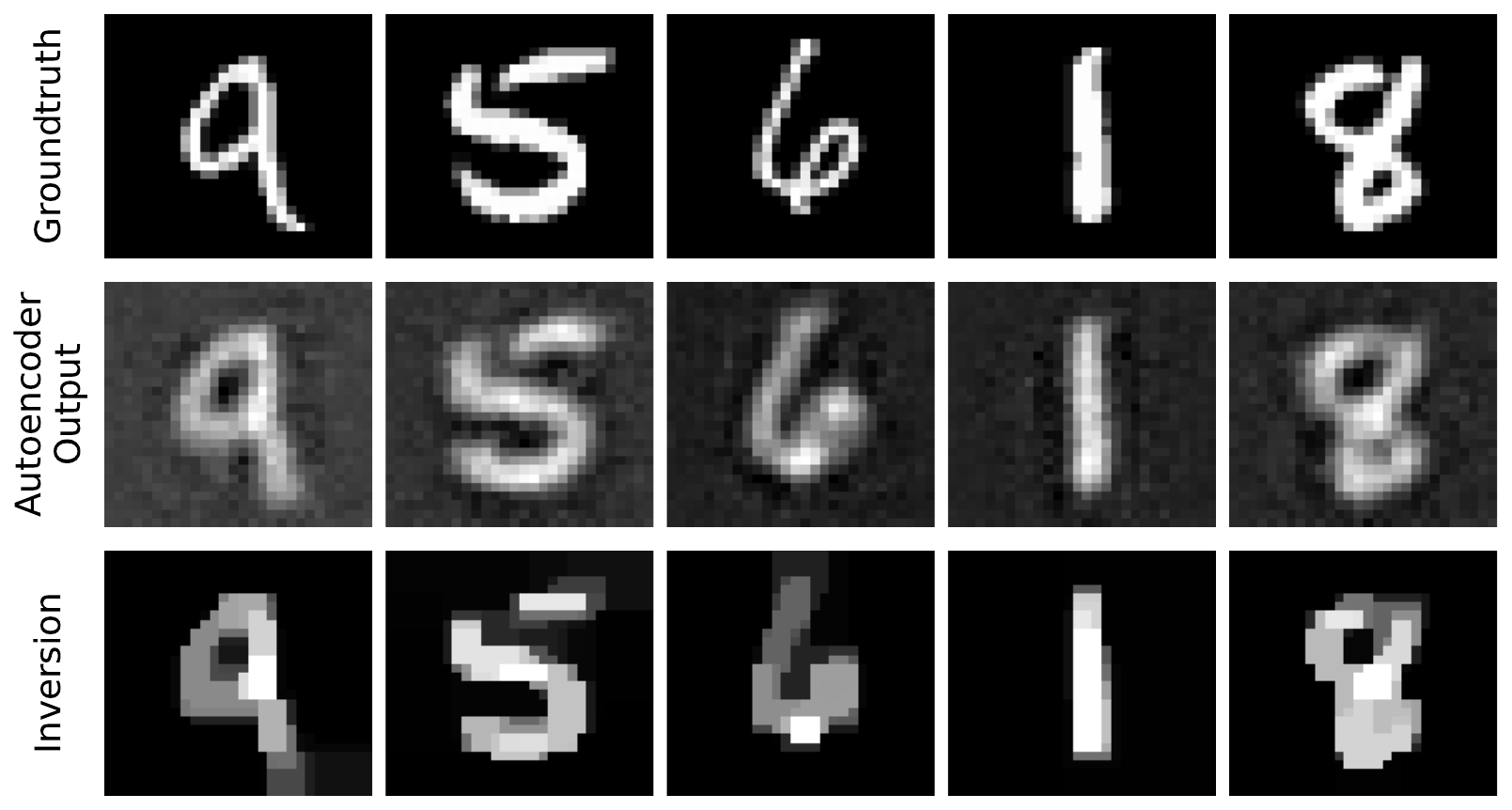}
  \caption{\textbf{(Severely under-trained network inversion image visualisations)} Comparison of the results of the network inversion (third row) with the output of the severely under-trained decoder (second row) on five MNIST sample images (first row).}
  \label{fig:S_Undertrained}
\end{figure}

\begin{figure}
  \centering
  \includegraphics[width=\linewidth]{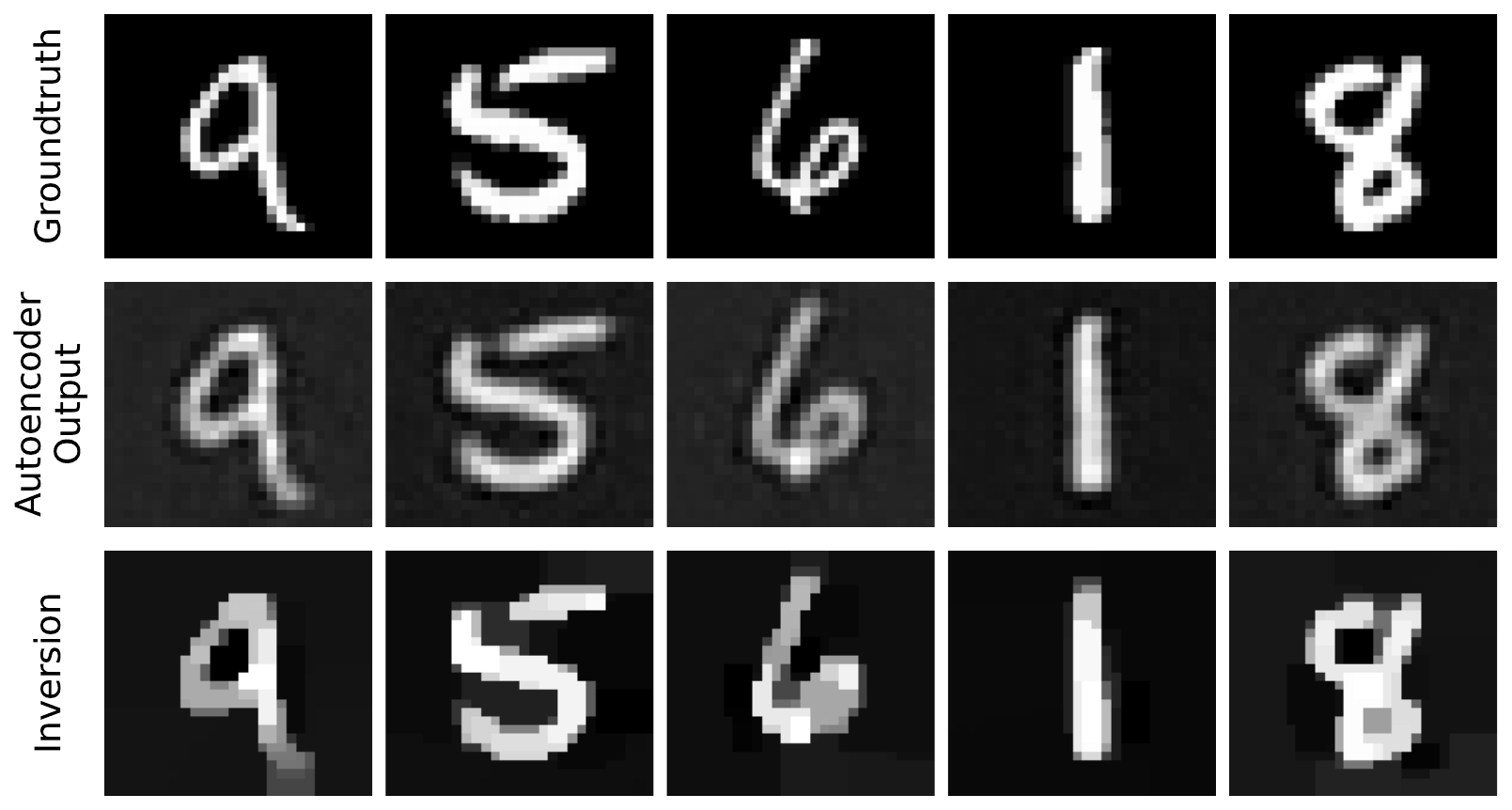}
  \caption{\textbf{(Moderately under-trained network inversion image visualisations)} Comparison of the results of the network inversion (third row) with the output of the under-trained decoder (second row) on five MNIST sample images (first row).}
  \label{fig:Undertrained}
\end{figure}

We also study how the training quality of the autoencoder influences the performance of the lifted inversion framework. We consider three training levels defined by the MSE reconstruction loss: (i) a severely under-trained model halted when the training loss reached $8.3 \times 10^{-3}$ (\Cref{fig:S_Undertrained}), (ii) a moderately under-trained model halted when the training loss reached $3.1 \times 10^{-3}$ (\Cref{fig:Undertrained}), and (iii) the near-converged model previously presented in \Cref{fig:Inversion} with loss $5 \times 10^{-4}$. All models are equivalent and trained with Adam to optimise the MSE loss.

For the inversion corresponding to (i) we set the regularisation parameter $\reghyp$ to $8 \times 10^{-2}$ and for (ii) to $ 6 \times 10^{-2}$. These values are set empirically based on the visual quality of the inverted images. Despite the degraded autoencoder reconstructions at higher loss levels, the lifted inversion framework still recovers an approximate inverse of the encoder; however, not as accurate as the baseline as shown in the last rows of \Cref{fig:Inversion}, \Cref{fig:S_Undertrained}, and \Cref{fig:Undertrained}, respectively.

\section{Conclusions \& Outlook}

This chapter presented a unified framework for the lifted training and inversion of neural networks. By reformulating network architectures using a block operator structure, we demonstrated that diverse models and various lifted training strategies (MAC-QP, Fenchel, Bregman) can be understood within a single mathematical paradigm. Lifted methods offer significant advantages, particularly in handling non-smooth activations, enabling distributed computation, and avoiding the need for back-propagation. Furthermore, the application of this framework to network inversion provides a robust, regularisation-based approach.

Future work will focus on addressing the memory overhead associated with auxiliary variables, extending theoretical guarantees to deep networks, and exploring the scalability of these methods to large-scale architectures and complex inverse problems.

\subsection{Limitations and Open Challenges}\label{subsec:limitations}

While the unified framework and the lifted training methodologies presented offer significant advantages, several limitations and open challenges remain.

\begin{itemize}
    \item \textbf{Computational and Memory Overhead:} Lifted methods inherently introduce a large number of auxiliary variables (one set for every training sample and every layer). While the implicit stochastic gradient method (ISGM) detailed in \Cref{sec:implementation} helps manage this by processing mini-batches, the memory footprint during the inner optimisation loop is still significantly larger than standard back-propagation. This can pose challenges when training very deep networks or handling high-dimensional data on memory-constrained hardware.
    \item \textbf{Theoretical Guarantees for Deep Networks:} While strong convergence results exist for the lifted inversion of single-layer networks (cf. \Cref{sec:lifted-inversion}), comprehensive theoretical guarantees for the training and inversion of deep, multi-layer networks within the lifted framework are still largely undeveloped due to the non-convexity of the joint optimisation problem.
    \item \textbf{Scalability and Scope:} The practical efficacy of lifted methods has primarily been demonstrated on moderate-sized MLPs, mostly focusing on linear inverse problems in imaging. Their scalability and performance compared to highly optimised back-propagation implementations for very large-scale architectures (e.g., large transformers) or complex non-linear inverse problems have yet to be fully explored.
\end{itemize}

\section{Acknowledgments}
Audrey Repetti and Xiaoyu Wang acknowledge the support by the EPSRC grant EP/X028860. Azhir Mahmood acknowledges support from the EPSRC Centre for Doctoral Training in Intelligent, Integrated Imaging In Healthcare (i4health), EP/S021930/1. Andreas Mang acknowledges support by the National Science Foundation (NSF) through the grant DMS-2145845. Alexandra Valavanis acknowledges support from the Queen Mary Principal’s Research Studentship at Queen Mary University of London (School of Mathematical Sciences). Any opinions, findings, and conclusions or recommendations expressed herein are those of the authors and do not necessarily reflect the views of the NSF or the EPSRC. Additionally, we thank Danilo Riccio for providing valuable feedback. Finally, we thank PhysicsX for providing computational resources that supported this work.

\bibliographystyle{elsarticle-num} \bibliography{references.bib}



\Backmatter
\end{document}